\documentclass[10pt,twocolumn,letterpaper]{article}

\usepackage[pagenumbers]{robosense}

\usepackage[table]{xcolor}
\usepackage{microtype}
\usepackage{graphicx}
\usepackage{booktabs}
\usepackage{amsmath}
\usepackage{amssymb}
\usepackage{mathtools}
\usepackage{amsthm}

\usepackage{color}
\usepackage{enumitem}
\usepackage{multirow}
\usepackage{makecell}

\usepackage{algorithmic}
\usepackage{algorithm}
\usepackage{etoolbox,siunitx}

\usepackage{fontawesome}
\usepackage{academicons}

\usepackage{tabularx}

\definecolor{robo_blue}{RGB}{66, 133, 244}
\definecolor{robo_red}{RGB}{231, 66, 52}
\definecolor{robo_yellow}{RGB}{251, 189, 5}
\definecolor{robo_green}{RGB}{51, 168, 82}
\definecolor{robo_gray}{RGB}{165, 165, 165}

\definecolor{lblue}{RGB}{66, 133, 244}

\usepackage[pagebackref=false,breaklinks,colorlinks,citecolor=lblue]{hyperref}

\def\paperID{1}
\def\confName{RoboSense}
\def\confYear{2025}

\newcommand\blfootnote[1]{%
\begingroup
\renewcommand\thefootnote{}{}\footnote{#1}%
\addtocounter{footnote}{-1}%
\endgroup
}

\title{\includegraphics[height=0.5cm]{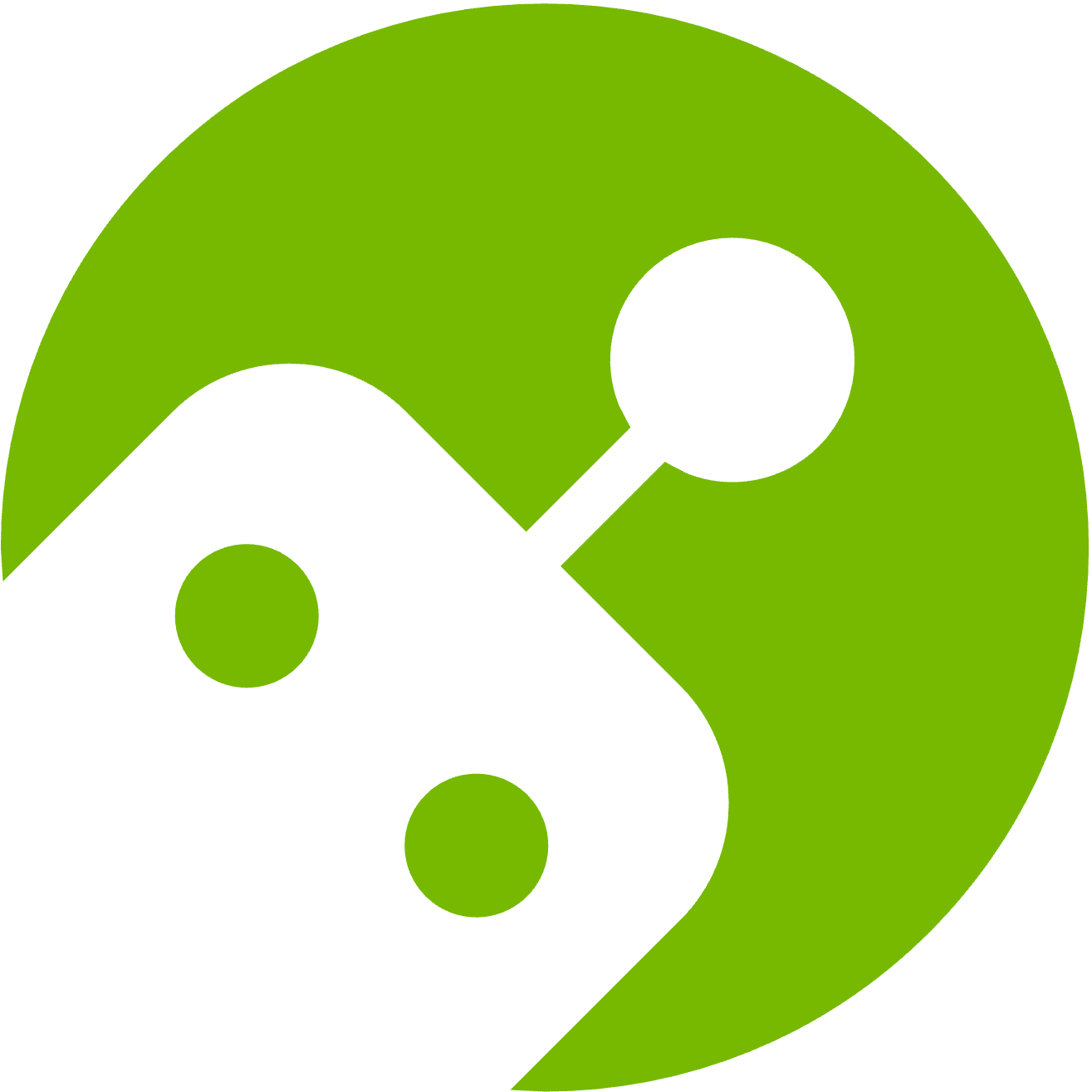}~The RoboSense Challenge: 
\\
Sense Anything, Navigate Anywhere, Adapt Across Platforms
}

\author{
    \textbf{Challenge \& Workshop Organizers}\\
    Lingdong Kong$^*$ \quad Shaoyuan Xie$^*$ \quad Zeying Gong$^*$ \quad Ye Li$^*$ \quad Meng Chu$^*$ \quad Ao Liang$^*$\\ 
    Yuhao Dong \quad Tianshuai Hu \quad Ronghe Qiu \quad Rong Li \quad Hanjiang Hu \quad Dongyue Lu \quad Wei Yin\\
    Wenhao Ding \quad Linfeng Li \quad Hang Song \quad Wenwei Zhang \quad Yuexin Ma \quad Junwei Liang\\
    Zhedong Zheng \quad Lai Xing Ng \quad
    Benoit R. Cottereau \quad Wei Tsang Ooi \quad Ziwei Liu\\
    \url{https://robosense2025.github.io}\\
\\
    \textbf{Technical Committee}\\
    Zhanpeng Zhang \quad Weichao Qiu \quad Wei Zhang\\
\\
    \textbf{Challenge Participants}\\
    Ji Ao \quad Jiangpeng Zheng \quad Siyu Wang \quad Guang Yang \quad Zihao Zhang \quad Yu Zhong
    \\
    Enzhu Gao \quad Xinhan Zheng \quad Xueting Wang \quad Shouming Li \quad Yunkai Gao \quad Siming Lan
    \\
    Mingfei Han \quad Xing Hu \quad Dusan Malic \quad Christian Fruhwirth-Reisinger \quad Alexander Prutsch
    \\
    Wei Lin \quad Samuel Schulter \quad Horst Possegger \quad Linfeng Li \quad Jian Zhao \quad Zepeng Yang
    \\
    Yuhang Song \quad Bojun Lin \quad Tianle Zhang \quad Yuchen Yuan \quad Chi Zhang \quad Xuelong Li
    \\
    Youngseok Kim \quad Sihwan Hwang \quad Hyeonjun Jeong \quad Aodi Wu \quad Xubo Luo \quad Erjia Xiao
    \\
    Lingfeng Zhang \quad Yingbo Tang \quad Hao Cheng \quad Renjing Xu \quad Wenbo Ding \quad Lei Zhou
    \\
    Long Chen \quad Hangjun Ye \quad Xiaoshuai Hao \quad Shuangzhi Li \quad Junlong Shen \quad Xingyu Li
    \\
    Hao Ruan \quad Jinliang Lin \quad Zhiming Luo \quad Yu Zang \quad Cheng Wang \quad Hanshi Wang
    \\
    Xijie Gong \quad Yixiang Yang \quad Qianli Ma \quad Zhipeng Zhang \quad Wenxiang Shi \quad Jingmeng Zhou
    \\
    Weijun Zeng \quad Kexin Xu \quad Yuchen Zhang \quad Haoxiang Fu \quad Ruibin Hu \quad Yanbiao Ma
    \\
    Xiyan Feng \quad Wenbo Zhang \quad Lu Zhang \quad Yunzhi Zhuge \quad Huchuan Lu \quad You He
    \\
    Seungjun Yu \quad Junsung Park \quad Youngsun Lim \quad Hyunjung Shim \quad Faduo Liang
    \\
    Zihang Wang \quad Yiming Peng \quad Guanyu Zong \quad Xu Li \quad Binghao Wang \quad Hao Wei
    \\
    Yongxin Ma \quad Yunke Shi \quad Shuaipeng Liu \quad Dong Kong \quad Yongchun Lin \quad Huitong Yang
    \\
    Liang Lei \quad Haoang Li \quad Xinliang Zhang \quad Zhiyong Wang \quad Xiaofeng Wang \quad Yuxia Fu
    \\
    Yadan Luo \quad Djamahl Etchegaray \quad Yang Li \quad Congfei Li \quad Yuxiang Sun \quad Wenkai Zhu
    \\
    Wang Xu \quad Linru Li \quad Longjie Liao \quad Jun Yan \quad Benwu Wang \quad Xueliang Ren \quad Xiaoyu Yue
    \\
    Jixian Zheng \quad Jinfeng Wu \quad Shurui Qin \quad Wei Cong \quad Yao He
}

\begin{document}

\maketitle

\blfootnote{\includegraphics[height=0.28cm]{figures/logo.png}~Official IROS 2025 RoboSense Challenge Report.~ ${(*)}$ Lingdong, Shaoyuan, Zeying, Ye, Meng, and Ao contributed equally to this work.}

\begin{abstract}
Autonomous systems are increasingly deployed in open and dynamic environments -- from city streets to aerial and indoor spaces -- where perception models must remain reliable under sensor noise, environmental variation, and platform shifts. However, even state-of-the-art methods often degrade under unseen conditions, highlighting the need for robust and generalizable robot sensing. The RoboSense 2025 Challenge is designed to advance robustness and adaptability in robot perception across diverse sensing scenarios. It unifies five complementary research tracks spanning language-grounded decision making, socially compliant navigation, sensor configuration generalization, cross-view and cross-modal correspondence, and cross-platform 3D perception. Together, these tasks form a comprehensive benchmark for evaluating real-world sensing reliability under domain shifts, sensor failures, and platform discrepancies. RoboSense 2025 provides standardized datasets, baseline models, and unified evaluation protocols, enabling large-scale and reproducible comparison of robust perception methods. The challenge attracted 143 teams from 85 institutions across 16 countries, reflecting broad community engagement. By consolidating insights from 23 winning solutions, this report highlights emerging methodological trends, shared design principles, and open challenges across all tracks, marking a step toward building robots that can sense reliably, act robustly, and adapt across platforms in real-world environments.
\end{abstract}

\begin{figure*}[t]
    \centering
    \includegraphics[width=\linewidth]{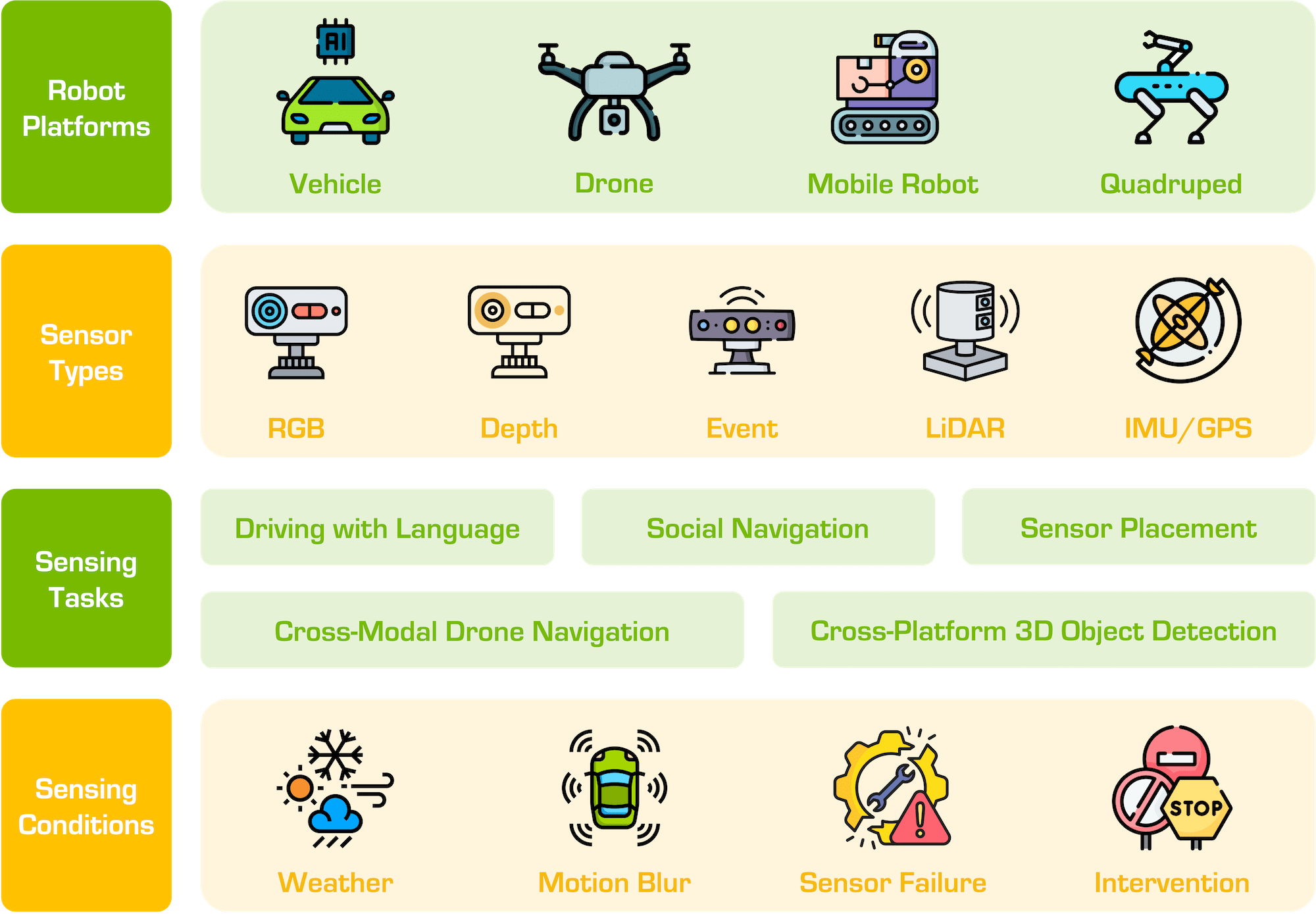}
    \vspace{-0.55cm}
    \caption{\textbf{Overview of the RoboSense 2025 Challenge.} This competition benchmarks robust robot sensing across platforms (vehicles, drones, mobile robots, and quadrupeds), sensor modalities (RGB, depth, LiDAR, language), task settings (driving with language, social navigation, sensor placement, cross-view matching, and cross-platform 3D detection), and adverse conditions (sensor noise, viewpoint changes, environmental corruptions, and domain shifts). Together, the five tracks evaluate the resilience and adaptability of modern perception systems under real-world variations.}
    \label{fig:teaser}
\end{figure*}

\section{Introduction}
\label{sec:intro}

Modern autonomous systems are increasingly deployed in open, dynamic, and unpredictable environments \cite{wang2026forging}. From autonomous vehicles navigating crowded urban streets \cite{xie2025drivebench,li2024place3d,bian2025dynamiccity,kong2022laserMix,zhu2025spiral}, to aerial drones operating under drastic viewpoint changes \cite{chandarana2017fly,zhang2025your,kong2025eventfly,liang2025pi3det,chu2024geotext-1652}, and legged robots traversing complex indoor spaces \cite{gong2025stairway,gong2025cognition,li2025_3eed,chaney2023m3ed}, these agents must perceive, reason, and act reliably under a wide range of conditions \cite{survey_vla4ad,wang2025alpamayo-r1,xu2025wod,hao2025mimo-embodied}. To achieve this, contemporary robotic systems rely on a diverse array of sensing modalities, including RGB cameras, LiDARs, depth sensors, and increasingly, language inputs, to support perception, decision making, and interaction \cite{caesar2020nuscenes,sun2020scalability,geiger2012kitti}.  

Despite remarkable progress in perception accuracy on curated benchmarks, the real-world reliability of these systems remains fragile \cite{xie2023robobev,kong2023robo3d,hao2024is,beemelmanns2024multicorrupt}. Models trained under specific platforms, viewpoints, weather conditions, or sensor configurations often exhibit severe performance degradation when exposed to distribution shifts or unseen environments \cite{hao2025safemap,hao2025mapfusion,hao2024mbfusion,hao2025msc-bench,wang2025nuc-net,wang2025monomrn,liu2025lalalidar,sun2024lidarseg,liang2025lidarcrafter}. Such failures are particularly concerning in safety-critical robotic applications, where robustness to sensor noise, environmental variation, and platform discrepancies is not optional but crucial \cite{survey_vla4ad,mavrogiannis2023core,xu2025visual}. These challenges underscore a growing gap between benchmark performance and real-world deployment, motivating the need for systematic evaluation of robustness and generalization in robot sensing \cite{robodrive_challenge_2024,robodepth_challenge_2023}.

The RoboSense Challenge is designed to confront this gap directly \cite{robosense_challenge_2025}. It establishes a unified framework for evaluating the robustness, adaptability, and cross-platform generalization of robot perception systems across diverse sensing scenarios \cite{xie2025drivebench,gong2025cognition,li2024place3d,chu2024geotext-1652,liang2025pi3det}. In contrast to traditional benchmarks that emphasize accuracy under fixed settings, RoboSense focuses on resilience under variation: assessing how models behave when confronted with novel viewpoints, degraded signals, unseen modalities, or platform shifts \cite{hu2022investigating,kong2025multi,jaritz2020xMUDA,kong2024robodepth}. This emphasis reflects a broader paradigm shift in embodied AI -- from optimizing performance in controlled datasets toward building agents that can perceive and act reliably in the wild \cite{hu2022seasondepth,hu2023towards,kong2023conDA,xie2025benchmarking}.

\subsection*{From RoboDrive to RoboSense}
The RoboSense 2025 Challenge builds upon and significantly extends the RoboDrive 2024 Challenge \cite{robodrive_challenge_2024}, which established the first large-scale benchmark for out-of-distribution (OoD) perception in autonomous driving at ICRA 2024. RoboDrive highlighted the vulnerability of state-of-the-art perception models to common corruptions and sensor failures in driving scenarios. While impactful, its scope was primarily limited to vehicle-centric perception and driving tasks.

RoboSense expands this vision to a broader, multi-domain setting that encompasses grounded reasoning, embodied navigation, sensor configuration generalization, and cross-platform perception across heterogeneous robotic platforms \cite{xie2025drivebench,gong2025cognition,li2024place3d,chu2024geotext-1652,liang2025pi3det}. This expansion reflects an important trend in embodied intelligence: the convergence of vision, language, and action. By unifying multiple perception tasks under a common robustness objective, RoboSense evaluates not only how well models perform within a single domain, but also how effectively they transfer, adapt, and remain interpretable across modalities, platforms, and environments, often without retraining or manual recalibration.

\subsection*{Challenge Tracks}
RoboSense 2025 features five complementary tracks, each targeting a critical frontier in robust robot sensing:
\begin{itemize}
    \item \textbf{Track 1: Driving with Language}\\
    This track investigates how vision-language models understand, reason, and respond to natural-language driving queries under visual corruptions and domain shifts. Participants develop models that answer perception, prediction, and planning questions based on multi-view visual inputs. Building on DriveBench \cite{xie2025drivebench} and RoboBEV \cite{xie2025benchmarking}, the track emphasizes task-specific prompting, spatial reasoning, and robustness in dynamic traffic scenes.

    \item \textbf{Track 2: Social Navigation}\\
    Beyond geometric path planning, autonomous agents must navigate in ways that respect human comfort and social norms. This track evaluates RGBD-based navigation systems capable of interpreting human behaviors and producing socially compliant trajectories. Built upon Falcon \cite{gong2025cognition}, this track benchmarks agents in photo-realistic, dynamic indoor environments from the Social-HM3D datasets.

    \item \textbf{Track 3: Sensor Placement}\\
    Real-world deployments often involve variations in sensor layout, calibration, and positioning. This track examines whether LiDAR-based 3D object detection models can generalize across diverse sensor placements without retraining, building on the Place3D \cite{li2024place3d} benchmark to stress-test robustness under configuration shifts.

    \item \textbf{Track 4: Cross-Modal Drone Navigation}\\
    This track focuses on language-guided cross-view retrieval, requiring models to associate aerial or satellite imagery with ground-level scenes based on free-form textual descriptions. Following GeoText-1652 \cite{chu2024geotext-1652}, it challenges models to perform spatial reasoning and cross-view alignment under strong viewpoint and appearance variations.

    \item \textbf{Track 5: Cross-Platform 3D Object Detection}\\
    Moving beyond vehicle-centric perception, this track evaluates 3D object detectors that transfer knowledge across heterogeneous platforms, including vehicles, drones, and quadruped robots. By extending the Pi3DET \cite{liang2025pi3det} dataset and baseline, it highlights geometric and data distribution gaps arising from platform diversity.
\end{itemize}

Together, these five tracks span reasoning, interaction, adaptation, and multi-platform perception, forming a comprehensive benchmark for evaluating the robustness and generalizability of modern robot sensing systems.

\subsection*{Scale \& Participation}
RoboSense 2025 attracted strong global participation, with $143$ registered teams from $85$ institutions, including $66$ universities and $19$ companies, across $16$ countries. This broad engagement reflects growing interest from the computer vision, robotics, and multimodal learning communities in robustness-oriented embodied AI research.

Across all tracks, participants developed methods capable of addressing real-world sensing challenges such as sensor misalignment, environmental corruptions, and domain shifts, while maintaining competitive performance on standard benchmarks. The diversity of participating teams, spanning academic research groups and industrial R\&D labs, further underscores the relevance of RoboSense to both foundational research and practical robotic deployment.
\section{Related Work}
\label{sec:related_work}

\subsection{Driving with Language}
Vision-language models (VLMs) \cite{liu2023llava, liu2024llava1.5, liu2024llavanext, qwen, wang2024qwen2, abdin2024phi} have demonstrated impressive reasoning, grounding, and multimodal understanding capabilities across a wide range of tasks \cite{brohan2023rt, chen2024spatialvlm, tian2024drivevlm, stone2023open, hong2024cogagent, yang2025octopus, dong2024insight, chen2023clip2Scene, chen2023towards, liu2023segment, liu2023uniseg, peng2024sam, wang2025pointlora, wang2025pixelthink, liu2024chain, liu2024coarse, chen2024drivinggpt, xu2024vlm, cui2024drive, yan2025ad-r1}. These advances have stimulated growing interest in applying VLMs to autonomous driving, where language offers a natural interface for querying perception, explaining decisions, and supporting high-level planning \cite{yang2023survey,li20224bev}. Compared with purely end-to-end perception or control models, language-driven reasoning provides greater interpretability and flexibility, enabling systems to articulate intermediate logic, uncertainty, and contextual constraints \cite{survey_vla4ad,wang2025alpamayo-r1,xie2025drivebench,survey_3d_4d_world_models,sima2024drivelm,worldlens}.

Motivated by these advantages, several benchmarks and datasets have emerged to evaluate VLMs in driving-related tasks, ranging from textual annotation and explainable perception to language-conditioned prediction and planning \cite{kim2018textual, xu2020explainable, wu2023language, malla2023drama}. These efforts have demonstrated promising capabilities in structured reasoning and cross-modal grounding \cite{qian2024nuscenes,li2025seeground,xu2025wod}. However, most existing evaluations remain focused on clean or narrowly controlled settings, leaving the robustness of VLM-based driving systems under real-world perturbations largely unexplored.

More broadly, deep neural networks are known to exhibit brittleness under out-of-distribution (OoD) conditions, a limitation that is especially critical in safety-sensitive domains such as autonomous driving \cite{xie2025benchmarking, kong2023robo3d, kong2024robodepth}. Recent studies have investigated hallucination, calibration, and trustworthiness issues in large multimodal models \cite{ji2023survey, wang2023evaluation, li2023evaluating, tu2023many}, revealing systematic failure modes when visual evidence is ambiguous or corrupted. Complementary work further shows that VLMs suffer from severe hallucination and grounding errors in driving scenarios under visual degradations \cite{xie2025drivebench}. These findings highlight a fundamental gap between impressive reasoning performance on curated benchmarks and reliable deployment under adverse sensing conditions. RoboSense addresses this gap by explicitly benchmarking the robustness of driving-oriented VLMs under controlled corruptions, sensor degradation, and domain shifts, emphasizing grounded reasoning and reliability rather than raw accuracy alone.

\subsection{Social Navigation}
Social navigation is a fundamental capability for autonomous robots operating in human-populated environments, where agents must balance task efficiency with safety, comfort, and social norms \cite{gong2025cognition}. Classical approaches primarily rely on geometric motion planning and collision avoidance, modeling pedestrian dynamics through social force models, velocity obstacles, and reactive control strategies \cite{helbing1995social, berg2008reciprocal, fox1997dynamic}. While these methods provide strong safety guarantees and computational efficiency, they often struggle to capture high-level human intent, social conventions, and long-term interaction dynamics in dense environments.

The availability of large-scale simulators and standardized benchmarks has catalyzed the development of learning-based social navigation systems \cite{biswas2021socnavbench, shen2021igibson, savva2019habitat}. These environments enable the study of complex multi-agent interactions under realistic visual and physical dynamics. Existing methods broadly fall into two categories. Model-based approaches explicitly encode interaction rules using social force models or reciprocal collision avoidance \cite{helbing1995social, wei2023transvae, ferrer2013robot, mavrogiannis2023core}, enabling interpretable and controllable behavior. In contrast, learning-based approaches leverage deep reinforcement learning to implicitly learn social behaviors from data \cite{chen2017decentralized, chen2017socially, everett2018motion}. Representative systems such as CADRL and Socially-Aware CADRL train policies to navigate among dynamic agents through value function approximation and policy optimization.

Recent advances further incorporate relational reasoning via graph neural networks and attention mechanisms to model complex crowd dynamics \cite{fan2020relational}. Emerging work also explores integrating vision-language models to inject semantic understanding and common-sense reasoning into navigation policies \cite{zhang2025nava, zhang2025mapnav, song2024vlmsocial, luo2024gson, gong2025stairway}, enabling agents to reason about social roles, intentions, and contextual cues. 

Despite this progress, robust evaluation of socially compliant navigation under diverse layouts, occlusions, and human motion patterns remains limited. RoboSense complements existing benchmarks by emphasizing robustness, scalability, and reproducibility in socially grounded navigation under realistic sensory uncertainty and dynamic interaction complexity.

\subsection{Sensor Placement}
Sensor configuration plays a critical role in determining the robustness, reliability and effectiveness of perception systems in autonomous driving and robotics \cite{joshi2008sensor, xu2022optimization,hu2022robustness, hu2023robustness}. Beyond sensor quality itself, the spatial arrangement, orientation, calibration, and redundancy of sensors directly affect observability, coverage, and downstream algorithmic performance \cite{hu2022investigating, li2024optimizing,hao2024mapdistill,wang2024voxel,cao2022monoscene}. Although sensor placement has a long history in robotics and control \cite{Vitus-RSS-10,1262512}, its systematic study in large-scale autonomous driving has only gained traction in recent years \cite{liu2019should,xu2022safebench}.

Early work primarily examined multi-sensor configurations mounted on ego vehicles, analyzing how multiple LiDARs or camera-LiDAR layouts improve detection accuracy and coverage \cite{hu2022investigating, li2024place3d, li2024influence, li2025unidrive}. These studies highlight the importance of complementary viewpoints and sensor diversity for mitigating occlusion and sparsity. More recent research has extended sensor placement to infrastructure-supported perception, exploring roadside LiDAR deployment for vehicle-to-everything (V2X) perception and cooperative sensing \cite{jin2022roadside, kim2023placement, cai2023analyzing, jiang2023optimizing}. Such settings introduce new trade-offs between coverage, cost, scalability, and communication latency.

Despite substantial progress in sensor placement strategies, most existing work implicitly assumes that perception models are trained and tuned for a fixed sensor configuration. In practice, however, sensor layouts frequently vary across vehicles, fleets, and deployment environments. The ability of perception models to generalize across multiple heterogeneous or evolving sensor configurations and viewpoints remains underexplored \cite{li2025unidrive, klinghoffer2023towards, hu2024pixel, zuo2025dvgt}. RoboSense directly targets this gap by evaluating robustness under cross-placement shifts, enabling systematic analysis of sensor-agnostic perception and transferable spatial representations.

\subsection{Cross-Modal Drone Navigation}
Cross-modal drone navigation studies how unmanned aerial vehicles (UAVs) perceive, reason, and act based on multimodal inputs such as images, depth, point clouds, videos, and natural language instructions. Early systems focused on mapping textual commands to executable flight actions using rule-based parsing, template matching, and symbolic planning \cite{huang2019flight, chandarana2017fly, 9103346}. With improved simulation environments and data availability, learning-based approaches emerged, enabling end-to-end grounding of language into trajectories via imitation learning or reinforcement learning \cite{blukis2019learning, 9155522}.

The broader literature on vision-and-language navigation (VLN) \cite{anderson2018vision, zhu2020vision, qi2021object, hong2021vln, zhang2025your, wang2021structured, hao2020towards, majumdar2020improving, thomason2020vision, georgakis2022cross} provides strong foundations for multimodal grounding and embodied reasoning. However, most VLN benchmarks are limited to ground-level indoor environments with relatively stable viewpoints and constrained scale variation. Extending these paradigms to aerial platforms introduces additional challenges, including extreme viewpoint changes, altitude-dependent scale shifts, motion blur, and partial observability.

Recent works address these challenges by incorporating geometric reasoning and cross-view alignment. Video2BEV \cite{ju2025video2bevtransformingdronevideos} reconstructs 3D scenes via Gaussian Splatting to generate consistent BEV representations for drone geolocalization. SaLPN \cite{chen2025scaleadaptiveuavgeolocalizationheightaware} dynamically adapts feature partitions based on altitude to mitigate scale inconsistency. ``Where am I'' \cite{ye2025icrossviewgeolocalizationnatural} formulates language-guided cross-view retrieval, while MMGeo \cite{Ji_2025_ICCV} unifies multimodal queries across image, text, depth, and point cloud modalities. Beyond perception, LLM-integrated UAV frameworks further explore memory, tool usage, and long-horizon reasoning for autonomous agents \cite{Tian_2025}. RoboSense builds upon these advances by providing a standardized benchmark for evaluating cross-view robustness and multimodal grounding under realistic aerial sensing variability.

\subsection{Cross-Platform 3D Object Detection}
LiDAR-based 3D object detection remains a core component of autonomous perception, enabling accurate localization and geometric understanding of surrounding environments \cite{mao20233d, qian20223d, wang2023multi, kong2023rethinking}. Existing approaches span grid-based representations \cite{deng2021voxel, li2022unifying, lee2024re, fan2021rangedet, zhang2022ri, wang2024club, song2024graphbev, li2024bevnext, shi2022pillarnet, chen2020every, zhu2021cylindrical}, point-based models \cite{qi2017pointnet++, yang20203dssd, yang2022dbq}, and hybrid architectures \cite{shishaoshuai2020pv, shi2023pv, li2022voxel}. These methods have been extensively benchmarked on vehicle-centric datasets such as KITTI \cite{geiger2012kitti}, nuScenes \cite{caesar2020nuscenes,fong2022Panoptic-nuScenes}, and Waymo Open \cite{sun2020scalability}, driving significant improvements in accuracy, efficiency, and robustness \cite{li2024bevnext, kong2023robo3d, robodrive_challenge_2024, ye2020hvnet, yang2018pixor, hong20224dDSNet, xu2024superflow, xu2025limoe}.

However, most existing benchmarks assume similar sensing platforms and operating conditions \cite{liang2023spsnet,liang2024boosting,liang2024suprnet,xu2025frnet}. As perception systems are increasingly deployed on heterogeneous robots, including drones and legged platforms, the ability to generalize across platforms becomes critical \cite{liang2025pi3det, xu2025lima}. Cross-platform scenarios introduce substantial distribution shifts arising from viewpoint geometry, motion dynamics, LiDAR sampling patterns, and environmental context.

Prior work on cross-dataset adaptation primarily addresses domain shifts within vehicle datasets using pseudo-labeling, self-training, and knowledge distillation \cite{yang2021st3d, yang2022st3d++, zhang2024detect, yuan2024reg, zhang2023uni3d, hu2023density}. While effective in limited settings, these techniques do not fully capture the broader heterogeneity of cross-platform sensing. Recent discussions highlight the need for unified benchmarks that explicitly stress cross-platform transferability \cite{wozniak2023applying}. RoboSense contributes to this direction by providing standardized datasets and evaluation protocols that enable systematic study of robust 3D perception across heterogeneous robotic platforms.
\section{RoboSense Challenge 2025}
\label{sec:robosense2025}

\begin{table*}[t]
\centering
\caption{The challenge results (\textbf{Phase 1} and \textbf{Phase 2}) of \textbf{Track 1: Driving with Language} in the \textbf{2025 RoboSense Challenge}. All scores are reported as percentages ($\%$). \underline{MCQ} stands for \textit{Multiple-Choice Question}, and \underline{VQA} stands for \textit{Visual Question Answering}. The average score of each phase is computed as a weighted mean based on the number of questions per task type. The final score is obtained by combining Phase 1 and Phase 2 scores with weights of 0.2 and 0.8, respectively.}
\vspace{-0.2cm}
\resizebox{\textwidth}{!}{%
\begin{tabular}{l|ccccccc|cccccccc|c}
\toprule
 & \multicolumn{7}{c|}{\textbf{Phase 1}} & \multicolumn{8}{c|}{\textbf{Phase 2}} & \\ 
\cmidrule(lr){2-8} \cmidrule(lr){9-16}
\textbf{Team Name} & \rotatebox{90}{\textcolor{robo_blue}{$\circ$} Perception-MCQs} & \rotatebox{90}{\textcolor{robo_blue}{$\circ$} Perception-VQAs-Object} & \rotatebox{90}{\textcolor{robo_blue}{$\circ$} Perception-VQAs-Scene} & \rotatebox{90}{\textcolor{robo_blue}{$\circ$} Prediction} & \rotatebox{90}{\textcolor{robo_blue}{$\circ$} Planning-VQAs-Scene} & \rotatebox{90}{\textcolor{robo_blue}{$\circ$} Planning-VQAs-Object} & \rotatebox{90}{\textcolor{robo_blue}{$\bullet$} \textbf{Weighted Average}} & \rotatebox{90}{\textcolor{robo_blue}{$\circ$} Perception-MCQs} & \rotatebox{90}{\textcolor{robo_blue}{$\circ$} Perception-VQAs-Object} & \rotatebox{90}{\textcolor{robo_blue}{$\circ$} Perception-VQAs-Scene} & \rotatebox{90}{\textcolor{robo_blue}{$\circ$} Prediction} & \rotatebox{90}{\textcolor{robo_blue}{$\circ$} Planning-VQAs-Scene} & \rotatebox{90}{\textcolor{robo_blue}{$\circ$} Planning-VQAs-Object} & \rotatebox{90}{\textcolor{robo_blue}{$\circ$} Corruption} & \rotatebox{90}{\textcolor{robo_blue}{$\bullet$} \textbf{Weighted Average}} & \rotatebox{90}{\textcolor{robo_blue}{$\bullet$} \textbf{Final}} \\ 
\midrule\midrule
\rowcolor{robo_blue!50}\multicolumn{17}{l}{\textcolor{white}{\textbf{Winning Teams}}}
\\
\rowcolor{robo_blue!20} \textcolor{robo_blue}{\faTrophy~\textbf{TQL}} 
& $92.45$ & $60.43$ & $47.03$ & $\mathbf{68.77}$ & $52.87$ & $61.08$ & $63.41$ 
& $53.06$ & $\mathbf{57.72}$ & $50.17$ & $\mathbf{69.92}$ & $50.03$ & $49.27$ & $\mathbf{100.00}$ & $\mathbf{60.40}$ & $\mathbf{61.00}$ 
\\
\rowcolor{robo_blue!15} \faTrophy~UCAS-CSU 
& $77.36$ & $63.79$ & $\mathbf{76.17}$ & $62.84$ & $\mathbf{75.21}$ & $\mathbf{64.83}$ & $\mathbf{66.14}$ 
& $84.69$ & $50.11$ & $\mathbf{64.32}$ & $51.45$ & $\mathbf{64.19}$ & $53.04$ & $99.04$ & $56.75$ & $58.63$
\\
\rowcolor{robo_blue!10} \faTrophy~AutoRobots 
& $67.92$ & $53.98$ & $64.45$ & $50.38$ & $69.46$ & $60.75$ & $57.17$ 
& $63.27$ & $50.65$ & $60.11$ & $59.58$ & $59.16$ & $\mathbf{53.72}$ & $99.04$ & $58.35$ & $58.11$
\\
\rowcolor{robo_blue!8} \faTrophy~CVML & $66.04$ & $58.52$ & $68.20$ & $50.19$ & $72.54$ & $\mathbf{64.84}$ & $59.38$ & $58.16$ & $42.79$ & $63.92$ & $49.59$ & $61.41$ & $52.16$ & $90.38$ & $52.89$ & $54.19$
\\
\rowcolor{robo_blue!5} \faTrophy~UQMM & $33.96$ & $42.70$ & $35.47$ & $59.20$ & $53.53$ & $58.16$ & $53.33$ & $70.41$ & $41.46$ & $37.31$ & $65.51$ & $51.38$ & $48.47$ & $51.92$ & $54.07$ & $53.92$
\\
\midrule
\rowcolor{robo_red!10}\multicolumn{17}{l}{\textcolor{robo_red}{\textbf{Other Teams}}}
\\
The Second & $\mathbf{94.34}$ & $39.77$ & $44.84$ & $59.20$ & $40.42$ & $54.48$ & $53.06$ & $\mathbf{98.98}$ & $35.17$ & $39.04$ & $61.56$ & $47.44$ & $50.31$ & $79.81$ & $54.07$ & $53.87$
\\
ECNU & $64.15$ & $\mathbf{64.20}$ & $73.59$ & $34.48$ & $66.62$ & $50.43$ & $50.39$ & $84.69$ & $54.08$ & $56.08$ & $46.46$ & $60.54$ & $51.39$ & $92.31$ & $54.28$ & $53.50$
\\
onedrive & $86.79$ & $44.63$ & $49.06$ & $59.00$ & $63.41$ & $61.56$ & $58.25$ & $71.43$ & $39.68$ & $52.16$ & $52.26$ & $51.41$ & $48.07$ & $79.81$ & $50.86$ & $52.34$
\\
TJJT & $64.15$ & $33.42$ & $32.11$ & $41.57$ & $52.13$ & $44.91$ & $42.66$ & $92.86$ & $33.18$ & $40.91$ & $63.76$ & $51.85$ & $50.08$ & $76.92$ & $54.58$ & $52.19$
\\
AutoL & $73.58$ & $30.02$ & $58.44$ & $47.32$ & $71.83$ & $62.62$ & $52.53$ & $89.80$ & $21.91$ & $40.39$ & $57.96$ & $51.95$ & $49.96$ & $93.27$ & $51.11$ & $51.39$
\\
Collab. Robot Lab & $67.92$ & $47.21$ & $62.66$ & $34.29$ & $61.86$ & $56.39$ & $47.90$ & $68.37$ & $43.27$ & $40.13$ & $18.12$ & $57.51$ & $\mathbf{53.90}$ & $80.77$ & $40.98$ & $42.36$
\\
\midrule
\rowcolor{gray!10}\multicolumn{17}{l}{\textcolor{gray}{\textbf{Baseline}}}
\\
Qwen2.5-VL-7B & $75.50$ & $29.20$ & $22.20$ & $59.20$ & $29.60$ & $31.20$ & $42.50$ & $78.60$ & $21.70$ & $19.30$ & $61.60$ & $30.80$ & $31.30$ & $81.70$ & $43.70$ & $43.50$ \\ 
\bottomrule
\end{tabular}%
}
\label{tab:res-cognitive-phases}
\end{table*}

\subsection*{Overview}
The RoboSense 2025 Challenge, held in conjunction with the 38th IEEE/RSJ International Conference on Intelligent Robots and Systems (IROS 2025), is a large-scale benchmark initiative aimed at advancing \emph{robust} and \emph{adaptive} robot sensing.

\subsection*{Challenge Tracks}
The following five tracks are established in this competition:
\begin{itemize}
    \item Track 1: Driving with Language
    \item Track 2: Social Navigation
    \item Track 3: Sensor Placement
    \item Track 4: Cross-Modal Drone Navigation
    \item Track 5: Cross-Platform 3D Object Detection
\end{itemize}

\subsection*{Challenge Phases}
To support both community onboarding and final-system benchmarking, we organized the competition in two phases:
\begin{itemize}
    \item \textbf{Phase 1 (June--August):} Preliminary exploration, baseline reproduction, and early model development.
    \item \textbf{Phase 2 (August--September):} Final system design, ablations, and official leaderboard submissions on hidden test sets.
\end{itemize}

\subsection{Track 1}

\subsubsection{Task Overview}
Track~1 benchmarks the robustness of VLMs for autonomous driving under visual degradations. Given multi-view sensory observations, participants develop VLM-based systems that answer language queries spanning \emph{perception}, \emph{prediction}, and \emph{planning}. Compared to standard driving QA settings, this track explicitly stresses \emph{grounded reasoning} and \emph{reliability} when the visual inputs are corrupted (\eg, blur, occlusion, adverse weather), reflecting realistic failure modes of on-vehicle sensors.

\subsubsection{Key Challenges}
\begin{itemize}
    \item \emph{Robust grounding under corruption:} Preserve faithful visual grounding and reduce hallucination when the visual evidence is degraded.
    \item \emph{Multi-task consistency:} Support diverse query types (MCQ and open-ended VQA) across perception, prediction, and planning, while maintaining coherent reasoning.
    \item \emph{Practical deployment constraints:} Achieve robust performance with feasible compute budgets and stable inference behavior.
\end{itemize}

\subsubsection{Information}
\begin{itemize}
    \item \textbf{Baseline Model:} Qwen2.5-VL from DriveBench \cite{xie2025drivebench}.
    \item \textbf{Estimated Training Requirements:} $4$$\times$ NVIDIA A100 GPUs, with an approximate training cycle of $12$ hours.
    \item \textbf{Baseline Performance:} Perception-MCQs@Accuracy = $75.5$, Perception-VQA@LLMScore = $27.8$, Prediction-MCQ@Accuracy = $59.2$, Planning-VQA@LLMScore = $30.8$.
    \item \textbf{Evaluation Server:} CodaBench at \url{https://www.codabench.org/competitions/9285}.
\end{itemize}

\begin{table*}[t]
\centering
\caption{The challenge results (\textbf{Phase 1} and \textbf{Phase 2}) of \textbf{Track 2: Social Navigation} in the \textbf{2025 RoboSense Challenge}. All scores are reported as percentages ($\%$). The final ranking is obtained only depending on Phase 2.}
\vspace{-0.2cm}
\resizebox{\textwidth}{!}{%
\begin{tabular}{l|ccccc|ccccc}
\toprule
& \multicolumn{5}{c|}{\textbf{Phase 1}} & \multicolumn{5}{c}{\textbf{Phase 2}}
\\ 
\cmidrule(lr){2-6} \cmidrule(lr){7-11}
\textbf{Team Name} 
& {\textcolor{robo_blue}{$\circ$} SR ↑~} 
& {\textcolor{robo_blue}{$\circ$} SPL ↑~} 
& {\textcolor{robo_blue}{$\circ$} PSC ↑} 
& {\textcolor{robo_blue}{$\circ$} H-Coll ↓} 
& {\textcolor{robo_blue}{$\bullet$} \textbf{Total ↑}} 
& {\textcolor{robo_blue}{$\circ$} SR ↑~} 
& {\textcolor{robo_blue}{$\circ$} SPL ↑~} 
& {\textcolor{robo_blue}{$\circ$} PSC ↑} 
& {\textcolor{robo_blue}{$\circ$} H-Coll ↓} 
& {\textcolor{robo_blue}{$\bullet$} \textbf{Total ↑}} 
\\ 
\midrule\midrule
\rowcolor{robo_blue!50}\multicolumn{11}{l}{\textcolor{white}{\textbf{Winning Teams}}}
\\
\rowcolor{robo_blue!20} \textcolor{robo_blue}{\faTrophy~\textbf{Are Ivan}} & $\mathbf{60.1}$ & $\mathbf{54.3}$ & $90.1$ & $38.4$ & $\mathbf{67.3}$ & $\mathbf{66.0}$ & $59.8$ & $\mathbf{86.3}$ & $\mathbf{32.4}$ & $\mathbf{70.2}$ 
\\
\rowcolor{robo_blue!15} \faTrophy~Xiaomi EV-AD VLA & $59.9$ & $53.3$ & $90.0$ & $37.4$ & $67.0$ & $65.6$ & $59.6$ & $86.1$ & $33.0$ & $69.9$
\\
\rowcolor{robo_blue!10} \faTrophy~AutoRobot & $55.7$ & $50.7$ & $89.5$ & $42.0$ & $64.3$ & $64.8$ & $\mathbf{60.1}$ & $86.1$ & $34.2$ & $69.8$ 
\\
\rowcolor{robo_blue!8} \faTrophy~DUO & $57.4$ & $50.8$ & $89.6$ & $38.9$ & $65.1$ & $65.2$ & $58.6$ & $86.1$ & $32.6$ & $69.5$
\\
\rowcolor{robo_blue!5} \faTrophy~CityU-ASL & $19.0$ & $15.8$ & $\mathbf{92.1}$ & $\mathbf{32.4}$ & $40.0$ & $64.4$ & $59.5$ & $85.8$ & $32.8$ & $69.4$ 
\\
\midrule
\rowcolor{robo_red!10}\multicolumn{11}{l}{\textcolor{robo_red}{\textbf{Other Teams}}}
\\
SocialDog & $53.7$ & $49.3$ & $89.5$ & $42.3$ & $63.1$ & $63.2$ & $56.6$ & $86.1$ & $34.2$ & $68.1$
\\
CORE Lab & $55.5$ & $50.6$ & $89.4$ & $42.4$ & $64.2$ & $54.0$ & $50.0$ & $86.3$ & $39.2$ & $62.5$
\\
MW & $54.8$ & $50.2$ & $89.4$ & $40.9$ & $63.8$ & $62.2$ & $55.2$ & $86.9$ & $30.8$ & $67.5$
\\
GRAM & $53.2$ & $46.2$ & $89.5$ & $43.9$ & $62.0$ & $56.8$ & $50.6$ & $86.6$ & $38.0$ & $63.9$
\\
\midrule
\rowcolor{gray!10}\multicolumn{11}{l}{\textcolor{gray}{\textbf{Baseline}}}
\\
Falcon & $55.8$ & $51.3$ & $89.5$ & $41.6$ & $64.6$ & $54.0$ & $50.0$ & $86.3$ & $39.2$ & $62.5$ 
\\ 
\bottomrule
\end{tabular}%
}
\label{tab:res-track2}
\end{table*}

\subsubsection{Evaluation Metrics}
Track~1 evaluates both \emph{discrete correctness} and \emph{language-answer quality}. For multiple-choice questions (MCQs), we report \emph{Accuracy}. For open-ended VQA queries, we report an \emph{LLMScore}, which measures the semantic correctness of the generated answer under a standardized LLM-based rubric. The official ranking aggregates performance across the track's four components (Perception-MCQs, Perception-VQA, Prediction-MCQs, Planning-VQA), emphasizing balanced robustness rather than specialization to a single query type. More details can be found in \cite{xie2025drivebench}.

\subsubsection{Leaderboard}
Table~\ref{tab:res-cognitive-phases} summarizes the top submissions on the evaluation server. Overall, participants substantially improved robustness under corruptions, with leading teams achieving consistent gains across both MCQ and VQA components. We observe that top-performing methods typically combine stronger visual encoders or corruption-aware augmentation with more reliable prompting or answer-calibration strategies, improving both factual grounding and response stability under degraded inputs.

\subsection{Track 2}

\subsubsection{Task Overview}
Track~2 benchmarks socially compliant navigation in photo-realistic indoor environments populated by dynamic human agents. Participants develop RGBD-based navigation systems that generate trajectories balancing \emph{task efficiency} and \emph{social compliance}, \eg, respecting personal space and avoiding socially unacceptable maneuvers. This track is built upon the \emph{Social-HM3D} benchmark \cite{gong2025cognition}, a large-scale simulation environment containing $844$ diverse indoor scenes (homes, offices, retail spaces, and public areas) with $0$--$6$ human agents per scene. Human motion is goal-driven and collision-aware (via ORCA), producing realistic interaction patterns beyond static-obstacle navigation.

\begin{table*}[t]
\centering
\caption{The challenge results (\textbf{Phase 2}) of \textbf{Track 3: Cross-Sensor Placement 3D Object Detection} in the \textbf{2025 RoboSense Challenge}. The final ranking is obtained only depending on mAP scores.}
\vspace{-0.2cm}
\renewcommand\arraystretch{1.2}
\resizebox{\textwidth}{!}{%
\begin{tabular}{p{2.40cm}<{\raggedright}|p{2.01cm}<{\centering}p{2.01cm}<{\centering}p{2.01cm}<{\centering}p{2.01cm}<{\centering}p{2.01cm}<{\centering}p{2.01cm}<{\centering}p{2.01cm}<{\centering}}
\toprule
\textbf{Team Name} 
& {\textcolor{robo_blue}{$\bullet$} \textbf{mAP} $\uparrow$} 
& \makecell[c]{\textcolor{robo_blue}{$\circ$} mATE $\downarrow$ \\ (m) } 
& \makecell[c]{\textcolor{robo_blue}{$\circ$} mASE $\downarrow$ \\ (1-IoU)} 
& \makecell[c]{\textcolor{robo_blue}{$\circ$} mAOE $\downarrow$ \\ (rad) } 
& \makecell[c]{\textcolor{robo_blue}{$\circ$} mAVE $\downarrow$ \\ (m/s)} 
& \makecell[c]{\textcolor{robo_blue}{$\circ$} mAAE $\downarrow$ \\ (1-acc)} 
& {\textcolor{robo_blue}{$\circ$} NDS $\uparrow$}
\\ 
\midrule\midrule
\rowcolor{robo_blue!50}\multicolumn{8}{l}{\textcolor{white}{\textbf{Winning Teams}}}
\\
\rowcolor{robo_blue!20} \textcolor{robo_blue}{\faTrophy~\textbf{LRP}} & $\mathbf{0.784}$ & $\mathbf{0.083}$ & $\mathbf{0.102}$ & $1.329$ & $2.571$ & $0.167$ & $0.657$
\\
\rowcolor{robo_blue!15} \faTrophy~Point Loom & $0.765$ & $0.099$ & $0.110$ & $1.067$ & $\mathbf{0.578}$ & $0.165$ & $\mathbf{0.688}$
\\
\rowcolor{robo_blue!10} \faTrophy~Smartqiu & $0.743$ & $0.101$ & $0.105$ & $\mathbf{1.062}$ & $0.615$ & $\mathbf{0.057}$ & $0.683$
\\
\rowcolor{robo_blue!8} \faTrophy~DZT328 & $0.726$ & $0.117$ & $0.129$ & $1.163$ & $0.817$ & $0.060$ & $0.651$
\\
\rowcolor{robo_blue!5} \faTrophy~seu\_zwk & $0.724$ & $0.110$ & $0.147$ & $1.109$ & $0.748$ & $0.062$ & $0.655$ 
\\
\midrule
\rowcolor{robo_red!10}\multicolumn{8}{l}{\textcolor{robo_red}{\textbf{Other Teams}}}
\\
laplace & $0.721$ & $0.108$ & $0.135$ & $1.072$ & $0.732$ & $0.066$ & $0.656$
\\
castiel972 & $0.714$ & $0.102$ & $0.112$ & $1.087$ & $0.689$ & $0.163$ & $0.65$
\\
Sensor Seekers & $0.707$ & $0.108$ & $0.116$ & $1.108$ & $0.701$ & $0.062$ & $0.655$
\\
hiepnt & $0.612$ & $0.116$ & $0.121$ & $1.221$ & $2.435$ & $0.404$ & $0.542$
\\
\midrule
\rowcolor{gray!10}\multicolumn{8}{l}{\textcolor{gray}{\textbf{Baseline}}}
\\
BEVFusion-L & $0.605$ & $0.121$ & $0.123$ & $1.164$ & $2.252$ & $0.398$ & $0.538$
\\ 
\bottomrule
\end{tabular}%
}
\label{tab:res-track3}
\end{table*}

\subsubsection{Key Challenges}
\begin{itemize}
    \item \emph{Social norm compliance:} Maintain safe distances, avoid collisions, and exhibit socially acceptable behavior in shared spaces.
    \item \emph{Realistic benchmarking:} Navigate in large-scale photo-realistic scenes with dynamic, collision-aware humans.
    \item \emph{Egocentric perception:} Operate from first-person observations under partial observability and frequent occlusions.
\end{itemize}

\subsubsection{Information}
\begin{itemize}
    \item \textbf{Baseline Model:} Falcon \cite{gong2025cognition}.
    \item \textbf{Estimated Training Requirements:} $4$$\times$ NVIDIA GeForce RTX 3090 GPUs; typical training completes within $1$--$2$ days.
    \item \textbf{Baseline Performance:} SR = $54.00\%$ on the evaluation set.
    \item \textbf{Evaluation Server:} EvalAI at \url{https://eval.ai/web/challenges/challenge-page/2557}.
\end{itemize}

\subsubsection{Evaluation Metrics}
The benchmark evaluates navigation quality across \textit{task completion} and \textit{social compliance}. \emph{Success Rate (SR)} measures the percentage of episodes where the agent reaches within 0.2\,m of the goal. \emph{Success weighted by Path Length (SPL)} captures path efficiency as
$\text{SPL} = \frac{1}{N} \sum_{i=1}^{N} S_i \frac{l_i}{\max(p_i, l_i)}$,
where $l_i$ is the shortest-path length, $p_i$ is the executed path length, and $S_i \in \{0,1\}$ indicates success. \emph{Personal Space Compliance (PSC)} measures the percentage of timesteps maintaining at least 0.5\,m distance from all humans. \emph{Human Collision Rate (H-Coll)} tracks episodes involving collisions with humans. The final ranking uses:
\[
\text{Total} = 0.4 \cdot \text{SR} + 0.3 \cdot \text{SPL} + 0.3 \cdot \text{PSC}.
\]

\subsubsection{Leaderboard.}
Table~\ref{tab:res-track2} reports the top submissions on the Phase~2 evaluation server.
The champion team \emph{Are Ivan} achieved the best score with \emph{Total = $70.22$}.
The second and third-place teams were \emph{Xiaomi EV-AD VLA} and \emph{AutoRobot}, with $69.94$ and $69.77$, respectively.
All results were officially verified on the EvalAI platform.

\subsection{Track 3}

\subsubsection{Task Overview}
Track~3 studies \emph{cross-sensor placement generalization} for LiDAR-based 3D object detection systems. In practical deployments, sensor configurations vary across vehicles and fleets due to differences in mounting positions, orientation, calibration, or hardware. Such configuration shifts often cause substantial degradation for detectors trained under a fixed setup. This track evaluates whether a model trained on one (or a subset of) sensor placements can generalize to \emph{unseen} placements with minimal performance drop, encouraging sensor-agnostic representations and robust geometric reasoning.

\begin{table*}[t]
\centering
\caption{The official leaderboard results of \textbf{Track 4: Text-to-Image Retrieval (Phase 2)} in the \textbf{roboText-190 Challenge}. 
All scores are given in percentage ($\%$). 
The final score is computed as $(\text{Recall@1} + \text{Recall@10}) / 2$. 
Higher scores indicate better retrieval performance. 
The best scores in each column are highlighted in \textbf{bold}.}
\vspace{-0.2cm}
\resizebox{\textwidth}{!}{%
\begin{tabular}{l|ccccl}
\toprule
\textbf{Team Name} & \textbf{Recall@1 (\%)} & \textbf{Recall@10 (\%)} & \textbf{Final Score (\%)} & \textbf{Award} & \textbf{Date} \\
\midrule\midrule
\rowcolor{robo_blue!50}\multicolumn{6}{l}{\textcolor{white}{\textbf{Winning Teams}}}
\\
\rowcolor{robo_blue!18} \textcolor{robo_blue}{\faTrophy~\textbf{lineng}} & $\mathbf{38.31}$ & $61.32$ & $\mathbf{49.82}$ & \textbf{Champion} & 2025-09-21
\\
\rowcolor{robo_blue!10} \faTrophy~rhao\_hur & $28.34$ & $\mathbf{66.11}$ & $47.23$ & \textbf{Runner-Up} & 2025-09-01
\\
\rowcolor{robo_blue!5} \faTrophy~Xiaomi EV-AD VLA & $31.33$ & $57.15$ & $44.24$ & \textbf{Third Place} & 2025-09-23
\\
\midrule
\rowcolor{robo_red!10}\multicolumn{6}{l}{\textcolor{robo_red}{\textbf{Other Top Teams}}}
\\
HiTSz-iLearn & $28.21$ & $52.37$ & $40.29$ & -- & 2025-09-19
\\
RED & $26.20$ & $49.65$ & $37.93$ & -- & 2025-09-19
\\
RoboSense2025 & $25.44$ & $49.10$ & $37.27$ & -- & 2025-08-31
\\
testliu & $23.14$ & $41.70$ & $32.42$ & -- & 2025-09-20
\\
robotes & $22.54$ & $41.21$ & $31.88$ & -- & 2025-09-20
\\
\midrule
\rowcolor{gray!10}\multicolumn{6}{l}{\textcolor{gray}{\textbf{Baseline}}}
\\
X-VLM (official baseline) &$25.44$ & $49.10$ & $37.27$  & Reference &   --
\\
\bottomrule
\end{tabular}%
}
\label{tab:res-track4}
\end{table*}

\subsubsection{Key Challenges}
\begin{itemize}
    \item \emph{Placement-induced distribution shift:} Changes in mounting position alter range-view statistics, sparsity patterns, and effective field-of-view.
    \item \emph{Occlusion and coverage variation:} Different placements expose different blind spots and visibility patterns, impacting detection recall.
    \item \emph{Calibration sensitivity:} Small extrinsic discrepancies can lead to systematic geometric bias in 3D localization.
    \item \emph{Generalization without retraining:} Methods should remain robust across placements without requiring per-configuration finetuning.
\end{itemize}

\subsubsection{Information}
\begin{itemize}
    \item \textbf{Baseline Model:} BEVFusion-L \cite{liu2023bevfusion}.
    \item \textbf{Estimated Training Requirements:} 1$\times$ NVIDIA RTX 4090 ($24$\,GB) for $\sim$$30$ hours.
    \item \textbf{Baseline Performance:} mAP = $0.605$ and NDS = $0.538$.
    \item \textbf{Evaluation Server:} CodaBench at \url{https://www.codabench.org/competitions/9284/}. Data is hosted at \url{https://huggingface.co/datasets/robosense/datasets/tree/main/track3-sensor-placement}.
\end{itemize}

\subsubsection{Evaluation Metrics}
We adopt standard nuScenes-style 3D detection metrics. The primary ranking metric is \emph{mAP}. We additionally report error-based metrics including
\emph{mATE}, \emph{mASE}, \emph{mAOE}, \emph{mAVE}, and \emph{mAAE}, as well as the composite \textbf{NDS} score:
\begin{equation*}
\begin{aligned}
\text{NDS} &=
\frac{1}{10}
\left(
5 \cdot \text{mAP}
+ \sum_{m }
\big( 1 - \min \left(1,m\right) \big)
\right), \\
m &\in \{\text{mATE}, \text{mASE}, \text{mAOE}, \text{mAVE}, \text{mAAE}\}.
\end{aligned}
\end{equation*}
The leaderboard is ranked by mAP. If the mAP difference is within $0.01$, NDS is used as a secondary tie-breaker.

\subsubsection{Leaderboard}
As shown in Table~\ref{tab:res-track3}, the baseline BEVFusion-L achieves mAP = $0.605$, while the top submission by Team \emph{LRP} reaches $0.784$, corresponding to an $18\%$ relative improvement. Beyond higher peak accuracy, many top solutions show improved stability under placement shifts, suggesting the value of placement-aware augmentation, geometry-consistent feature learning, and robust training schedules.

\subsection{Track 4}

\subsubsection{Task Overview}
Track~4 benchmarks language-guided cross-view retrieval for drone navigation and geolocalization. It is built upon the extended \emph{GeoText-190} dataset derived from \emph{GeoText-1652} \cite{chu2024geotext-1652}, containing $33{,}516$ text queries and $11{,}172$ aerial images from $190$ unseen categories spanning urban, suburban, and coastal environments. Each aerial image is paired with fine-grained spatially grounded captions following an \emph{image--text--bbox} structure. Phase~2 emphasizes \emph{cross-scene generalization}: models must retrieve the correct aerial image from a gallery purely based on a free-form natural language description, with no overlap between the hidden test set and training/Phase~1 categories.

\begin{table*}[t]
\centering
\caption{The official leaderboard results of \textbf{Track 5: Cross-Platform 3D Object Detection} in the \textbf{RoboSense Challenge 2025}. 
All scores are given in percentage ($\%$). 
The final score is reported as the overall mean Average Precision (mAP) across evaluation classes. 
Higher scores indicate better cross-platform detection performance. 
The best scores in each column are highlighted in \textbf{bold}.}
\vspace{-0.2cm}
\resizebox{\textwidth}{!}{%
\begin{tabular}{l|ccccccl}
\toprule
\textbf{Team Name} & \textbf{mAP (\%)} & \textbf{Car AP@0.50} & \textbf{Car AP@0.70} & \textbf{Ped. AP@0.25} & \textbf{Ped. AP@0.50} & \textbf{Award} & \textbf{Date} \\
\midrule\midrule
\rowcolor{robo_blue!50}\multicolumn{8}{l}{\textcolor{white}{\textbf{Winning Teams}}}
\\
\rowcolor{robo_blue!18} \textcolor{robo_blue}{\faTrophy~\textbf{youngseok\_kim}} 
& $\mathbf{58.54}$ & $\mathbf{64.17}$ & $25.36$ & $\mathbf{64.39}$ & $\mathbf{52.92}$ & \textbf{Champion} & 2025-09-16
\\
\rowcolor{robo_blue!10} \faTrophy~castiel972 
& $55.66$ & $56.30$ & $\mathbf{29.27}$ & $60.63$ & $55.02$ & \textbf{Runner-Up} & 2025-09-16
\\
\rowcolor{robo_blue!5} \faTrophy~linyongchun 
& $55.61$ & $56.14$ & $26.44$ & $60.60$ & $55.07$ & \textbf{Third Place} & 2025-09-16
\\
\midrule
\rowcolor{robo_red!10}\multicolumn{8}{l}{\textcolor{robo_red}{\textbf{Other Top Teams}}}
\\
DUTLu\_group & $54.29$ & $58.76$ & $30.89$ & $55.27$ & $49.81$ & -- & 2025-09-16
\\
arkn1ghts & $43.95$ & $36.59$ & $17.40$ & $60.21$ & $51.31$ & -- & 2025-09-11
\\
ljjincheng & $37.58$ & $31.82$ & $12.49$ & $49.49$ & $43.35$ & -- & 2025-09-16
\\
lvgroup & $36.12$ & $28.78$ & $11.66$ & $49.05$ & $43.45$ & -- & 2025-09-15
\\
\midrule
\rowcolor{gray!10}\multicolumn{8}{l}{\textcolor{gray}{\textbf{Baseline}}}
\\
PV-RCNN + ST3D++ \cite{liang2025pi3det} & $45.07$ & $52.65$ & $25.37$ & $46.99$ & $41.49$ & Reference & --
\\
\bottomrule
\end{tabular}%
}
\label{tab:res-track5}
\end{table*}

\subsubsection{Key Challenges}
\begin{itemize}
    \item \emph{Extreme viewpoint shift:} Align ground-level semantics with aerial imagery under drastic perspective and scale changes.
    \item \emph{Fine-grained spatial grounding:} Resolve relational cues (\eg, ``next to'', ``behind'', ``at the corner of'') that require spatial reasoning beyond global semantics.
    \item \emph{Open-world generalization:} Maintain retrieval quality on unseen categories and novel geographic contexts in Phase~2.
    \item \emph{Robust representation learning:} Handle appearance variation caused by lighting, seasonal changes, and occlusions in aerial views.
\end{itemize}

\subsubsection{Information}
\begin{itemize}
    \item \textbf{Baseline Model:} A standard dual-encoder retrieval baseline (image encoder + text encoder) is provided with training and evaluation scripts in the Track~4 repository.
    \item \textbf{Estimated Training Requirements:} The baseline can be trained on $1$--$2$ modern GPUs (\eg, RTX 3090/A100) within one day, depending on encoder choice and batch size.
    \item \textbf{Evaluation Server:} CodaBench at \url{https://www.codabench.org/competitions/10311/}.
\end{itemize}

\subsubsection{Evaluation Metrics}
The official Phase~2 score is the average of Recall@1 and Recall@10:
\[
\text{Final Score} = \frac{\text{Recall@1} + \text{Recall@10}}{2}.
\]
Rankings are determined solely by this score on the hidden test subset covering $190$ unseen aerial categories.

\subsubsection{Leaderboard}
Table~\ref{tab:res-track4} reports the top submissions on the Phase~2 evaluation server. The champion team \emph{TeleAI} achieved Recall@1 = $38.31\%$, Recall@10 = $61.32\%$, and a final score of $49.82$. The second and third-place teams were \emph{rhao\_hur} and \emph{Xiaomi EV-AD VLA}, with $47.23$ and $44.24$, respectively.
All results were officially verified on the CodaBench platform.

\subsection{Track 5}

\subsubsection{Task Overview}
Track~5 benchmarks \emph{cross-platform adaptation} for LiDAR-based 3D object detection across heterogeneous robotic platforms. While most prior detection benchmarks assume vehicle-mounted LiDAR with relatively consistent viewpoints and motion patterns, real deployments increasingly include drones and quadruped robots with substantially different sensing geometry and dynamics. Track~5 evaluates whether detectors can transfer from a labeled \emph{vehicle} source domain to unlabeled \emph{drone} (Phase~1) and \emph{quadruped} (Phase~2) target domains, encouraging domain-adaptive learning under strong platform-induced distribution shifts.

\subsubsection{Key Challenges}
\begin{itemize}
    \item \emph{Platform-induced geometric shift:} Viewpoint, height, and motion dynamics differ substantially across vehicle, drone, and quadruped platforms.
    \item \emph{Sensor diversity:} LiDAR sampling patterns and sparsity characteristics vary, changing object appearance and local geometry.
    \item \emph{Label scarcity on targets:} Drone and quadruped domains are unlabeled, requiring reliable pseudo-labeling and adaptation strategies.
    \item \emph{Stability under adaptation:} Methods should improve target performance without collapsing under noisy pseudo labels.
\end{itemize}

\subsubsection{Information}
\begin{itemize}
    \item \textbf{Baseline Model:} PV-RCNN \cite{shishaoshuai2020pv}, with ST3D++ as the baseline adaptation framework. Code and checkpoints are provided in \url{https://github.com/robosense2025/track5}.
    \item \textbf{Estimated Training Requirements:} $\sim$2$\times$ NVIDIA A100 ($40$\,GB) for $\sim$$24$ hours under the standard ST3D++ training protocol.
    \item \textbf{Baseline Performance:} On \textbf{Drone} (Phase~1), $45.07\%$ 3D AP@0.5 (R40) for Car; on \textbf{Quadruped} (Phase~2), $28.53\%$ (Car) and $41.49\%$ (Pedestrian) 3D AP@0.5 (R40).
    \item \textbf{Evaluation Server:} CodaBench at \url{https://www.codabench.org/competitions/9179/}. Dataset access is provided via \url{https://huggingface.co/datasets/robosense/datasets/tree/main/track5-cross-platform-3d-object-detection}.
\end{itemize}

\subsubsection{Evaluation Metrics}
Track~5 follows standard 3D detection evaluation with \textbf{3D AP@0.5 (R40)} computed per class on the target domains. For Phase~1 (Drone), ranking is determined by the Car class AP, while Phase~2 (Quadruped) considers both Car and Pedestrian. The final leaderboard aggregates Phase~1 and Phase~2 performance to reflect end-to-end cross-platform transfer ability across both target platforms. Higher AP indicates more accurate localization and classification under platform shifts. More details can be found in \cite{liang2025pi3det}.

\subsubsection{Leaderboard}
Track~5 submissions show clear progress beyond the ST3D++ baseline under strong platform discrepancies. Table~\ref{tab:res-track5} reports the top submissions on the Phase~2 evaluation server. We observe that leading methods consistently improve target-domain AP via stronger pseudo-label filtering, platform-aware augmentation, and more stable adaptation schedules, suggesting that robustness in cross-platform perception hinges on both representation transferability and adaptation stability.
\section{Challenge Solutions}
\label{sec:solution}
In this section, we summarize the key components of each of the solutions that contributed to this challenge.

\subsection{Track 1: Driving with Language}
This section introduces the key innovations and implementation details of the five winning solutions in Track 1.
\subsubsection{Team~ [\textcolor{robo_blue}{TQL}]}
\begin{figure*}[t]
    \centering
    \includegraphics[width=1\linewidth]{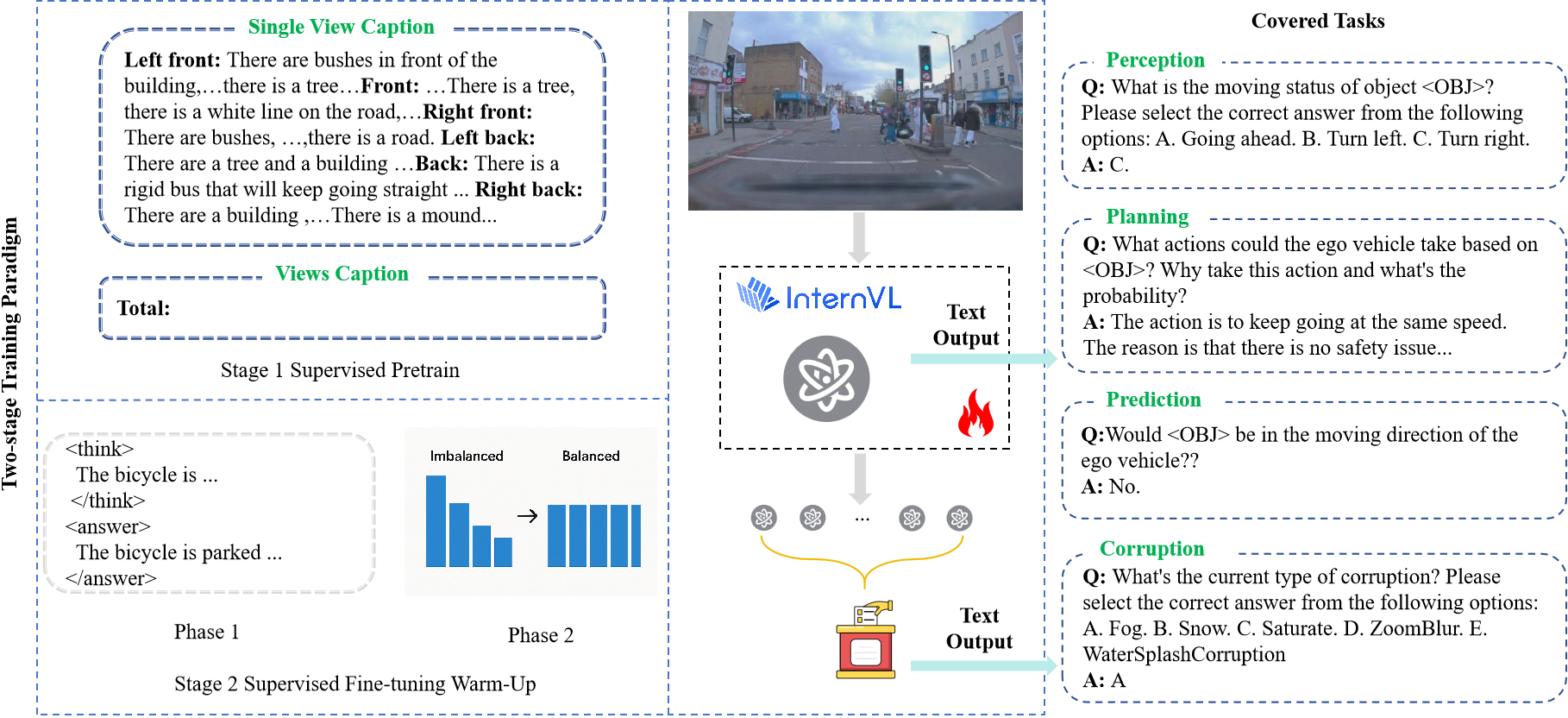}
    \vspace{-0.6cm}
    \caption{Team~[\textcolor{robo_blue}{TQL}]'s dataset expansion and training pipeline. Phase 1 uses pseudo-labels from InternVL3-8B for pre-training, six-view concatenation for multi-view learning, and CoT data from InternVL3-14B-Instruct for reasoning enhancement. Phase 2 applies balanced multi-view SFT with category-wise voting ensemble for final predictions.}
    \label{fig:track1_TQL_data_expand}
\end{figure*}

This team introduced a two-stage optimization framework designed to enhance the robustness and generalization capability of multi-view VLMs for driving. Motivated by the limitations of existing models, such as (1) vulnerability to visual degradations, (2) insufficient coordination across perception, prediction, and planning tasks, and (3) performance bias under imbalanced data distributions, the proposed approach strengthens both representation learning and reasoning stability. As shown in Fig.~\ref{fig:track1_TQL_data_expand}, the model builds upon the InternVL3-8B \cite{zhu2025internvl3} backbone and integrates pseudo-label pretraining, long-tail rebalancing, and multi-model ensembling to achieve reliable multi-view understanding under corrupted visual conditions.

\noindent\textbf{\faLightbulbO~Key Innovations:}
\begin{itemize}
    \item \textit{Pseudo-label pretraining with chain-of-thought (CoT):} Large-scale pseudo-annotations were generated using InternVL3-14B \cite{zhu2025internvl3} to inject structured reasoning signals into the model, facilitating improved multi-task comprehension across perception, prediction, and planning.
    
    \item \textit{Multi-view fusion via spatially ordered image concatenation:} Six camera views (front, rear, and lateral) were concatenated following a fixed spatial sequence to provide panoramic situational awareness, ensuring comprehensive scene coverage and enhanced spatial reasoning.

    \item \textit{Long-tail rebalancing and corruption-aware fine-tuning:} A task-level reweighting strategy was adopted to equalize the distributions of perception, prediction, planning, and corruption tasks, while hybrid training on official and synthetic datasets further mitigated data imbalance and improved resilience.
\end{itemize}

\noindent\textbf{\faGear~Implementation Details:}\\
The framework employs InternVL3-8B \cite{zhu2025internvl3} as the base model, with the Vision Transformer~\cite{dosovitskiy2020image} backbone frozen and optimization applied to the MLP adapters and language modeling layers. Training was conducted in two stages: pseudo-label pretraining to enrich input comprehension, followed by instruction fine-tuning with balanced and augmented data for robustness enhancement. All experiments were executed on 8$\times$A100 GPUs using DeepSpeed ZeRO-2 \cite{deepseed}, with a learning rate of $2\times10^{-5}$ and batch size of 128. Ensemble inference with category-wise voting yielded competitive performance across all tasks, achieving good accuracy on corruption-related evaluations and surpassing the Qwen2.5-VL-7B \cite{bai2025qwen2} baseline in overall robustness.

\subsubsection{Team~ [\textcolor{robo_blue}{UCAS-CSU}]}
\begin{figure*}[t]
    \centering
    \includegraphics[width=\textwidth]{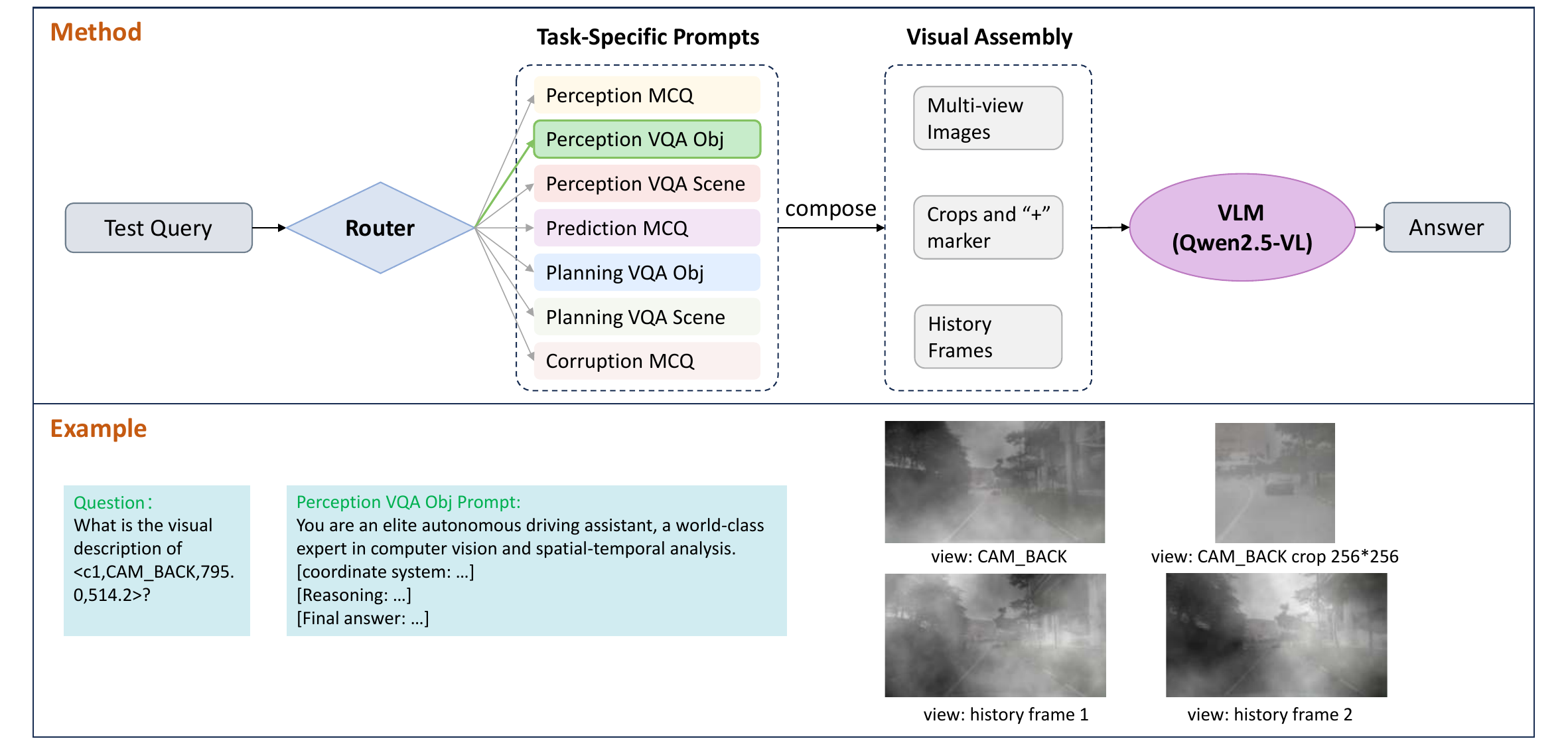}
    \vspace{-0.6cm}
    \caption{Team~[\textcolor{robo_blue}{UCAS-CSU}]'s mixture-of-prompts framework. A router classifies test queries and selects task-specific expert prompts, which are combined with multi-view images, visual markers, region crops, and adaptive historical frames before invoking the VLM.}
\label{fig:track1_UCAS_CSU_method}
\end{figure*}

The methodology pipeline is shown in Fig.~\ref{fig:track1_UCAS_CSU_method}. This team presented a structured prompting framework that substantially improves the reliability and spatial reasoning capability of VLMs for autonomous driving. Their approach is motivated by three critical challenges: (1) spatial misinterpretation in multi-view settings, (2) prompt interference across heterogeneous driving tasks, and (3) the need for adaptive temporal reasoning. To address these issues, the team designed a modular pipeline centered on systematic prompt engineering, spatial grounding, and dynamic visual composition, thereby enhancing the reasoning precision and robustness of Qwen2.5-VL-72B \cite{bai2025qwen2} across perception, prediction, planning, and corruption detection tasks.

\noindent\textbf{\faLightbulbO~Key Innovations:}
\begin{itemize}
    \item \textit{Mixture-of-prompts routing mechanism:} A lightweight router classifies each query and dispatches it to task-specific expert prompts, effectively eliminating prompt interference among diverse question types.
    
    \item \textit{Task-specific prompting with explicit spatial and temporal grounding:} Each expert prompt encodes camera coordinate systems, spatial rules (\eg, back-view constraints), role-playing instructions, and structured reasoning via Chain-of-Thought (CoT) \cite{wei2022chain} and Tree-of-Thought (ToT) \cite{yao2023tree} guidance, ensuring consistent and interpretable multi-view reasoning.
    
    \item \textit{Adaptive visual assembly for context-aware perception:} A modular visual composer integrates multi-view images, object crops with visual markers, and history frames selected by task type, enabling precise visual grounding and corruption-aware scene understanding.
\end{itemize}

\noindent\textbf{\faGear~Implementation Details:}\\
The framework is implemented on Qwen2.5-VL-72B-Instruct \cite{bai2025qwen2}, utilizing task-specific inference configurations to optimize decoding behavior. Deterministic tasks (\eg, MCQs) employ low temperature and top-p values for stable outputs, while descriptive tasks adopt higher sampling for richer contextual responses. Inference is parallelized across multiple workers with fault-tolerant checkpointing for efficient experimentation. Empirical results demonstrate significant gains in both clean and corrupted conditions, highlighting that explicit prompting, spatial reasoning, and adaptive context integration jointly improve the robustness and interpretability of driving-oriented VLMs.

\subsubsection{Team~ [\textcolor{robo_blue}{AutoRobots}]}
This team proposed a unified framework that integrates task-aware inference routing with CoT~\cite{wei2022chain} augmented fine-tuning to improve robustness and reasoning in multi-view autonomous driving scenarios. Their approach is motivated by the observation that generic VLMs often rely on linguistic priors rather than grounded visual cues, particularly under degraded inputs. To address this limitation, the framework combines a lightweight query analyzer for dynamic prompt routing and visual selection with supervised fine-tuning on rationalized data, enhancing both perception accuracy and temporal understanding across diverse driving tasks.

\noindent\textbf{\faLightbulbO~Key Innovations:}
\begin{itemize}
    \item \textit{Task-aware prompt routing:} A query analyzer automatically classifies each question and selects the most suitable prompt, controlling view inclusion and temporal policy. It disables history for short perception and corruption queries, while enabling strided temporal sampling for prediction and planning tasks.
    \item \textit{Chain-of-thought augmented fine-tuning:} The model is fine-tuned on the DriveLM dataset \cite{sima2024drivelm} with concise reasoning traces generated by ChatGLM \cite{glm2024chatglm}, injecting structured stepwise logic that strengthens semantic alignment and interpretability across perception, prediction, and planning.
    \item \textit{Coupled training and inference design:} The same router metadata used during inference guides data alignment during training, ensuring consistent temporal context exposure and stable reasoning behavior under multi-view and corrupted conditions.
\end{itemize}

\noindent\textbf{\faGear~Implementation Details:}\\
The framework builds upon Qwen2.5-VL-72B \cite{bai2025qwen2} and employs a LoRA-based supervised instruction fine-tuning scheme with rank~8 and scaling factor~16 \cite{hu2022lora}, keeping the vision tower and multimodal projector frozen. Optimization uses AdamW \cite{loshchilov2017decoupled} with a learning rate of $2\times10^{-5}$, a cosine decay schedule, and 3~epochs of training. During inference, deterministic decoding (low temperature and top-p) is used for structured tasks, while open-ended queries adopt a more explorative sampling strategy. The method outperformed the official baseline by around 15 points and demonstrated strong robustness across perception, prediction, and planning under corrupted visual inputs.

\subsubsection{Team~ [\textcolor{robo_blue}{CVML}]}
\begin{figure*}
  \centering
  \begin{subfigure}{0.64\linewidth}
    \centering
    \includegraphics[width=0.99\linewidth]{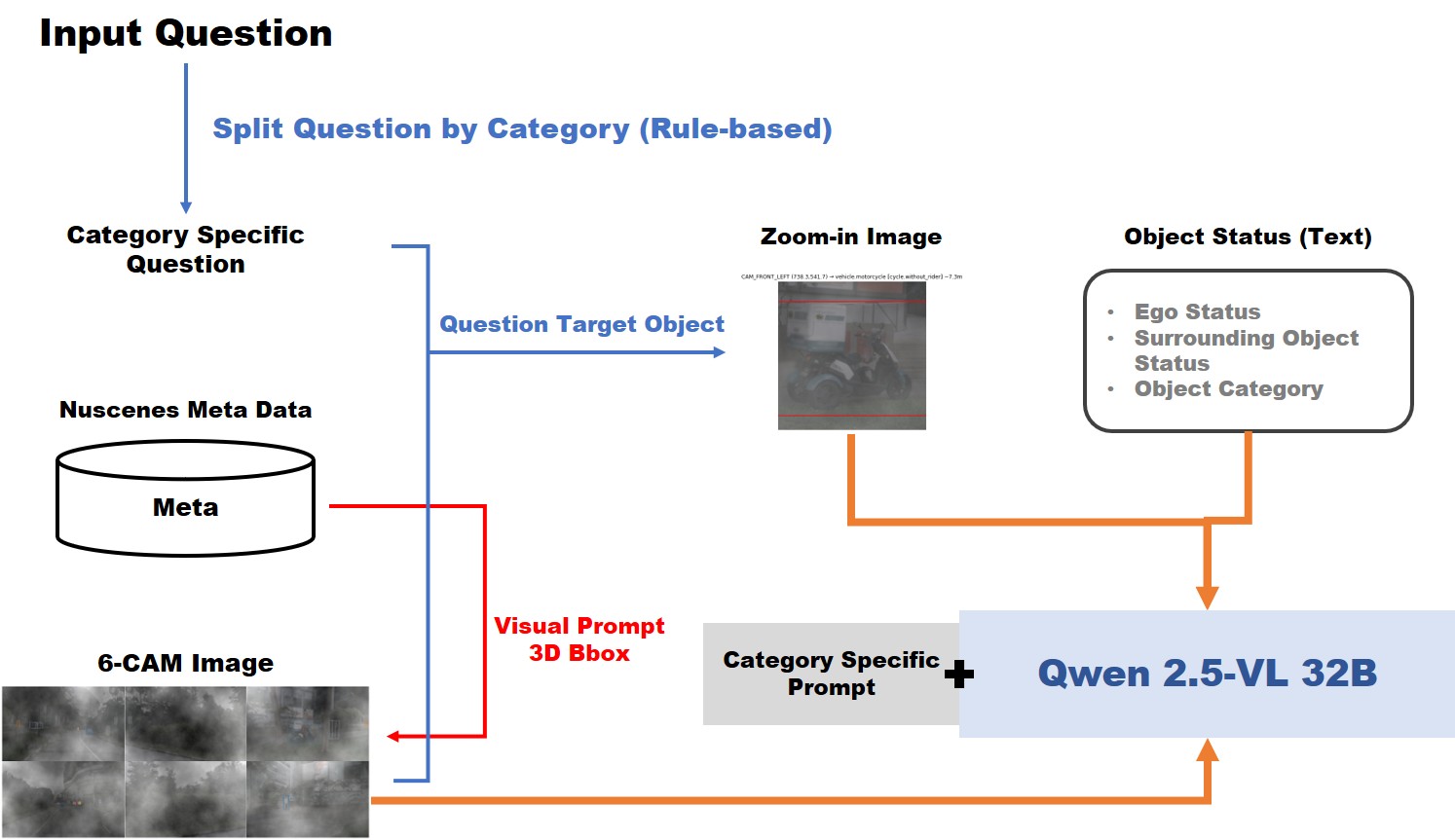}
    \caption{Phase-2 overview.}
  \end{subfigure}
  \hfill
  \begin{subfigure}{0.35\linewidth}
    \centering
    \includegraphics[width=0.9\linewidth]{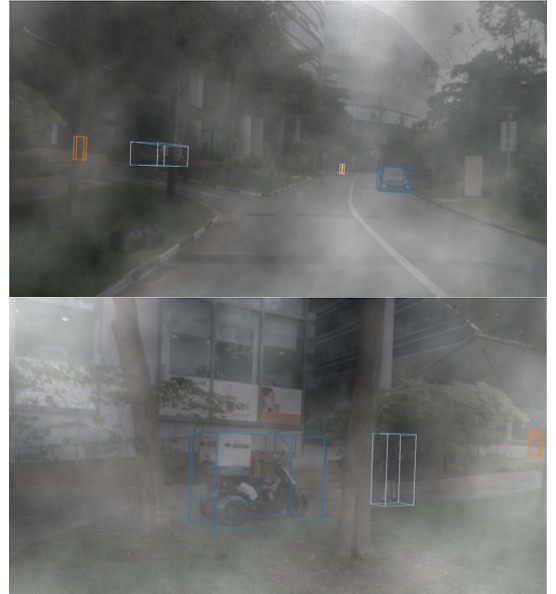}
    \caption{Example of nuScenes Images with 3D bounding box visual prompt.}
  \end{subfigure}
  \vspace{-0.2cm}
  \caption{Team~[\textcolor{robo_blue}{CVML}]'s Phase-2 metadata-grounded reasoning framework. A rule-based router categorizes questions, then injects nuScenes metadata, zoom-in crops, and object/ego status into task-specific prompts for Qwen~2.5-VL~32B.}
  \label{fig:track1_CVML_method}
\end{figure*}

As shown in Fig.~\ref{fig:track1_CVML_method}, this team developed a vision–language reasoning framework that enhances the robustness and interpretability of driving systems through metadata grounding and task-specific prompting. Their design is motivated by the observation that VLMs often exhibit superficial reasoning, relying excessively on linguistic priors when visual cues are incomplete or corrupted. To address this limitation, the proposed approach integrates explicit scene metadata, geometric priors, and structured prompting, thereby reinforcing spatial awareness and logical consistency in driving-oriented question answering. The resulting system demonstrates improved resilience across diverse corruptions while maintaining coherent reasoning ability.

\noindent\textbf{\faLightbulbO~Key Innovations:}
\begin{itemize}
    \item \textit{Metadata-grounded context integration:} The framework injects structured \textit{nuScenes} metadata, covering object categories, bounding box coordinates, ego-vehicle pose, and HD map priors, into the model’s prompt template. This explicit grounding allows the model to reason about spatial relations and temporal dependencies that are difficult to infer solely from visual tokens.
    \item \textit{Task-specific structured prompting:} Distinct prompt templates are designed for perception, prediction, and planning questions. Each template follows a hierarchical reasoning structure (\textit{Observation} $\rightarrow$ \textit{Reasoning} $\rightarrow$ \textit{Answer}) and includes explicit domain knowledge such as traffic rule adherence and motion constraints, improving both factual correctness and interpretability.
    \item \textit{Geometric and visual priors for corruption robustness:} The team incorporates geometric cues such as vanishing-point alignment, edge-gradient descriptors, and spatial heatmaps as auxiliary prompts or low-level features. These cues help stabilize model predictions under visual degradation (\eg, motion blur, fog, and occlusion) by providing additional geometric orientation and structural awareness.
\end{itemize}

\noindent\textbf{\faGear~Implementation Details:}\\
The system is built upon Qwen2.5-VL-32B \cite{bai2025qwen2}, employing few-shot CoT~\cite{wei2022chain} prompting in Phase~1 and metadata-augmented zero-shot prompting in Phase~2. The inference pipeline adopts a self-consistency decoding strategy, aggregating multiple reasoning paths for reliability, while a rule-based verification layer filters logically inconsistent outputs to ensure safety-aligned decision making. The framework achieved a good overall accuracy improvement, with robustness exceeding 96\% under heavy visual corruptions. Ablation studies confirm that structured prompting and metadata grounding jointly contribute to performance gains, establishing a strong and interpretable baseline for vision–language reasoning in autonomous driving.

\begin{figure*}
  \centering
  \includegraphics[width=\textwidth]{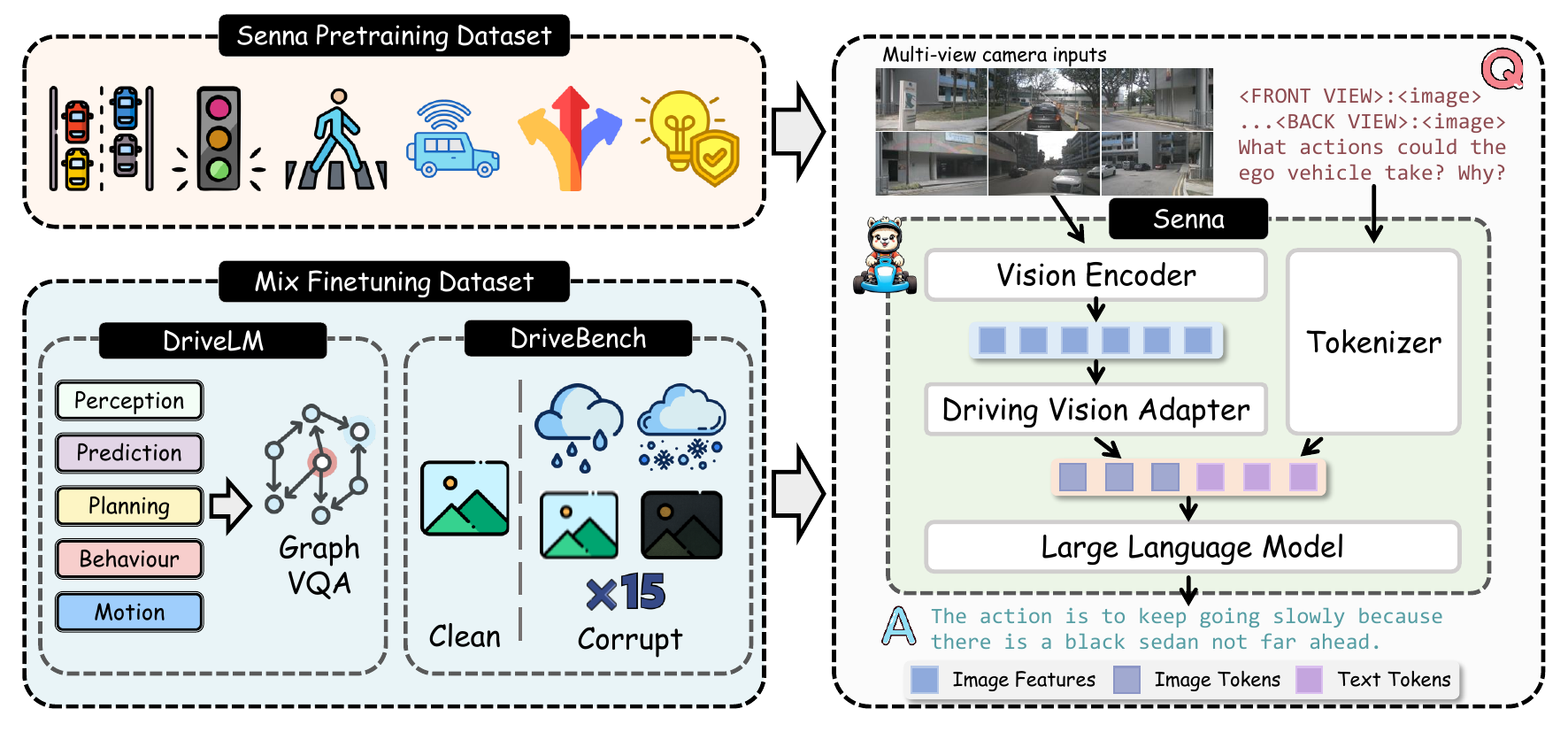}
  \vspace{-0.6cm}
    \caption{Team~[\textcolor{robo_blue}{UQMM}]'s fine-tuning pipeline for Senna-VLM \cite{jiang2024senna}. The model is pretrained on the Senna dataset, then fine-tuned using DriveLM \cite{sima2024drivelm} for graph-structured VQA and DriveBench for corruption-aware reasoning.}
  \label{fig:track1_UQMM_overview}
\end{figure*}

\subsubsection{Team~ [\textcolor{robo_blue}{UQMM}]}
This team proposed a robust fine-tuning strategy for driving VLMs that enhances multi-view reasoning and resilience against visual corruptions in Fig.~\ref{fig:track1_UQMM_overview}. Their approach builds upon Senna-VLM \cite{jiang2024senna}, a domain-specific architecture for autonomous driving that integrates a Driving Vision Adapter (DVA) for efficient multi-camera compression and spatial reasoning. Motivated by the observation that generic driving VLMs often struggle with corrupted inputs and limited diversity in question-answer structures, the team developed a multi-source LoRA~\cite{hu2022lora} fine-tuning framework combining diverse datasets and corruption-aware supervision to strengthen model generalization across perception, prediction, and planning tasks.

\noindent\textbf{\faLightbulbO~Key Innovations:}
\begin{itemize}
    \item \textit{Multi-source LoRA fine-tuning:} The model was fine-tuned using a mixture of complementary datasets, including DriveLM~\cite{sima2024drivelm} for graph-structured reasoning and DriveBench~\cite{xie2025drivebench} for corruption-aware QA. This multi-source supervision improved both linguistic diversity and resilience under degraded visual conditions.
    
    \item \textit{Driving vision adapter (DVA) for multi-view integration:} A custom adapter compresses visual tokens from multiple camera views while preserving spatial context, mitigating token overload, and enabling the model to reason effectively across 360° surround-view inputs.
    
    \item \textit{Balanced corruption-aware training:} Fine-tuning strategies progressively combined clean and corrupted data, ensuring robustness to 15 types of synthetic degradations such as weather, blur, and sensor noise. This approach improved model stability and corruption identification accuracy without sacrificing clean-scene performance.
\end{itemize}

\begin{figure*}[t]
    \centering
    \includegraphics[width=\linewidth]{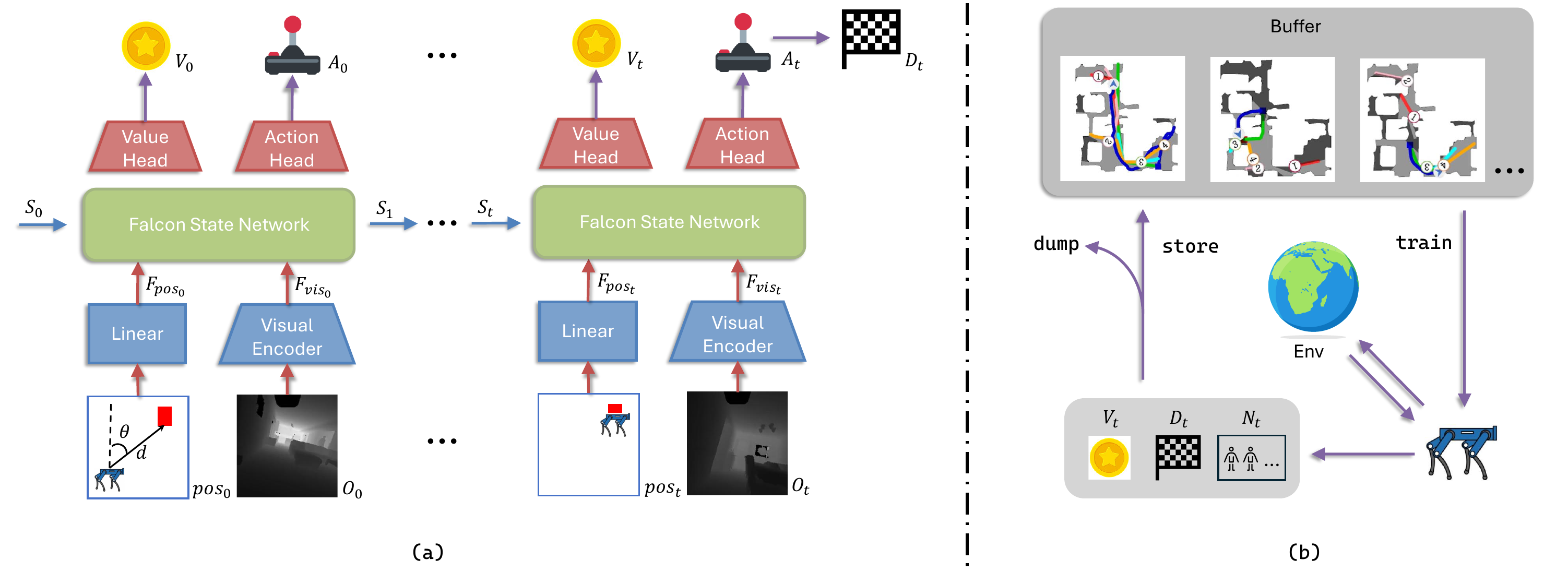}
    \vspace{-0.6cm}
    \caption{Team~[\textcolor{robo_blue}{Are Ivan}]'s PER-Falcon framework. (a) Recurrent policy architecture processing GPS and depth inputs. (b) Positive Episode Replay mechanism storing high-return trajectories for auxiliary DDPPO training.}
    \label{fig:track2_Are_Ivan_overview}
\end{figure*}

\noindent\textbf{\faGear~Implementation Details:}\\
The framework employs Senna-VLM \cite{jiang2024senna} initialized from official pretrained weights and optimized with parameter-efficient LoRA fine-tuning (rank~128, $\alpha$~=~256)~\cite{hu2022lora}. Training was performed on a single NVIDIA H100 GPU using AdamW~\cite{loshchilov2017decoupled} with a learning rate of $2\times10^{-5}$ and cosine decay scheduling. Multiple fine-tuning stages were explored, ranging from clean data to combined datasets with 15 corruption types and graph-based supervision. The best configuration (DriveBench + DriveLM) improved overall accuracy over the Qwen2.5-VL \cite{bai2025qwen2} baseline and attained strong robustness across perception, prediction, and planning tasks. These results demonstrate that integrating domain-specific priors with multi-source LoRA fine-tuning effectively enhances reasoning accuracy and corruption resilience for driving-oriented VLMs.

\subsubsection{Summary \& Discussion of Track 1}
This track benchmarks vision-language reasoning in autonomous driving, evaluating how VLMs perceive, reason, and plan under both clean and corrupted visual conditions. The top-performing teams collectively demonstrated several converging trends: the adoption of structured prompting and chain-of-thought reasoning, multi-view fusion for spatial completeness, and multi-source fine-tuning strategies for robustness. Most solutions relied on parameter-efficient tuning (\eg, LoRA or adapter-based methods) to scale large VLMs while maintaining interpretability.

Across the five winning teams, robustness under corruption emerged as the most critical differentiator. Teams that explicitly combined reasoning supervision with corruption-aware data achieved the highest overall stability. These findings underline the growing importance of language-driven reasoning and structured supervision for reliable autonomous driving. Looking forward, extending this paradigm to temporal reasoning, multimodal fusion (LiDAR–camera–language), and closed-loop decision-making offers a promising direction toward fully explainable embodied intelligence.

\subsection{Track 2: Social Navigation}
This section introduces the key innovations and implementation details of the three winning solutions in Track 2.
\subsubsection{Team~ [\textcolor{robo_blue}{Are Ivan}]}
This team introduced PER-Falcon, as shown in Fig. \ref{fig:track2_Are_Ivan_overview}. It is a Reinforcement Learning (RL) framework designed to improve socially compliant navigation in dynamic human environments. Building upon the Falcon \cite{gong2025cognition} baseline, PER-Falcon integrates a Positive Episode Replay (PER) mechanism that selectively reuses high-return trajectories during policy optimization, thereby accelerating the acquisition of complex and socially aware navigation behaviors. The method explicitly prioritizes valuable experiences, those that exhibit efficient maneuvers and safe human-robot interactions, to achieve more stable and sample-efficient learning in crowded indoor scenes.

\begin{figure*}[t]
    \centering
    \includegraphics[width=\linewidth]{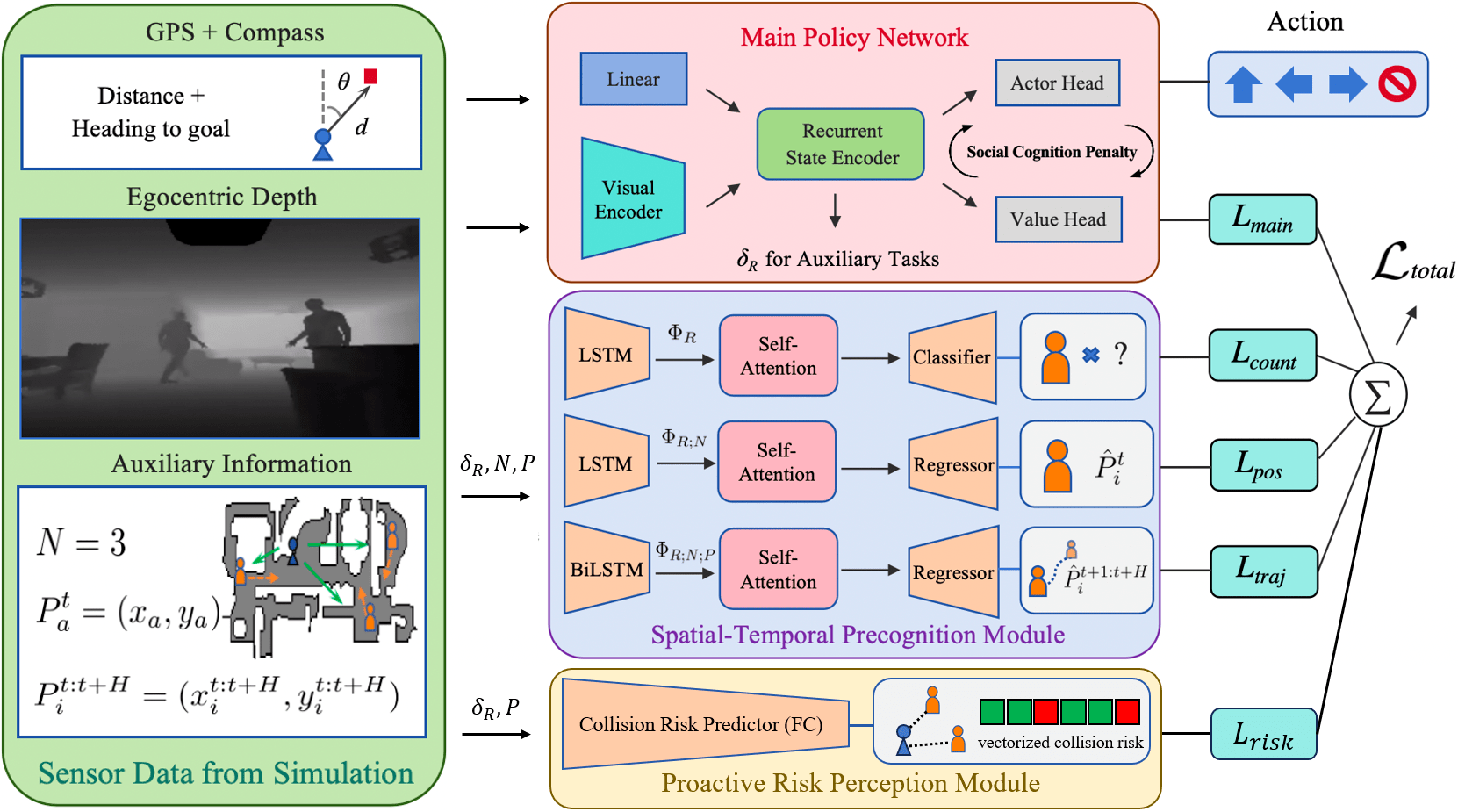}
    \caption{Team~[\textcolor{robo_blue}{Xiaomi EV-AD VLA}]'s Proactive Risk Perception framework. The risk perception module operates alongside Falcon's policy network, predicting distance-based collision risks for nearby humans to enhance spatial awareness and collision avoidance.}
    \label{fig:track2_XiaomiEV_overview}
\end{figure*}

\noindent\textbf{\faLightbulbO~Key Innovations:}
\begin{itemize}
    \item \textit{Positive episode replay:} Introduces a prioritized experience replay strategy that periodically identifies successful trajectories with undiscounted returns above a fixed threshold and reuses them for auxiliary DDPPO updates, ensuring greater learning focus on high-value experiences.
    \item \textit{Secondary batch construction and auxiliary DDPPO update:} Implements two dedicated modules to realize PER-one for extracting and storing positive trajectories, and another for performing auxiliary gradient updates on the secondary batch-allowing efficient reinforcement of social behaviors without disrupting DDPPO stability.
    \item \textit{Reward rebalancing and collision penalty tuning:} Slightly increases the human-collision penalty from $-0.015$ to $-0.017$ to better discourage unsafe maneuvers, achieving a balance between navigation efficiency and social compliance.
\end{itemize}

\noindent\textbf{\faGear~Implementation Details:}\\
The framework is trained on the Social-HM3D~\cite{gong2025cognition} dataset using a DDPPO-based recurrent policy\cite{wijmans2019dd}. The PER module is activated every 20 update steps, replaying trajectories with returns above~10, corresponding to the success-reward threshold. Training is conducted on 4$\times$V100 GPUs for approximately 40~hours using Habitat-sim’s distributed DDPPO implementation. Experimental results show consistent improvements across all metrics on the Phase-2 test set, achieving a success rate of~0.660, an SPL of~0.598, and a total score of~0.702-surpassing the official Falcon baseline by~7.8~percentage points. Ablation studies confirm that positive-episode selection, rather than negative replay, is key to performance gains, demonstrating the effectiveness of PER-Falcon in promoting future-aware, socially compliant navigation.

\begin{figure*}
    \centering
    \includegraphics[width=\linewidth]{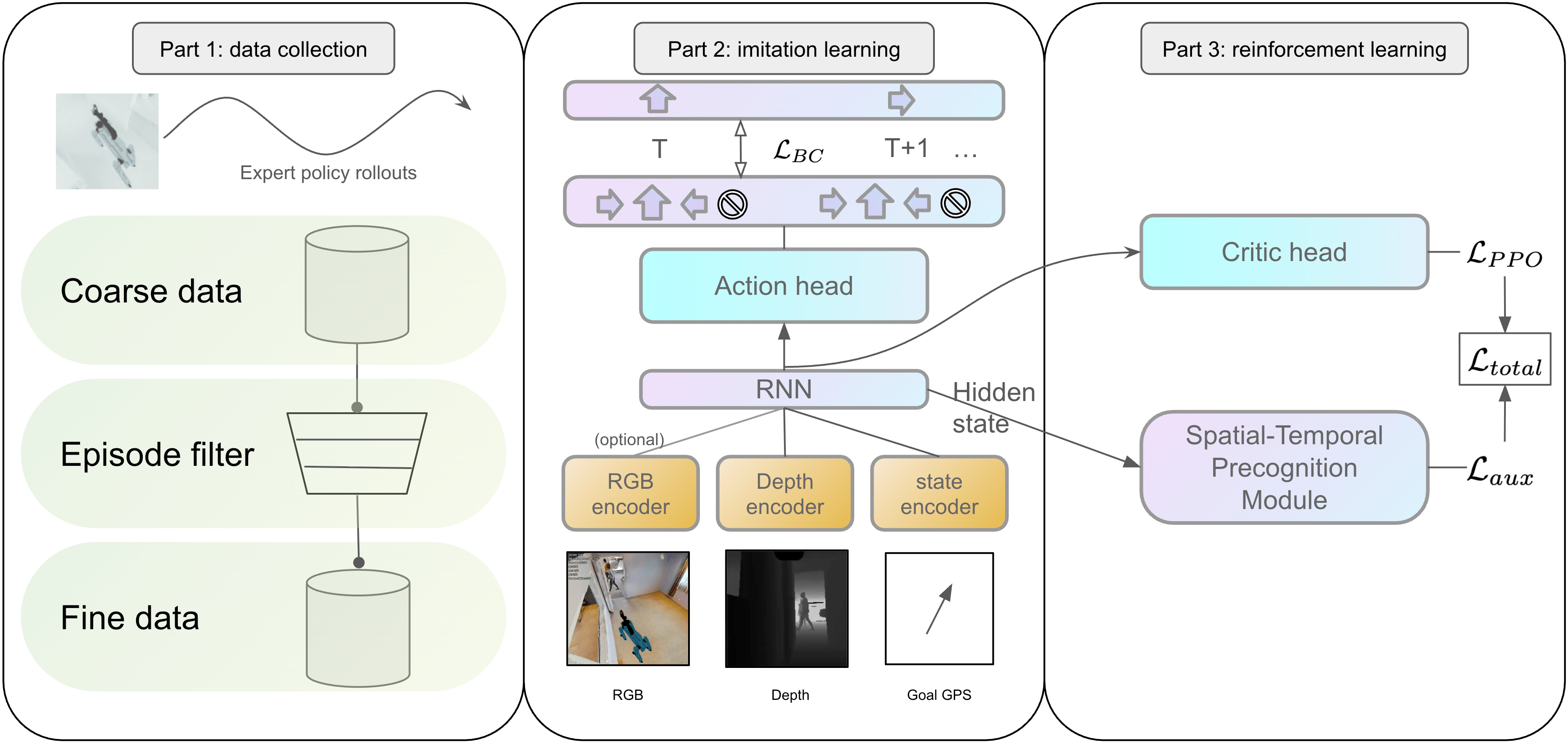}
    \vspace{-0.6cm}
    \caption{Team~[\textcolor{robo_blue}{AutoRobot}]'s two-stage IL-RL training pipeline. The framework progresses from (1) expert demonstration collection and filtering, through (2) Behavioral Cloning pre-training, to (3) DDPPO fine-tuning with spatial-temporal precognition modules.}
    \label{fig:track2_AutoRobot_arch}
\end{figure*}

\subsubsection{Team~ [\textcolor{robo_blue}{Xiaomi EV-AD VLA}]}
This team proposed a Proactive Risk Perception (PRP) framework to enhance social navigation safety and efficiency in dynamic environments, as shown in Fig. \ref{fig:track2_XiaomiEV_overview}. Built upon the Falcon~\cite{gong2025cognition} baseline, the approach introduces a neural risk-perception module that enables agents to anticipate potential collisions before they occur, improving both navigation stability and social norm compliance. By explicitly quantifying collision risk through distance-aware supervision, the model encourages proactive avoidance behaviors that balance goal efficiency with human-centric spatial comfort.

\noindent\textbf{\faLightbulbO~Key Innovations:}
\begin{itemize}
    \item \textit{Proactive risk perception module:} Extends the spatial-temporal reasoning of Falcon with a lightweight network predicting continuous risk scores for surrounding humans. The module generates dense supervisory signals guiding the agent to maintain safe interpersonal distances.
    
    \item \textit{Distance-weighted auxiliary loss:} Employs a distance-based loss emphasizing near-collision states, providing stronger gradients in high-risk zones and promoting earlier evasive actions.
    
    \item \textit{Integrated auxiliary learning and policy optimization:} Jointly trains the risk-perception branch with Falcon’s auxiliary tasks (population, position, trajectory estimation) under a unified DDPPO objective, achieving better synergy between situational awareness and decision reliability.
\end{itemize}

\noindent\textbf{\faGear~Implementation Details:}\\
The method is trained via DDPPO~\cite{wijmans2019dd} on the Social-HM3D~\cite{gong2025cognition} dataset. Depth images are encoded by ResNet-50, temporal dependencies are modeled with a two-layer LSTM, and the PRP branch adds two fully connected layers sharing the same hidden state as Falcon’s auxiliary tasks. Training is performed on 4$\times$A40 GPUs for 75 M steps. On the Phase-2 test set, the model achieved a success rate of 0.656, with a SPL of 0.596, a PSC of 0.861, and an overall score of 0.699. The narrow margin (0.0028) behind the winning entry validates the effectiveness of proactive risk perception in achieving socially compliant and collision-aware navigation.

\subsubsection{Team~ [\textcolor{robo_blue}{AutoRobot}]}
This team proposed a two-stage training paradigm, \textit{From Imitation to Interaction}, that unifies Imitation Learning (IL) and Reinforcement Learning (RL) for socially compliant navigation. The core motivation stems from the complementary strengths of both paradigms: IL provides a stable policy initialization from expert demonstrations, while RL refines the policy through interaction-based adaptation. The resulting hybrid framework, as shown in Fig. \ref{fig:track2_AutoRobot_arch}, enables agents to navigate efficiently and safely in dynamic human environments by progressively bootstrapping from demonstration-driven behavior to environment-aware interaction.

\noindent\textbf{\faLightbulbO~Key Innovations:}
\begin{itemize}
    \item \textit{Two-stage IL-RL training pipeline:} The framework first pre-trains a policy via Behavioral Cloning (BC) on high-quality expert trajectories to establish a robust foundation, then fine-tunes it using DDPPO with auxiliary social-awareness tasks.
    
    \item \textit{Spatial-temporal precognition module:} During RL fine-tuning, three auxiliary objectives -- human count estimation, current position tracking, and future trajectory forecasting -- enhance the agent’s predictive awareness of nearby humans, leading to smoother, safer navigation.
    
    \item \textit{Self-reinforcing data cycle:} The enhanced RL policy is reused to generate additional expert demonstrations, iteratively improving the IL dataset and creating a virtuous learning loop that strengthens both imitation and interaction performance over time.
\end{itemize}

\noindent\textbf{\faGear~Implementation Details:}\\
The agent’s policy network is based on the Falcon~\cite{gong2025cognition} architecture, composed of a ResNet-50 visual encoder, a two-layer GRU temporal module, and policy-value heads. IL pre-training uses cross-entropy loss between predicted and expert actions, trained for 20~epochs with Adam ($1\times10^{-5}$). The RL phase fine-tunes the IL policy via DDPPO with Generalized Advantage Estimation ($\gamma{=}0.99$, $\lambda{=}0.95$) for 6M environment steps. Training employs 8~parallel Habitat environments on 4$\times$V100 GPUs. Results on the Social-HM3D benchmark show substantial gains over both IL-only and RL-only baselines, achieving SR 0.648, SPL 0.601, and PSC 0.861 with a total score of 0.698, demonstrating that the IL+RL synergy yields safer, more socially aware navigation in dynamic settings.

\subsubsection{Team~ [\textcolor{robo_blue}{DUO}]}
\begin{figure}
  \centering
    \includegraphics[width=\linewidth]{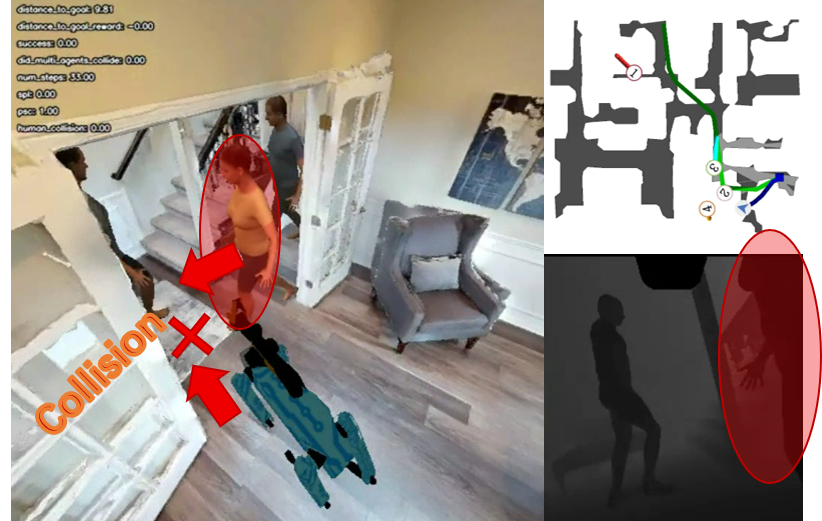}
    \vspace{-0.6cm}
    \caption{Team~[\textcolor{robo_blue}{DUO}]'s motivation for occlusion robustness. Example scenario where a pedestrian suddenly approaches from the side, obstructing the robot's view and causing a potential collision.}
    \label{fig:track2_DUO_collision}
\end{figure}

This team presented a lightweight yet effective enhancement to the Falcon~\cite{gong2025cognition} framework through image-based augmentation, as shown in Fig. \ref{fig:track2_DUO_collision}. It targets the problem of human occlusion in dynamic social navigation scenarios. Their method, termed \textbf{Cutout-Augmented Falcon}, integrates the classic Cutout augmentation technique into the RGB-D input stream to simulate partial visual occlusions caused by pedestrians or environmental obstacles. By forcing the model to reconstruct missing spatial cues from incomplete depth observations, the approach significantly strengthens robustness and situational awareness without altering the underlying model architecture.

\begin{figure*}
  \centering
    \includegraphics[width=\textwidth]{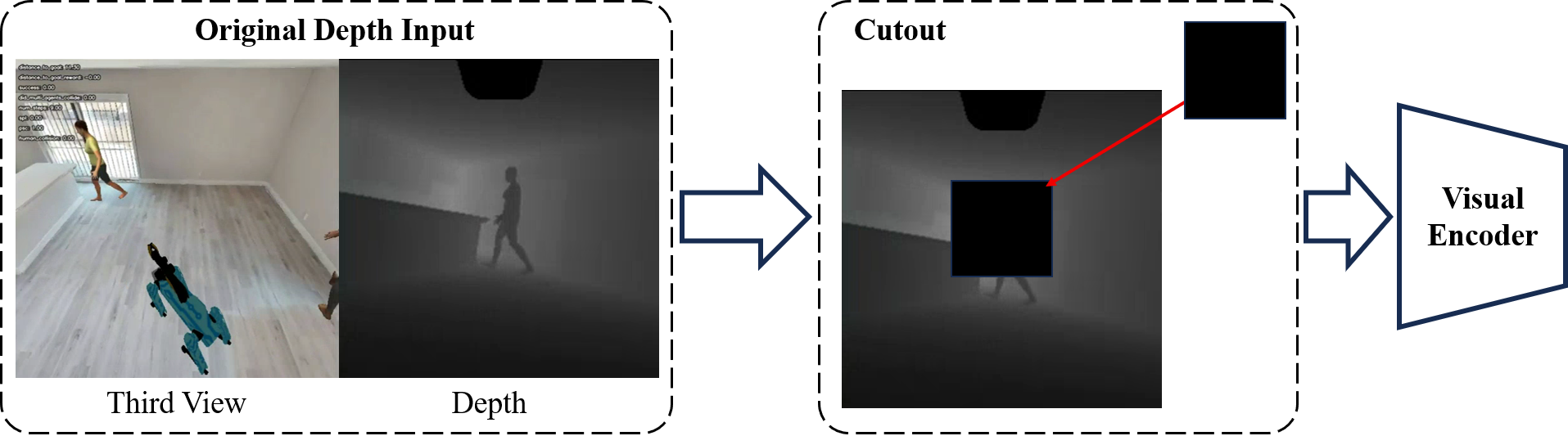}
    \vspace{-0.6cm}
    \caption{Team~[\textcolor{robo_blue}{DUO}]'s Cutout augmentation strategy. Random rectangular patches are applied to depth images before ResNet-50 encoding to simulate real-world occlusions.}
  \label{fig:track2_DUO_cutout}
\end{figure*}

\begin{figure*}[t]
    \centering
    \begin{subfigure}{0.48\linewidth}
        \includegraphics[width=\textwidth]{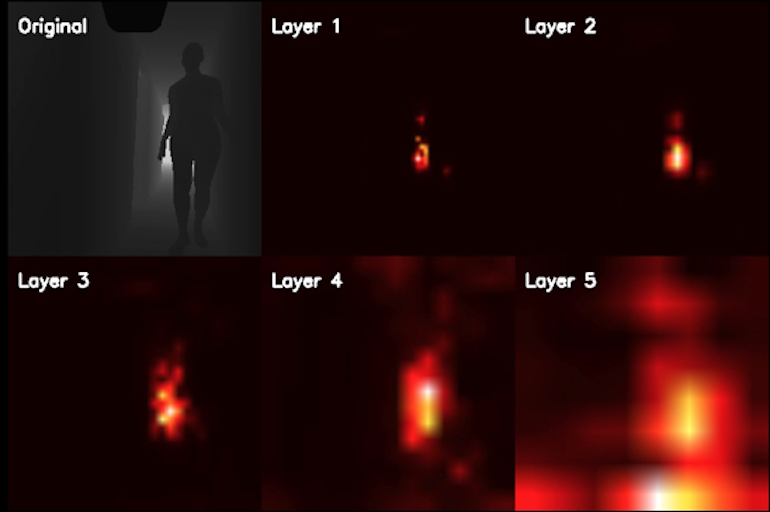}
        \caption{Spatial attention map for the baseline Falcon model.}
        \label{fig:short-a}
    \end{subfigure}
    \hfill
    \begin{subfigure}{0.48\linewidth}
        \includegraphics[width=\textwidth]{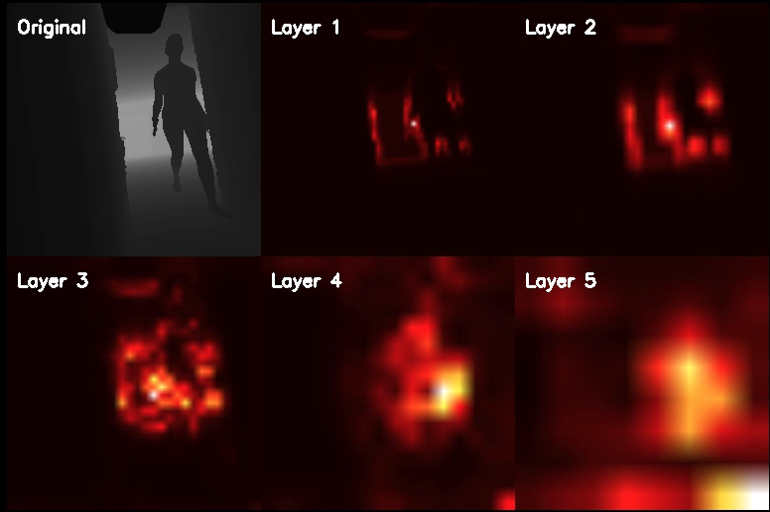}
        \caption{Spatial attention map for our Cutout-augmented model.}
        \label{fig:short-b}
    \end{subfigure}
    \vspace{-0.2cm}
    \caption{Team~[\textcolor{robo_blue}{DUO}]'s attention visualization comparison. (a) Baseline Falcon model. (b) Cutout-augmented model. The augmented model focuses more strongly on human bodies while suppressing irrelevant background details.}
    \label{fig:track2_DUO_example}
\end{figure*}

\noindent\textbf{\faLightbulbO~Key Innovations:}
\begin{itemize}
    \item \textit{Cutout-based occlusion simulation:} Introduces random black rectangular patches into depth frames during training to mimic partial occlusions, enabling the model to generalize better to real-world visibility constraints.
    
    \item \textit{Integration with reinforcement learning:} Seamlessly embeds Cutout augmentation within the DDPPO training pipeline, preserving the stability of reinforcement learning updates while enhancing resilience to viewpoint limitations.
    
    \item \textit{Attention-shifted feature learning:} Visualization of encoder attention maps reveals that the augmented model attends more strongly to human contours and motion cues while suppressing irrelevant background details, improving human-robot interaction awareness.
\end{itemize}

\noindent\textbf{\faGear~Implementation Details:}\\
Built upon the Falcon~\cite{gong2025cognition} architecture, the model employs a ResNet-50 visual encoder and LSTM policy head, trained using DDPPO~\cite{wijmans2019dd} on the Social-HM3D~\cite{gong2025cognition} and RoboSense Track 2 datasets. The Cutout mask aspect ratio ranges from~[0.3, 3.33] with relative scale~[0.02, 0.33]. Training is conducted for 10 M steps on 4$\times$RTX 3090 GPUs with eight parallel simulation environments. Quantitative results show consistent improvements over the baseline Falcon model: on the RoboSense Track 2 dataset, SR by 11.2\%, H-Coll by 6.6\%, and total score by 7 points, demonstrating that Cutout provides an efficient, plug-and-play solution for enhancing occlusion robustness in social navigation without increasing model complexity.

\subsubsection{Team~ [\textcolor{robo_blue}{CityU-ASL}]}
\begin{figure*}[t]
    \centering 
    \includegraphics[width=\linewidth]{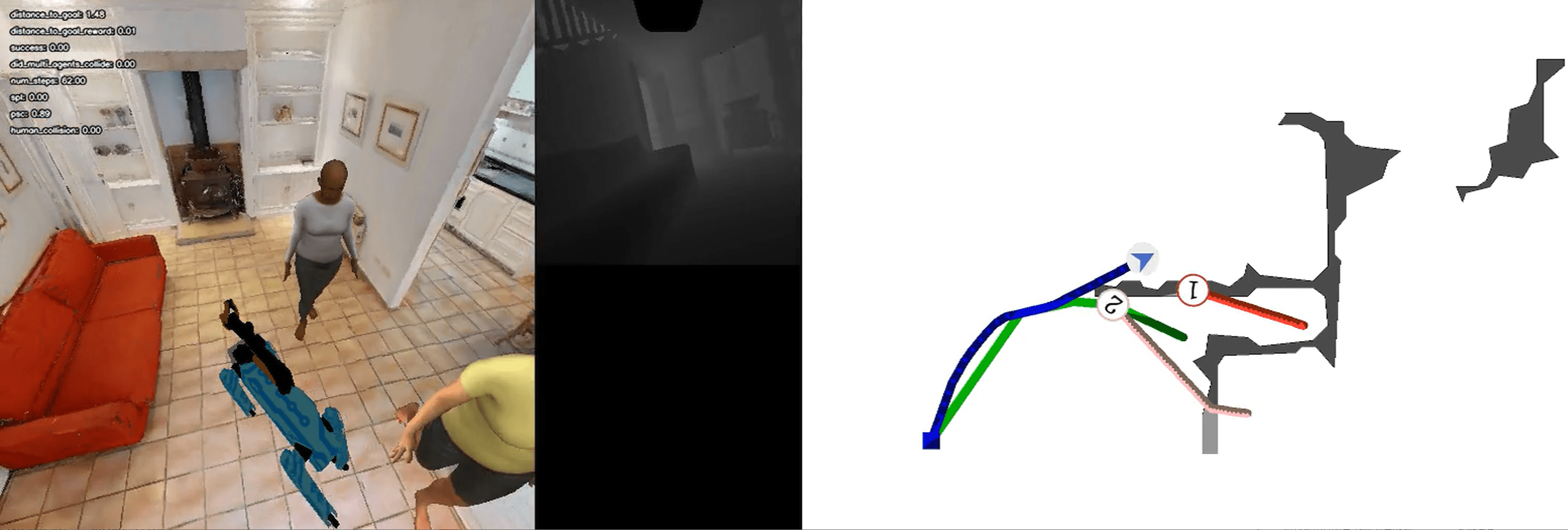}
    \vspace{-0.6cm}
    \caption{Team~[\textcolor{robo_blue}{CityU-ASL}]'s qualitative demonstration. Left: third-person RGB view. Center: robot's first-person depth observation. Right: top-down map with trajectory history (blue: robot, green: ground truth, other colors: humans).}
    \label{fig:track2_CityU_ASL_snapshot} 
\end{figure*}

This team proposed a Hybrid Parameter Optimization strategy to improve socially compliant navigation within the Falcon~\cite{gong2025cognition} framework, as shown in Fig. \ref{fig:track2_CityU_ASL_snapshot}. Their method aims to efficiently balance task completion, path efficiency, and social compliance by systematically tuning key reward function parameters that govern navigation behavior. Rather than modifying the underlying architecture, the approach focuses on optimizing the proportional relationships between reward terms through coupled parameter adjustment and grid search, providing an effective and low-cost pathway to enhance the real-world practicality of Falcon in dynamic human environments.

\noindent\textbf{\faLightbulbO~Key Innovations:}
\begin{itemize}
    \item \textit{Proportional-constrained parameter coupling:} Establishes a fixed ratio between the \textit{task completion reward} and the \textit{facing human distance threshold} parameters, synchronously adjusting them to maintain a consistent trade-off between efficiency and safety. This coupling significantly reduces the dimensionality of the search space and accelerates convergence.
    
    \item \textit{Grid search over coupled parameters:} Performs systematic traversal of the coupled parameters and the \textit{trajectory cover penalty}, achieving comprehensive exploration without excessive computational overhead. This hybrid scheme ensures global coverage while maintaining optimization efficiency.
    
    \item \textit{Reward balancing for social compliance:} Fine-tunes the \textit{trajectory cover penalty} to encourage early evasive maneuvers when potential human-robot collisions are predicted. The optimized reward composition achieves balanced improvements across Success Rate (SR), Success weighted by Path Length (SPL), and Path Safety Compliance (PSC).
\end{itemize}

\noindent\textbf{\faGear~Implementation Details:}\\
The optimization was applied to the original Falcon policy trained with DDPPO~\cite{wijmans2019dd} on the Social-HM3D~\cite{gong2025cognition} dataset. Experiments were conducted on 8$\times$A5000 GPUs with a total of 15 M training steps and a learning rate of $2.5\times10^{-4}$. The best-performing configuration, $\delta=0.2$ (ratio between task completion reward and human distance threshold) and a trajectory cover penalty of $-2.5\times10^{-4}$ -- achieved an SR of 0.770, SPL of 0.706, and PSC of 0.911. These results represent up to a 15\% improvement in success rate compared to the baseline Falcon model, demonstrating that careful parameter coupling and grid-based tuning can yield simultaneous gains in efficiency, safety, and social compliance without altering the model architecture.

\subsubsection{Summary \& Discussion of Track 2}
This track focuses on socially compliant navigation in dynamic human environments, where autonomous agents must move efficiently while respecting human safety, comfort, and behavioral norms. The benchmark evaluates how navigation policies reason about social interactions and adapt to visual occlusions, dynamic obstacles, and crowded layouts. Across all teams, the primary goal was to develop policies that combine motion efficiency with anticipatory and human-aware decision making.

Several converging strategies emerged among the top solutions. Learning efficiency and policy stability were emphasized through hybrid training schemes, which improved sample efficiency and policy robustness without altering the base architecture.

Quantitatively, all five teams achieved notable improvements over the baseline Falcon~\cite{gong2025cognition} across key metrics, including Success Rate (SR), Success weighted by Path Length (SPL), and Path Safety Compliance (PSC). Among them, proactive risk modeling and auxiliary social-awareness tasks consistently produced the highest PSC scores, while hybrid IL-RL schemes yielded superior SR and SPL trade-offs. These findings highlight that high-quality social navigation arises not solely from architectural complexity but from behavioral shaping and reward-aware optimization that align model learning with human-centric objectives.

Looking ahead, future research may extend these ideas toward multi-agent coordination, multimodal perception (\eg, audio or language cues for intent inference), and sim-to-real transfer for embodied robots operating in real crowds. This track thus establishes a strong foundation for bridging classical navigation and socially intelligent autonomy, illustrating that socially compliant motion can emerge from the synergy between perception robustness, anticipatory reasoning, and adaptive policy learning.

\subsection{Track 3: Sensor Placement}
This section introduces the key innovations and implementation details of the three winning solutions in Track 3.
\subsubsection{Team~ [\textcolor{robo_blue}{LRP}]}
This team proposed GBlobs, a local geometric representation designed to enhance LiDAR-based 3D object detection robustness across varying sensor placements. Extending from the previous work \cite{malic2025gblobs}, this method addresses the common issue of ``geometric shortcuts'', where detectors overfit to absolute Cartesian coordinates, causing poor cross-placement generalization. By encoding each local neighborhood as a Gaussian blob, defined by its mean and covariance, GBlobs compel the network to focus on object-centric geometric structures instead of global position cues. This enables the detector to learn shape- and appearance-based representations that remain consistent under sensor configuration shifts.

\begin{figure*}[t]
    \centering 
    \includegraphics[width=\linewidth]{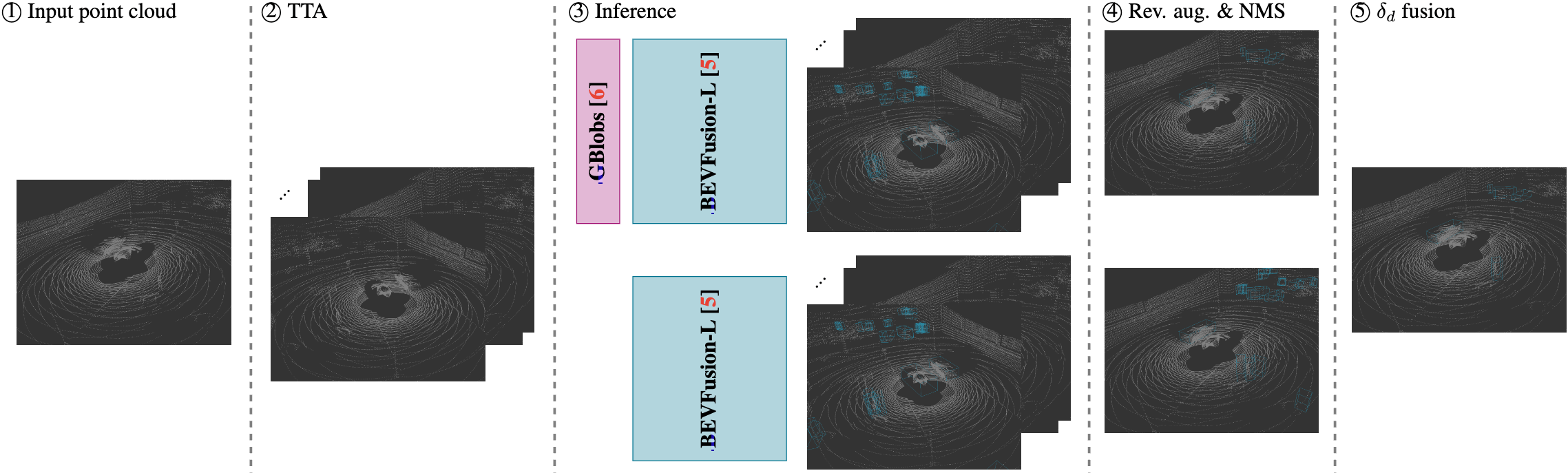}
    \vspace{-0.6cm}
    \caption{Team~[\textcolor{robo_blue}{LRP}]'s GBlobs detection pipeline. Input point clouds \textcircled{1} are augmented \textcircled{2}, then processed by both baseline BEVFusion-L and GBlobs-trained models \textcircled{3}. Predictions are de-augmented and fused via NMS \textcircled{4}, then combined using distance-threshold fusion \textcircled{5}.}
    \label{fig:track3_LRP_framework} 
\end{figure*}

\begin{figure}[t]
    \centering
    \begin{subfigure}{0.48\linewidth}
        \includegraphics[width=\textwidth]{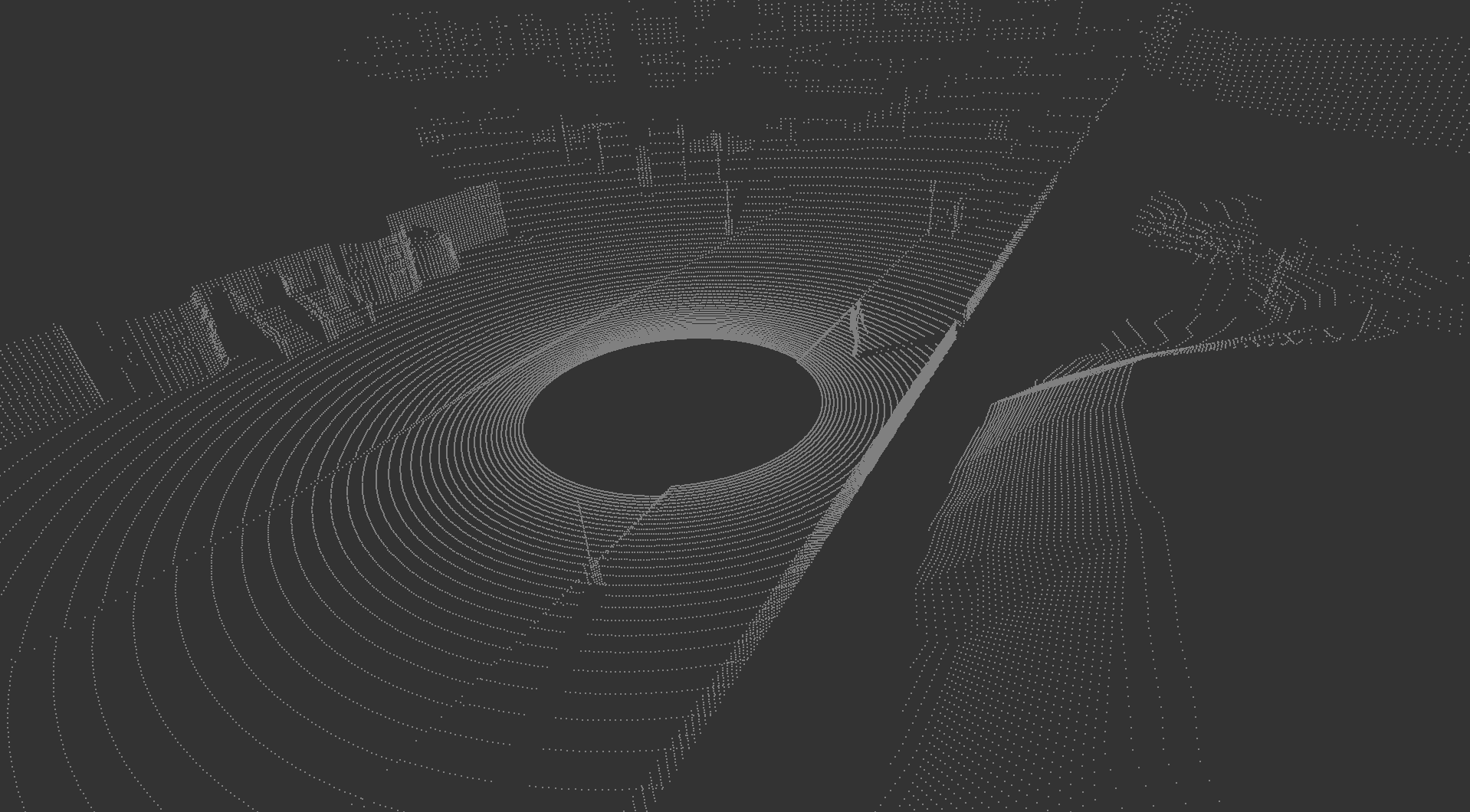}
        \caption{Training Data}
        \label{fig:track3_LRP_data_train}
    \end{subfigure}
    \hfill
    \begin{subfigure}{0.48\linewidth}
        \includegraphics[width=\textwidth]{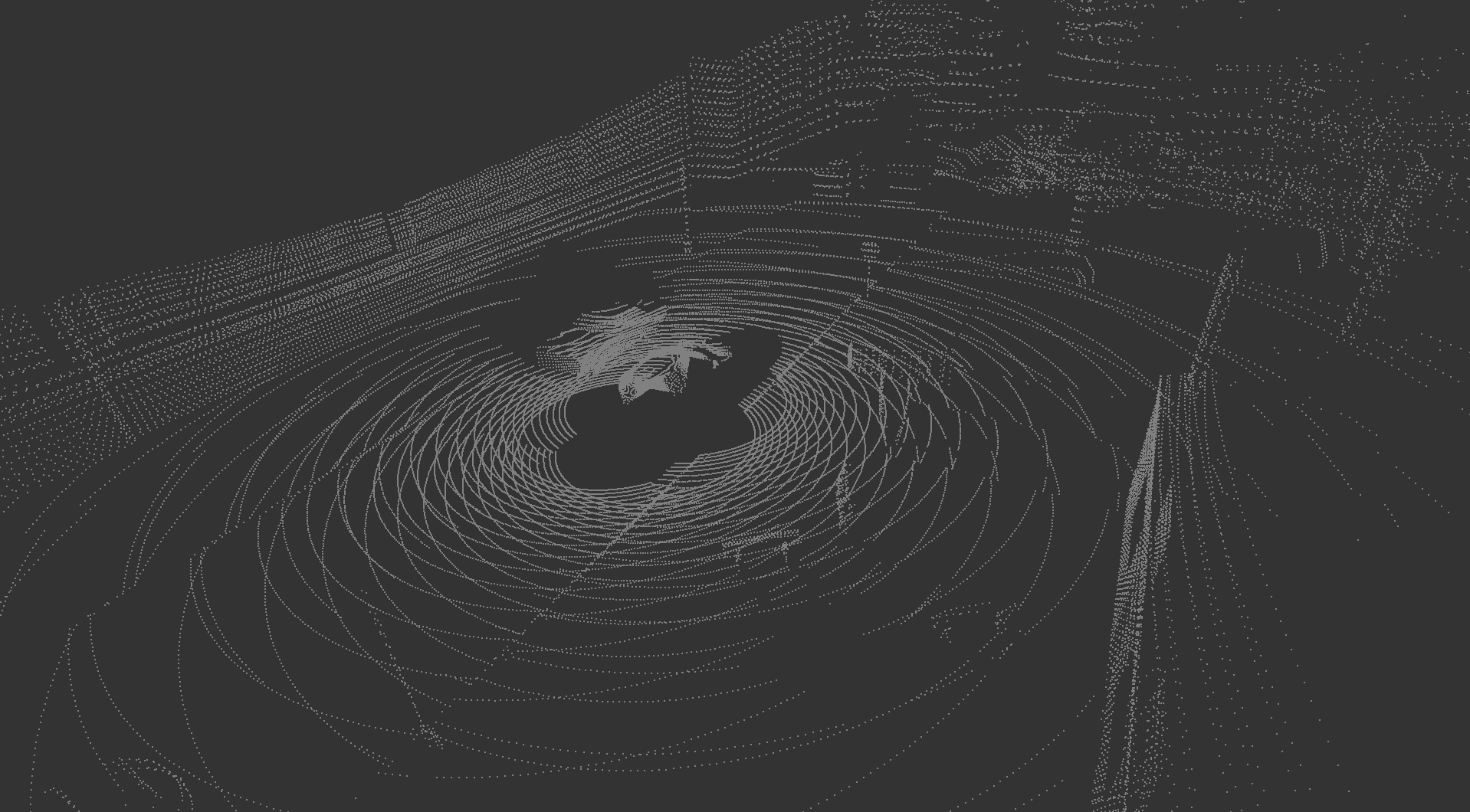}
        \caption{Test Data}
        \label{fig:track3_LRP_data_test}
    \end{subfigure}
    \vspace{-0.2cm}
    \caption{Team~[\textcolor{robo_blue}{LRP}]'s dataset visualization. (a) Training data and (b) test data LiDAR frames from the RoboSense Challenge 2025 Track 3 Sensor Placement dataset, showing viewpoint variations.}
    \label{fig:track3_LRP_dataset}
\end{figure}

\noindent\textbf{\faLightbulbO~Key Innovations:}
\begin{itemize}
    \item \textit{Local geometric encoding with GBlobs:} Replaces absolute point coordinates with localized Gaussian representations, effectively eliminating the geometric shortcut and improving the understanding of intrinsic object geometry.
    
    \item \textit{Dual-model hybrid detection:} Combines a GBlobs-based detector for near-range regions (within 30\,m) and a standard Cartesian-coordinate detector for sparse, far-range regions. The fused prediction balances robustness and completeness across varying point densities.
    
    \item \textit{Test-time augmentation (TTA) and fusion:} Enhances inference stability through rotation, translation, and scaling augmentations; predictions from both detectors are de-augmented, post-processed via NMS, and fused using a distance-based threshold.
\end{itemize}

\noindent\textbf{\faGear~Implementation Details:}\\
The framework builds upon the LiDAR-only BEVFusion-L~\cite{liu2023bevfusion} detector and is trained on the CARLA-based synthetic dataset provided by Place3D \cite{li2024place3d}. Training uses class-balanced sampling for 90 epochs with Adam optimizer and cyclical learning rate (max 0.001), voxel size $[0.075, 0.075, 0.2]$\,m, and accumulated 10-frame sequences. During inference, both GBlobs and global detectors process ten augmented frames per sample. Predictions are fused using a distance threshold $\delta_d=30$\,m, assigning near-range detections to the GBlobs model and far-range detections to the global model. On the official leaderboard, the proposed system demonstrates that explicitly modeling local geometric structure yields strong generalization to unseen LiDAR placements.

\subsubsection{Team~ [\textcolor{robo_blue}{Point Loom}]}
This team developed an enhanced LiDAR-only 3D object detection framework to improve robustness under variable sensor placements. Their approach builds on the BEVFusion~\cite{liu2023bevfusion} LiDAR branch and introduces three complementary strategies: temporal sequence enhancement, placement-mixed training, and test-time augmentation (TTA), to strengthen perception consistency across heterogeneous configurations. The central insight is that robustness to sensor variation can be achieved without altering network architecture by integrating temporal richness, diverse spatial exposure, and inference stabilization within a unified pipeline.

\begin{figure*}
  \centering
    \includegraphics[width=\linewidth]{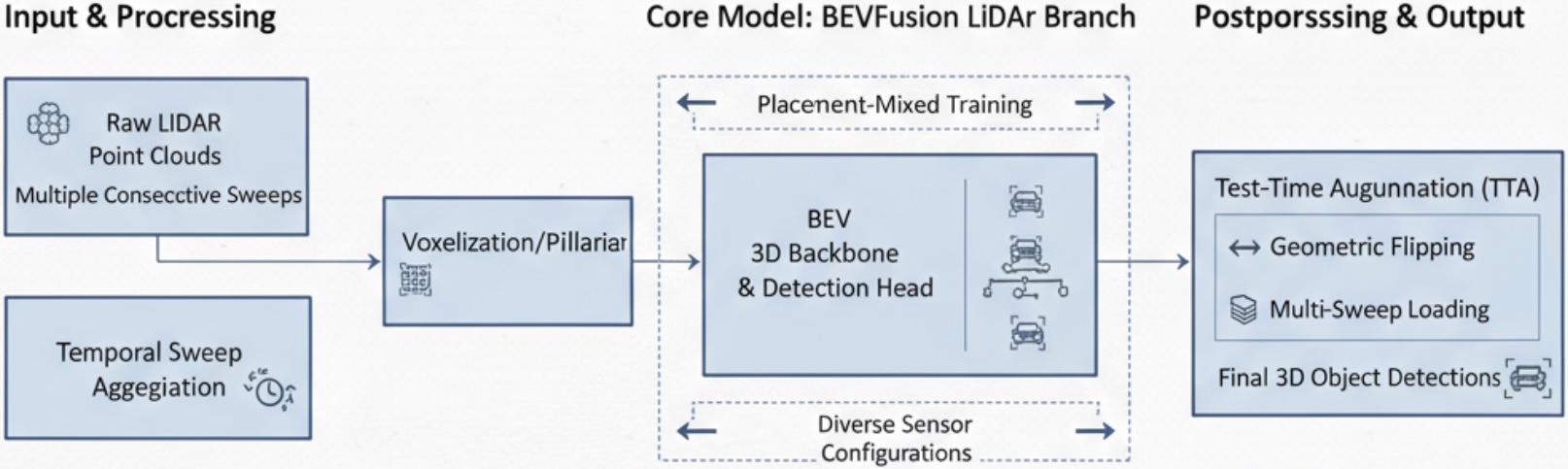}
    \vspace{-0.6cm}
    \caption{Team~[\textcolor{robo_blue}{Point Loom}]'s multi-strategy enhancement framework. The pipeline integrates temporal sequence aggregation, placement-mixed training, and test-time augmentation to achieve generalizable 3D detection across sensor placements.}
    \label{fig:track3_Point_Loom_method}
\end{figure*}

\noindent\textbf{\faLightbulbO~Key Innovations:}
\begin{itemize}
    \item \textbf{Temporal sequence enhancement:} Aggregates multiple consecutive LiDAR sweeps before voxelization, enriching spatial density and motion continuity to mitigate sparsity and occlusion in single-frame inputs.
    \item \textbf{Placement-mixed training:} Merges training data from different LiDAR setups to expose the detector to varied viewpoints, encouraging the model to learn placement-invariant geometric representations.
    \item \textbf{Inference-time augmentation:} Applies geometric flipping and multi-sweep loading at test time; predictions are fused through averaging and NMS to improve stability and reduce directional bias.
\end{itemize}

\noindent\textbf{\faGear~Implementation Details:}\\
The framework is implemented using MMDetection3D~\cite{mmdet3d} with the official BEVFusion LiDAR-only configuration. Temporal sweeps are aligned to a common reference frame prior to voxelization, and placement-mixed datasets are constructed by combining samples from multiple sensor layouts. Models are trained for 90 epochs using the Adam optimizer (max LR 0.001) and class-balanced sampling. During inference, five augmented passes are fused via distance-weighted averaging. On the official Track 3 benchmark, the full model achieved a mAP of 90.4 under fixed placements and 74.6 under variable placements, outperforming the BEVFusion baseline (87.9 / 63.0) and demonstrating the effectiveness of combining temporal cues, placement diversity, and inference augmentation for scalable 3D perception.

\subsubsection{Team~ [\textcolor{robo_blue}{Smartqiu}]}
This team presented a refined LiDAR-only detection framework aimed at improving cross-placement generalization for 3D object detection. Building upon the BEVFusion~\cite{liu2023bevfusion} model, the method emphasizes \textbf{data-centric robustness} through multi-sweep aggregation and placement-mixed training, avoiding any architectural modifications. The underlying motivation is that spatial sparsity and placement bias are major causes of performance degradation when LiDAR sensors are mounted in different vehicle positions. By jointly leveraging temporal consistency and placement diversity, the framework encourages the detector to learn geometry-aware, placement-invariant features.

\begin{figure}
  \centering
    \includegraphics[width=\linewidth]{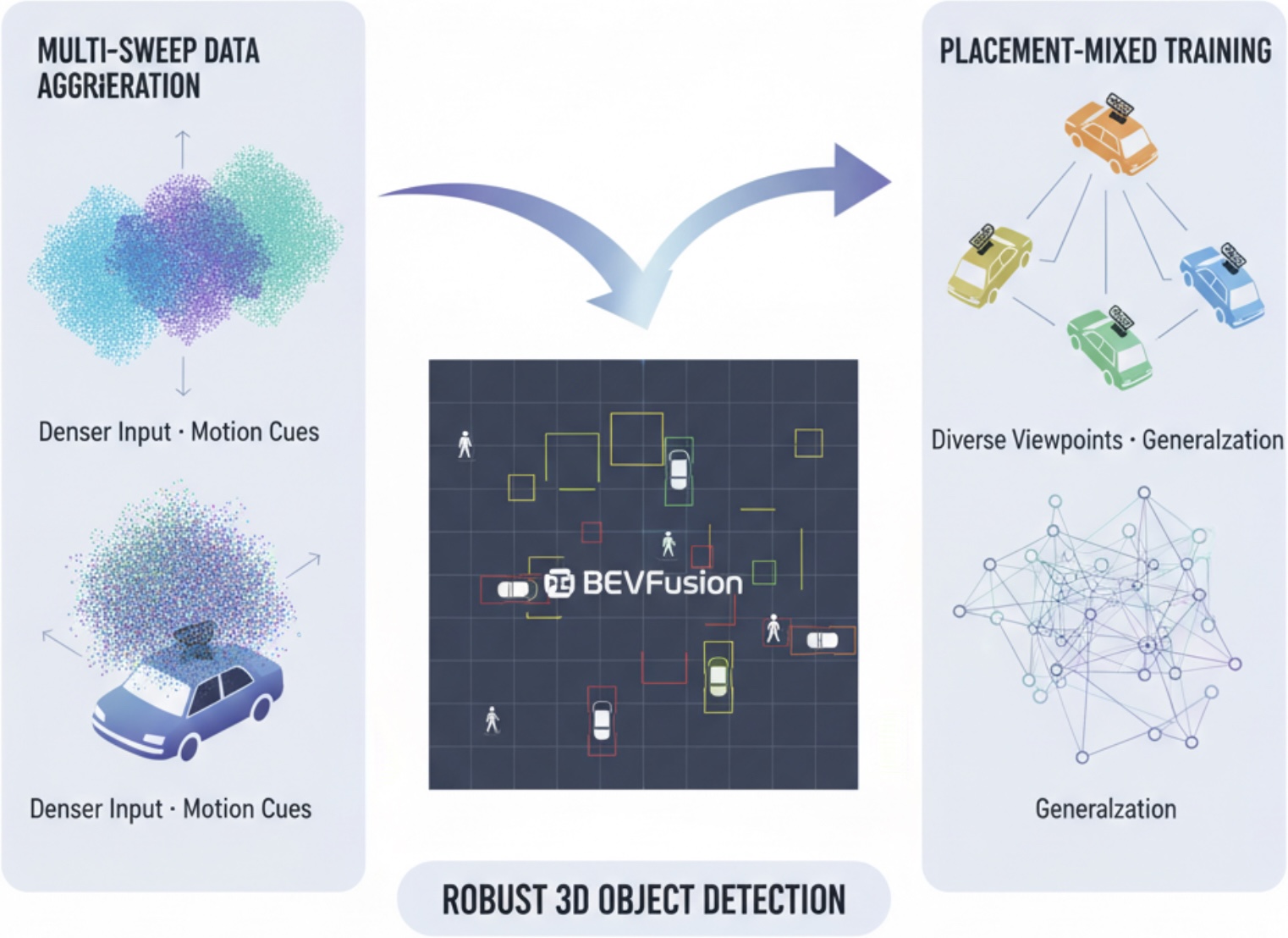}
    \caption{Team~[\textcolor{robo_blue}{Smartqiu}]'s data-centric robustness approach. The framework combines multi-sweep temporal aggregation with placement-mixed training to improve cross-placement generalization without architectural modifications.}
    \label{fig:track3_Smartqiu_method}
\end{figure}

\noindent\textbf{\faLightbulbO~Key Innovations:}
\begin{itemize}
    \item \textit{Multi-sweep aggregation:} Consecutive LiDAR sweeps are temporally aligned and merged before voxelization, forming a denser and motion-consistent 3D representation that improves detection of partially occluded or distant objects.
    
    \item \textit{Placement-mixed training:} Datasets collected from multiple LiDAR mounting positions are unified into a single training set, exposing the detector to diverse viewpoints and input distributions. This simulates real-world sensor variability and encourages placement-agnostic learning.
    
    \item \textit{Architecture-agnostic integration:} Both strategies are directly applied within the BEVFusion-L \cite{liu2023bevfusion} pipeline, requiring no additional parameters or architectural redesign, ensuring high compatibility and ease of deployment across different vehicle platforms.
\end{itemize}

\noindent\textbf{\faGear~Implementation Details:}\\
The model is implemented based on the official BEVFusion-L \cite{liu2023bevfusion}. Training follows the standard settings, including Adam optimizer, cosine learning rate schedule, and voxelization consistent with the baseline. Multi-sweep aggregation fuses sequential frames during preprocessing, while placement-mixed training combines samples from diverse sensor layouts into a unified split. The enhanced model demonstrates clear performance gains over the baseline: mAP 0.605 → 0.729 → 0.743 when successively adding multi-sweep and placement-mixing strategies. These results confirm that data-centric enhancement can yield substantial robustness improvements under variable sensor configurations without additional computational or architectural cost.

\subsubsection{Team~ [\textcolor{robo_blue}{DZT328}]}
This team proposed PlaceRecover, a Transformer-based point cloud recovery framework that reconstructs canonical, placement-invariant representations from LiDAR data collected under diverse sensor configurations. The method addresses a central challenge in 3D detection -- the loss of geometric consistency when transferring models across vehicles with differing LiDAR layouts. By leveraging implicit neural representations (INR) and point-wise transformers, PlaceRecover effectively disentangles geometry from the sensor perspective, restoring a consistent global spatial structure before detection. The framework operates as a pre-detection reconstruction stage and integrates seamlessly with BEVFusion-L~\cite{liu2023bevfusion} to improve cross-placement robustness and geometric fidelity. Its modular design and strong empirical performance for 3D object detection under different sensor placements.

\begin{figure}
    \centering
    \includegraphics[width=\linewidth]{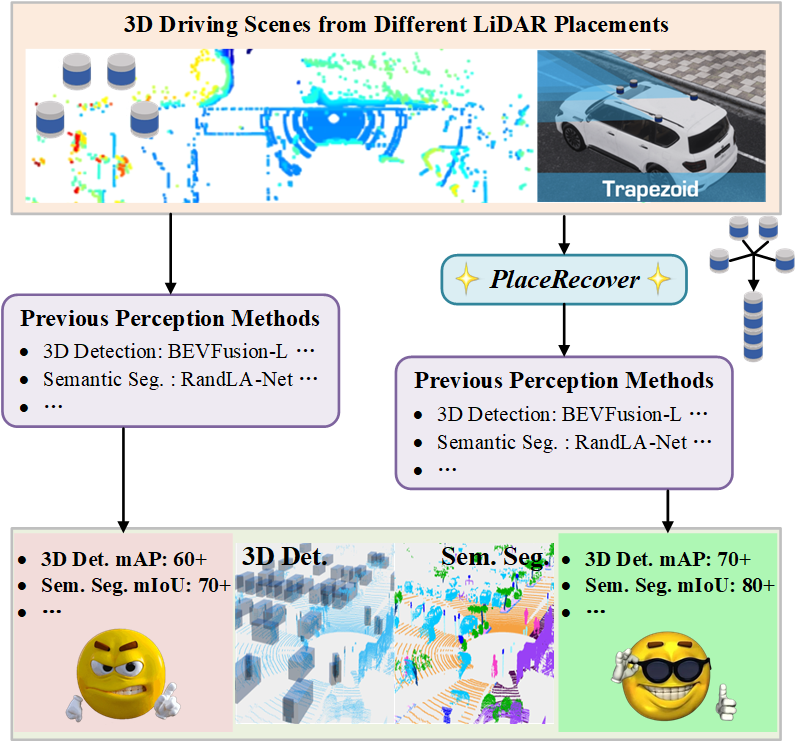}
    \vspace{-0.6cm}
    \caption{Team~[\textcolor{robo_blue}{DZT328}]'s PlaceRecover performance improvement. Point cloud perception methods show significant accuracy gains across different sensor placements when augmented with PlaceRecover.}
    \label{fig:track3_DZT328_abstract}
\end{figure}

\begin{figure*}
    \centering
    \includegraphics[width=\linewidth]{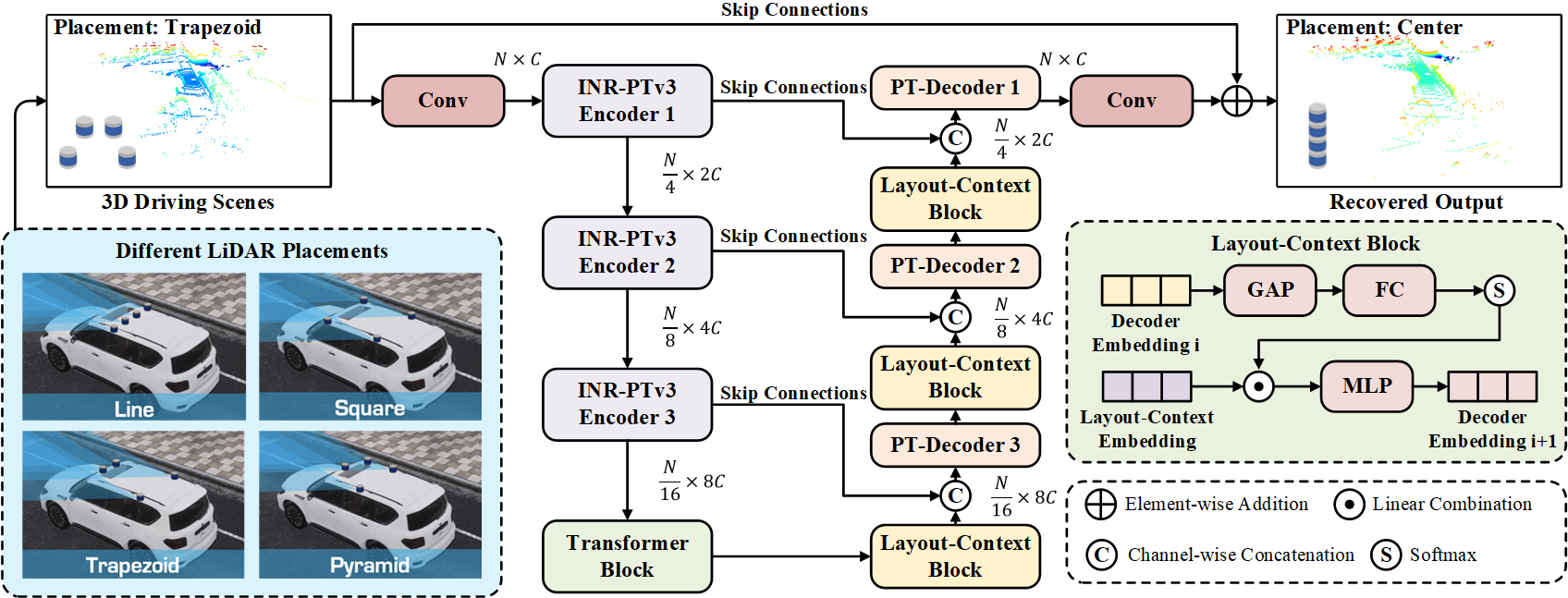}
    \vspace{-0.6cm}
    \caption{Team~[\textcolor{robo_blue}{DZT328}]'s PlaceRecover architecture. The framework employs INR-PointTransformer encoders, a central Transformer block, and PointTransformer decoders with Layout-Context Blocks to recover placement-invariant point clouds.}
    \label{fig:track3_DZT328_method}
\end{figure*}

\begin{figure}
    \centering
    \includegraphics[width=\linewidth]{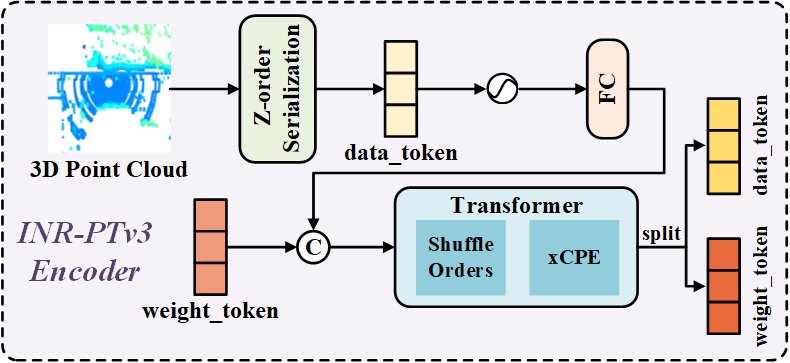}
    \vspace{-0.6cm}
    \caption{Team~[\textcolor{robo_blue}{DZT328}]'s INR-PTv3 encoder structure. Input point clouds are serialized into data tokens, combined with learnable weight tokens, then refined through a Transformer with Shuffle Orders and xCPE modules.}
    \label{fig:track3_DZT328_encoder}
\end{figure}

\begin{figure*}
    \centering
    \includegraphics[width=\linewidth]{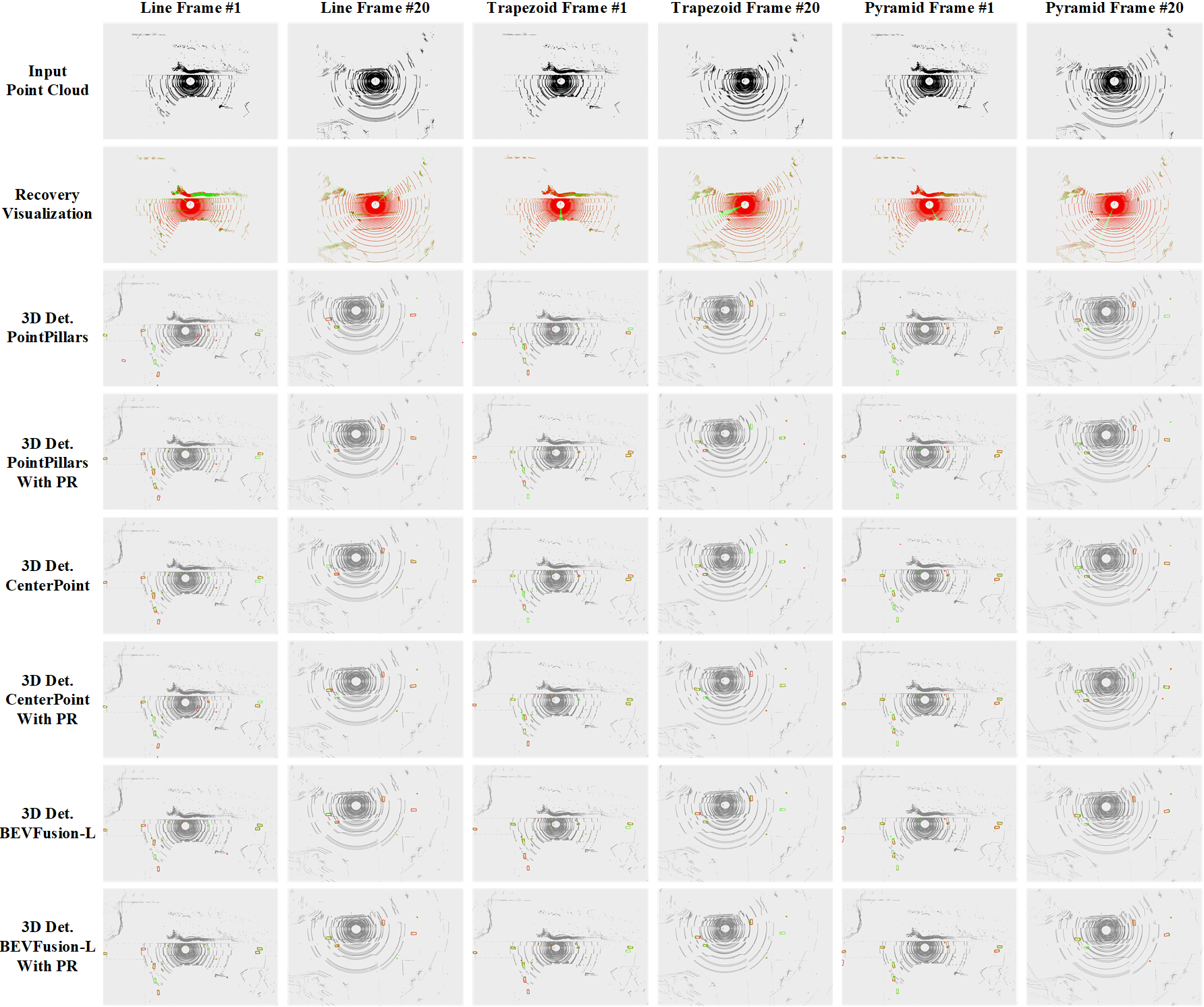}
    \vspace{-0.6cm}
    \caption{Team~[\textcolor{robo_blue}{DZT328}]'s qualitative comparison on validation set. Row 1: input point clouds from different placements (Line, Trapezoid, Pyramid). Row 2: recovery results (red: PlaceRecover predictions, green: ground truth at Center placement). Rows 3-8: detection results from PointPillars, CenterPoint, and BEVFusion-L with/without PlaceRecover (red: predictions, green: ground truth).}
    \label{fig:track3_DZT328_visualization}
\end{figure*}

\noindent\textbf{\faLightbulbO~Key Innovations:}
\begin{itemize}
    \item \textit{INR-PointTransformer encoder:} Couples PointTransformer V3 with an implicit neural representation (INR) module to encode spatial priors independent of sensor placement. Learnable INR weight tokens project layout-dependent features into a shared canonical latent space, promoting geometric alignment across configurations.
    
    \item \textit{Layout-aware decoder with context blocks:} Employs a Layout-Context Block that adaptively refines the decoded features by injecting layout embeddings, enabling the network to restore high-frequency geometric details even when input point clouds are sparse or unevenly distributed.
    
    \item \textit{Cross-placement 3D detection integration:} Reconstructed point clouds are directly consumed by BEVFusion-L \cite{liu2023bevfusion} for downstream 3D detection, demonstrating significant cross-layout robustness. The model improves mAP by +11.9 over the baseline and achieves notable gains in Chamfer Distance and Earth Mover’s Distance metrics, validating the geometric precision of reconstruction.
\end{itemize}

\noindent\textbf{\faGear~Implementation Details:}\\
The model is implemented using MMDetection3D~\cite{mmdet3d} in PyTorch and trained on the RoboSense 2025 Track 3 dataset. The encoder and decoder each comprise three transformer layers with conditional positional encoding (xCPE) and residual skip connections. Training employs the Adam optimizer with cosine annealing, a batch size of 1, and a learning rate of $5\times10^{-5}$ for 10 epochs on a single RTX 4090 GPU. The encoder is initialized from pre-trained PointTransformer V3 weights to accelerate convergence. Quantitative evaluations show strong geometric fidelity (Chamfer Distance: 0.027, Earth Mover’s Distance: 0.105). When integrated with BEVFusion-L \cite{liu2023bevfusion}, PlaceRecover achieves mAP 0.807 / mATE 0.097 on the validation set and mAP 0.726 / mATE 0.117 on the test set, outperforming the baseline (0.605 mAP) by a large margin. The results demonstrate that explicitly reconstructing placement-invariant geometry through INR-Transformer design establishes a robust and generalizable foundation for LiDAR perception under varying sensor layouts.

\subsubsection{Team~ [\textcolor{robo_blue}{seu\_zwk}]}
This team introduced a unified and robust 3D object detection framework designed to generalize effectively across heterogeneous LiDAR placements. Their approach, titled \textbf{Layout-Robust 3D Detection via Multi-Representation Fusion (MRF)}, directly addresses the issue of viewpoint-dependent feature distortion and distribution shifts caused by sensor heterogeneity -- such as differing LiDAR beam counts, mounting heights, and spatial orientations. By combining geometric complementarity across multiple representations and motion-guided temporal reasoning, the framework achieves stable and layout-invariant perception across a wide range of driving configurations.

\noindent\textbf{\faLightbulbO~Key Innovations:}
\begin{itemize}
    \item \textit{Multi-representation fusion module (MRFM):} Integrates point-, voxel-, and BEV-level features through view-wise self-attention and point-voxel cross-attention, enabling the model to leverage both fine-grained geometry and global context. This fusion allows the detector to remain stable under varying sampling densities and layout-induced distortions.
    
    \item \textit{Motion-guided spatio-temporal fusion (MG-STF):} Utilizes motion residuals between consecutive frames as temporal priors, injecting them into a transformer-based fusion head to refine dynamic object trajectories and improve inter-frame consistency for moving targets.
    
    \item \textit{Consistency-regularized learning objective:} Incorporates a Hybrid Temporal Consistency (HTC) loss that enforces feature coherence between temporally adjacent frames, jointly constraining alignment between historical and current representations to maintain smooth temporal transitions.
\end{itemize}

\noindent\textbf{\faGear~Implementation Details:}\\
The proposed framework is implemented upon PV-RCNN++~\cite{shi2023pv} using MMDetection3D~\cite{mmdet3d}. Each input frame is processed through PointNet++ and sparse 3D convolutions to produce three complementary representations -- point-level, voxel-level, and BEV-level -- which are then fused by MRFM. MG-STF introduces an auxiliary motion extractor based on flow-guided residual features, feeding temporal embeddings into the fusion transformer to enhance time-domain reasoning. The network is trained for 90 epochs with the Adam optimizer, cosine learning rate schedule, and batch size of 8 on 4$\times$A100 GPUs. Quantitatively, the model achieves mAP 0.724 and NDS 0.655 on the official RoboSense dataset, surpassing BEVFusion-L by +11.6\% and +13.9\%, respectively. Detailed class-wise analysis highlights substantial improvements on dynamic object categories -- pedestrian (+43.4\%), motorcycle (+7.3\%), and bicycle (+15.4\%) -- showcasing the model’s ability to reason about motion and geometry simultaneously. These results validate that jointly modeling multi-representation fusion and temporal consistency establishes a principled and scalable path toward layout-robust 3D detection across diverse sensor configurations.

\begin{figure}
    \centering
    \includegraphics[width=\linewidth]{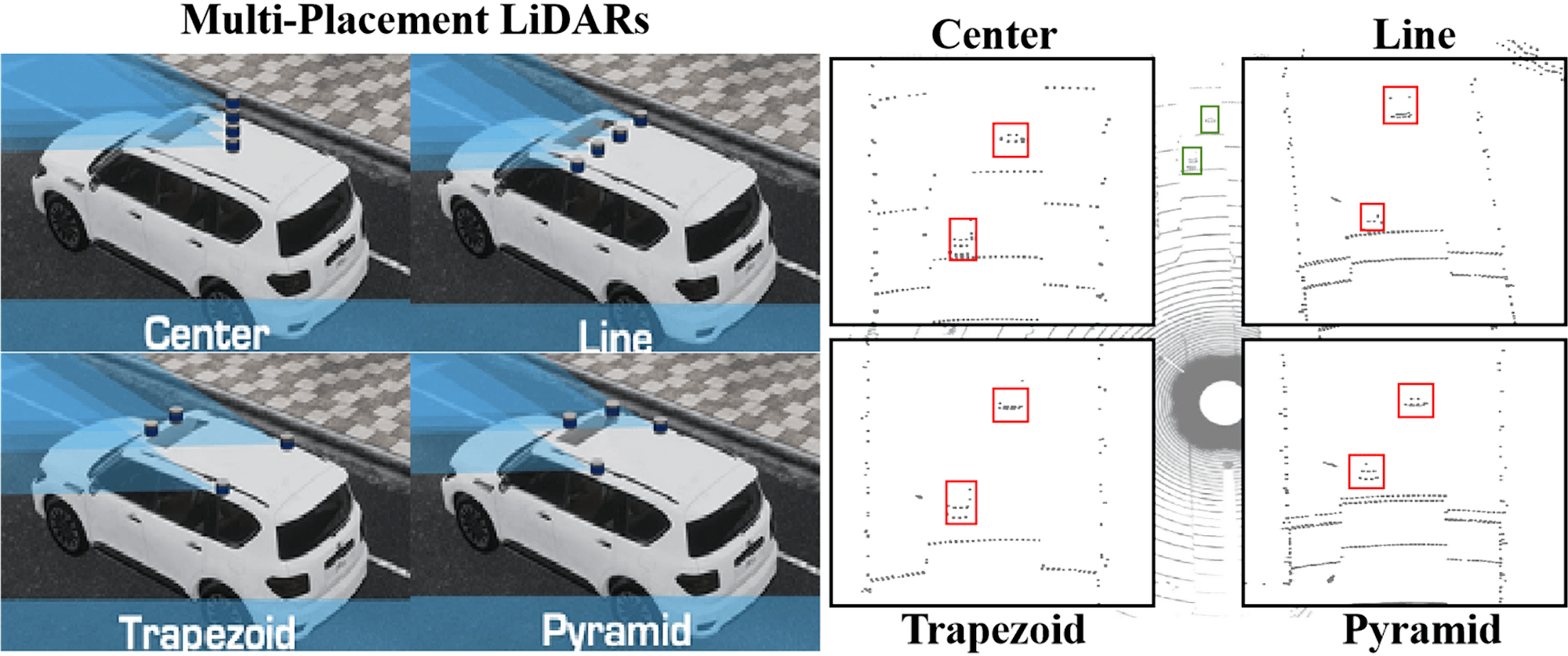}
    \vspace{-0.6cm}
    \caption{Team~[\textcolor{robo_blue}{seu\_zwk}]'s motivation for layout-robust detection. Illustration of viewpoint-dependent feature distortion and distribution shifts caused by heterogeneous LiDAR sensor configurations.}
    \label{fig:track3_seu_zwk_intro}
\end{figure}

\begin{figure*}
    \centering
    \includegraphics[width=0.85\linewidth]{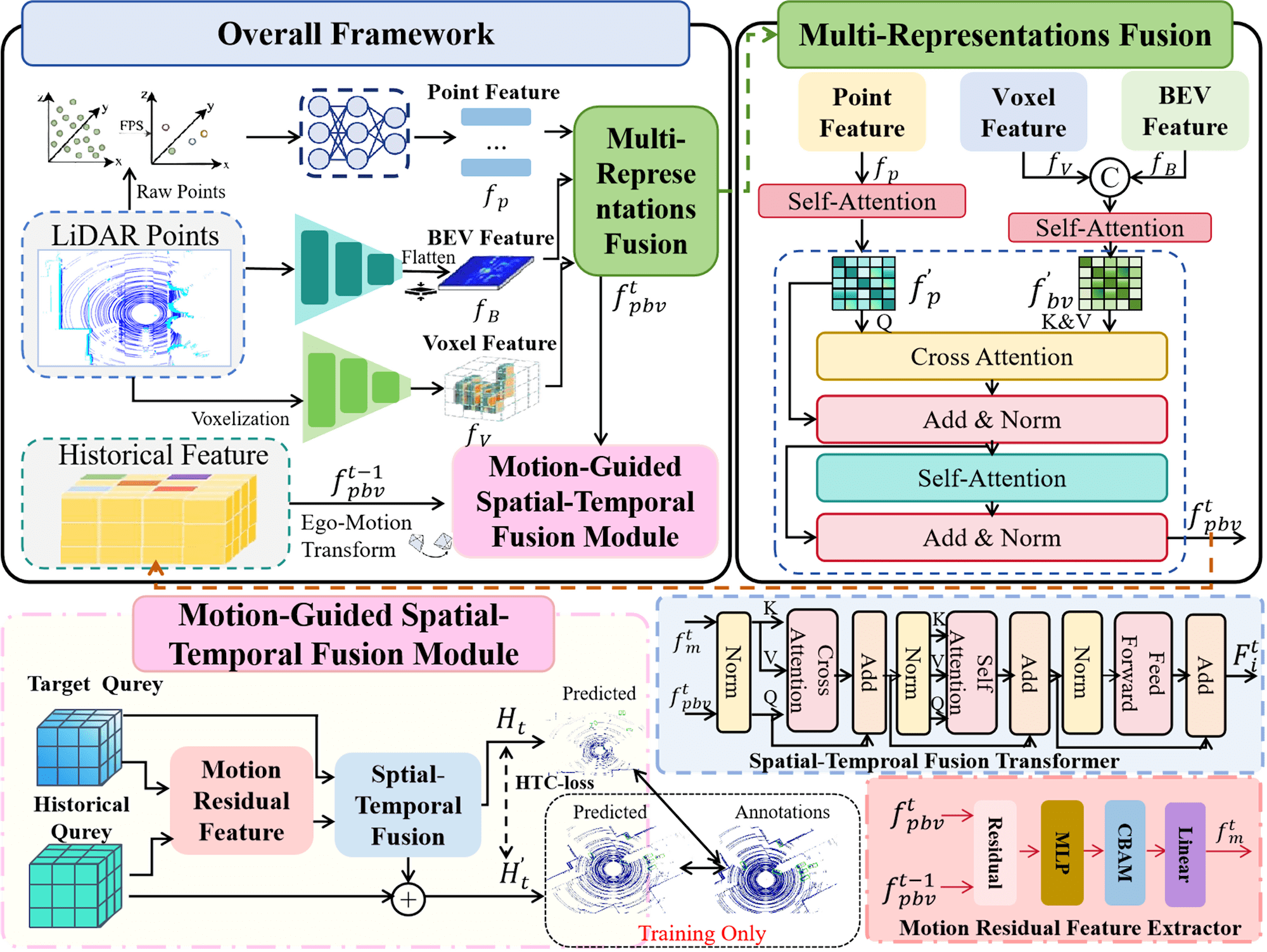}
    \vspace{-0.1cm}
    \caption{Team~[\textcolor{robo_blue}{seu\_zwk}]'s Multi-Representation Fusion framework. The MRFM fuses point ($\mathbf{f}_{p}$), voxel ($\mathbf{f}_{v}$), and BEV ($\mathbf{f}_{B}$) features via self- and cross-attention to produce $\mathbf{f}^{t}_{pbv}$. Motion-Guided Spatial-Temporal Fusion then combines temporal residuals for final prediction $\mathbf{H}_{t}$.}
    \label{fig:track3_seu_zwk_method}
\end{figure*}

\subsubsection{Summary \& Discussion of Track 3}
This track addresses the challenge of sensor placement variability in LiDAR-based 3D object detection: a key barrier to deploying perception models across diverse robotic platforms. Variations in mounting height, orientation, and beam configuration introduce spatial distribution shifts that can degrade accuracy. The goal is to develop models that retain reliable performance under unseen placements, advancing toward layout-invariant 3D perception.

Two major trends emerged. The first is \textit{representation-level adaptation}, where models align or reconstruct geometric features to mitigate layout-induced bias. Approaches that fuse multi-level representations (point, voxel, and BEV) or recover canonical point clouds through implicit neural representations consistently achieved higher cross-layout robustness. The second trend is \textit{data-centric generalization}, which improves robustness through multi-sweep aggregation, placement-mixed training, and coordinate-independent encodings. These strategies expose detectors to diverse viewpoints and temporal continuity without altering network design.

Empirically, most solutions achieved substantial gains over the BEVFusion-L~\cite{liu2023bevfusion} baseline, with double-digit improvements in mAP and lower translation error. The strongest systems balanced structural modeling with temporal reasoning and dataset diversity, achieving both stability and efficiency.  

Future work may extend these ideas to \textit{cross-platform adaptation} (\eg, vehicle-drone-quadruped transfer) and multimodal alignment under varying extrinsics. The findings suggest that robust 3D perception depends less on architectural complexity and more on learning geometric structure that remains consistent regardless of sensor placement.

\subsection{Track 4: Cross-Modal Drone Navigation}
This section introduces the key innovations and implementation details of the three winning solutions in Track 4.
\subsubsection{Team~ [\textcolor{robo_blue}{TeleAI}]}
This team presented a Parameter-Efficient Mixture-of-Experts (PE-MoE) framework for robust cross-modal geo-localization and drone navigation, targeting retrieval scenarios across drastically different viewpoints such as aerial, satellite, and ground-level imagery. The method aims to address two major challenges: (1) large-scale domain shifts between modalities due to spatial and appearance discrepancies, and (2) semantic mismatch between verbose training captions and concise, query-style test descriptions. By integrating domain-aligned textual preprocessing, expert-specialized representations, and parameter-efficient fine-tuning, the framework achieves strong cross-modal alignment with minimal computational overhead. The design emphasizes modularity and transferability, making it highly suitable for scalable real-world navigation systems involving heterogeneous visual sensors.

\begin{figure*}[t]
    \centering
    \includegraphics[width=\textwidth]{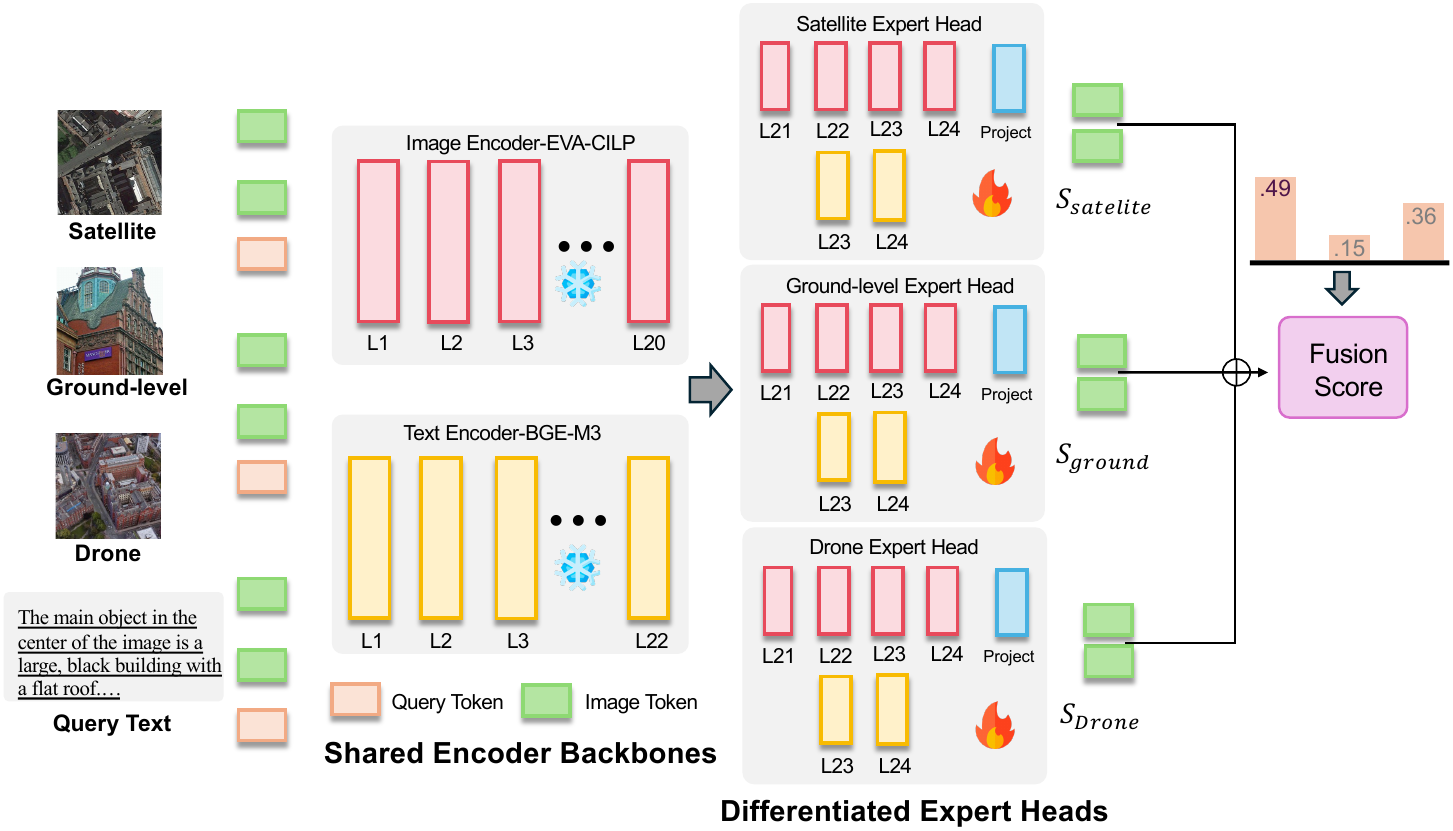} 
    \vspace{-0.6cm}
    \caption{Team~[\textcolor{robo_blue}{TeleAI}]'s Parameter-Efficient Mixture-of-Experts framework. A shared backbone extracts features, which are processed by a dynamic gating network and specialized expert heads (satellite, drone, ground) to produce modality-aware retrieval scores.}
    \label{fig:track4_TeleAI_framework}
\end{figure*}

\noindent\textbf{\faLightbulbO~Key Innovations:}
\begin{itemize}
    \item \textit{Domain-aligned data preprocessing:} The training corpus is partitioned by sensing modality (satellite, drone, ground) to isolate platform-specific biases. Each textual caption is further decomposed into short, query-style statements using a rule-based splitting pipeline, narrowing the semantic gap between training data and test inputs. This preprocessing improves language consistency and facilitates downstream expert specialization.
    
    \item \textit{Parameter-efficient Mixture-of-Experts design:} Built upon frozen EVA-CLIP and BGE-M3 encoders, the framework introduces three lightweight expert heads, each dedicated to a distinct sensing platform. These heads are fine-tuned with LoRA-style adapters and share a unified projection space for joint optimization. A compact gating network dynamically routes incoming text queries to the most relevant expert based on modality priors, fusing the outputs through weighted similarity aggregation for final retrieval.
    
    \item \textit{Platform-aware augmentation and text sanitization:} To enhance alignment, visual augmentations (random rotations, color jitter, and resolution scaling) are applied selectively to satellite and aerial imagery to emulate real-world perturbations. Meanwhile, language sanitization automatically removes directional terms (\eg, “north,” “left side”) that could introduce modality-specific biases, ensuring consistent geometric reasoning across views.
\end{itemize}

\noindent\textbf{\faGear~Implementation Details:}\\
The shared vision and text encoders remain frozen, while the expert heads and gating network are optimized jointly through a two-stage curriculum. In the first stage, each expert head is trained on its platform-specific subset using a symmetric contrastive loss; in the second stage, the gating network is trained end-to-end to adaptively weight expert outputs based on cross-modal similarity signals. Optimization uses the AdamW optimizer ($2\times10^{-5}$) with cosine scheduling, a batch size of 128, and a maximum of 25 epochs on 4$\times$A100 GPUs. To prevent overfitting, training incorporates modality-balanced sampling and online hard example mining. Quantitatively, the model achieves a Recall@1 = 38.31 on the University-1652 benchmark, surpassing the baseline by +12.87 points. The proposed PE-MoE demonstrates that robust cross-modal retrieval can be realized without large-scale retraining, highlighting the potential of parameter-efficient multi-expert modeling for practical drone navigation and geo-localization tasks.

\subsubsection{Team~ [\textcolor{robo_blue}{rhao\_hur}]}
This team introduced HCCM (Hierarchical Cross-Granularity Contrastive and Matching Learning), a unified framework designed to enhance fine-grained visual-language alignment and compositional semantic reasoning for natural language-guided drone navigation. The method addresses two key challenges of the cross-view retrieval setting: (1) existing vision-language models primarily emphasize global alignment, neglecting the hierarchical structure of local-to-global semantics; and (2) entity-level partitioning approaches are brittle under complex drone scenes characterized by large fields of view and overlapping regions. HCCM proposes a hierarchy-aware modeling strategy that jointly captures regional and global semantics through contrastive and matching objectives, reinforced by a momentum-based distillation mechanism for training stability and generalization.

\begin{figure}[t]
    \centering
    \includegraphics[width=\linewidth]{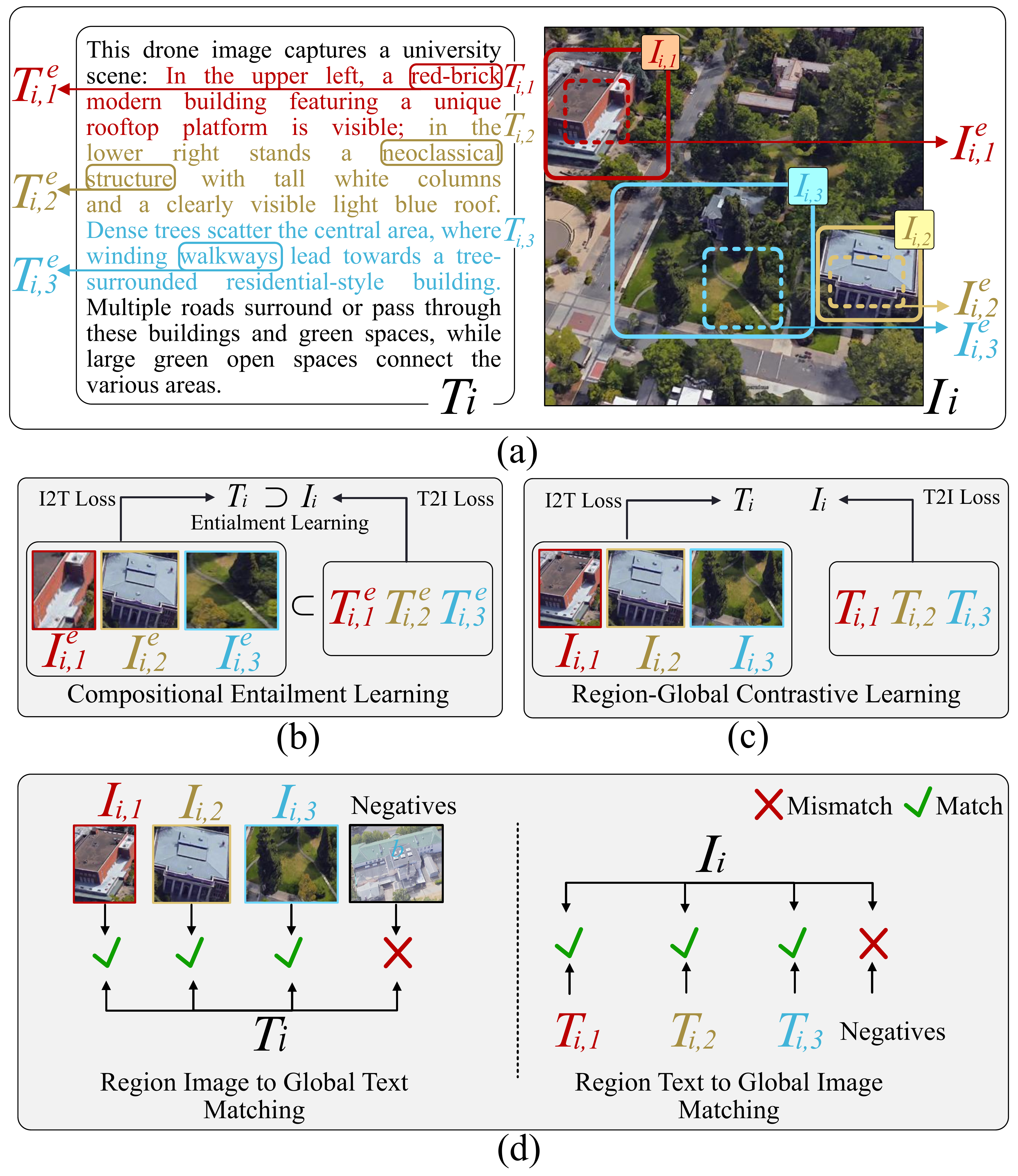}
    \vspace{-0.6cm}
    \caption{Team~[\textcolor{robo_blue}{rhao\_hur}]'s motivation for hierarchical alignment. Illustration of the semantic gap between global and local features in cross-view drone navigation scenarios.}
  \label{fig:track4_rhao_hur_motivation}
\end{figure}

\begin{figure}[t]
    \centering
    \includegraphics[width=\linewidth]{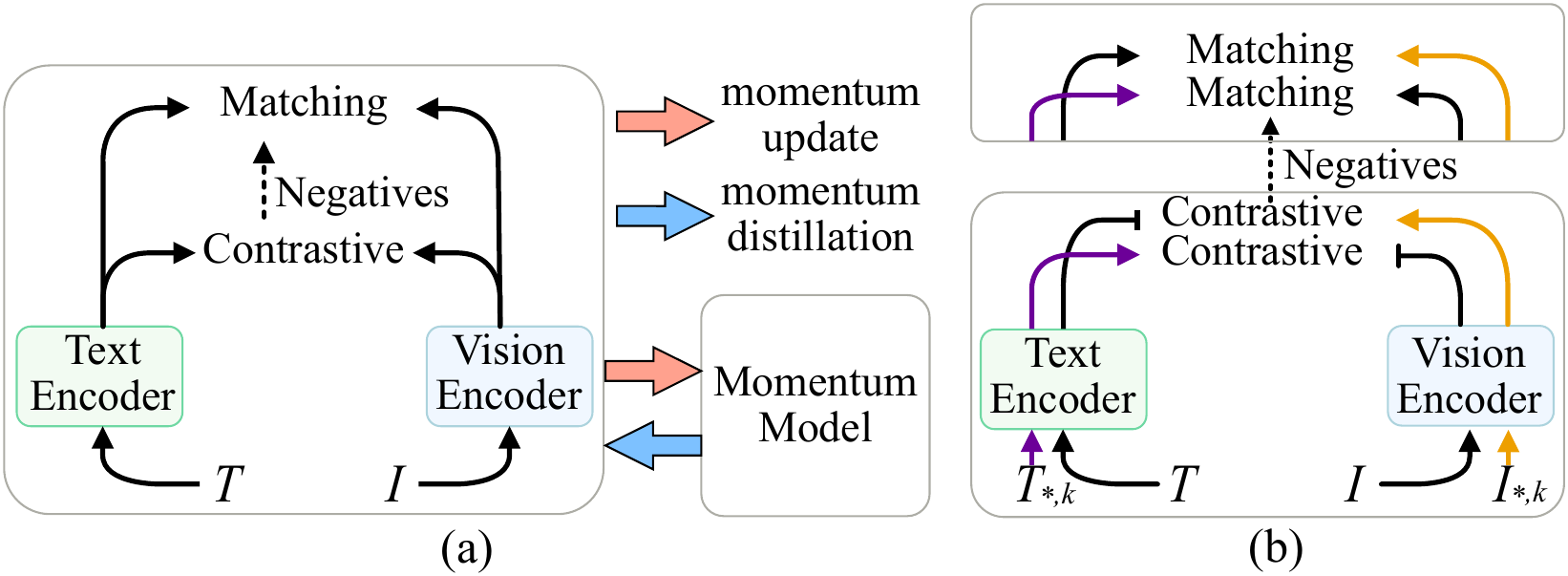}
    \vspace{-0.6cm}
    \caption{Team~[\textcolor{robo_blue}{rhao\_hur}]'s hierarchical contrastive learning modules. Detailed architecture of Region-Global ITC and ITM components for multi-granularity alignment.}
  \label{fig:track4_rhao_hur_module}
\end{figure}

\begin{figure*}[t]
    \centering
    \includegraphics[width=\linewidth]{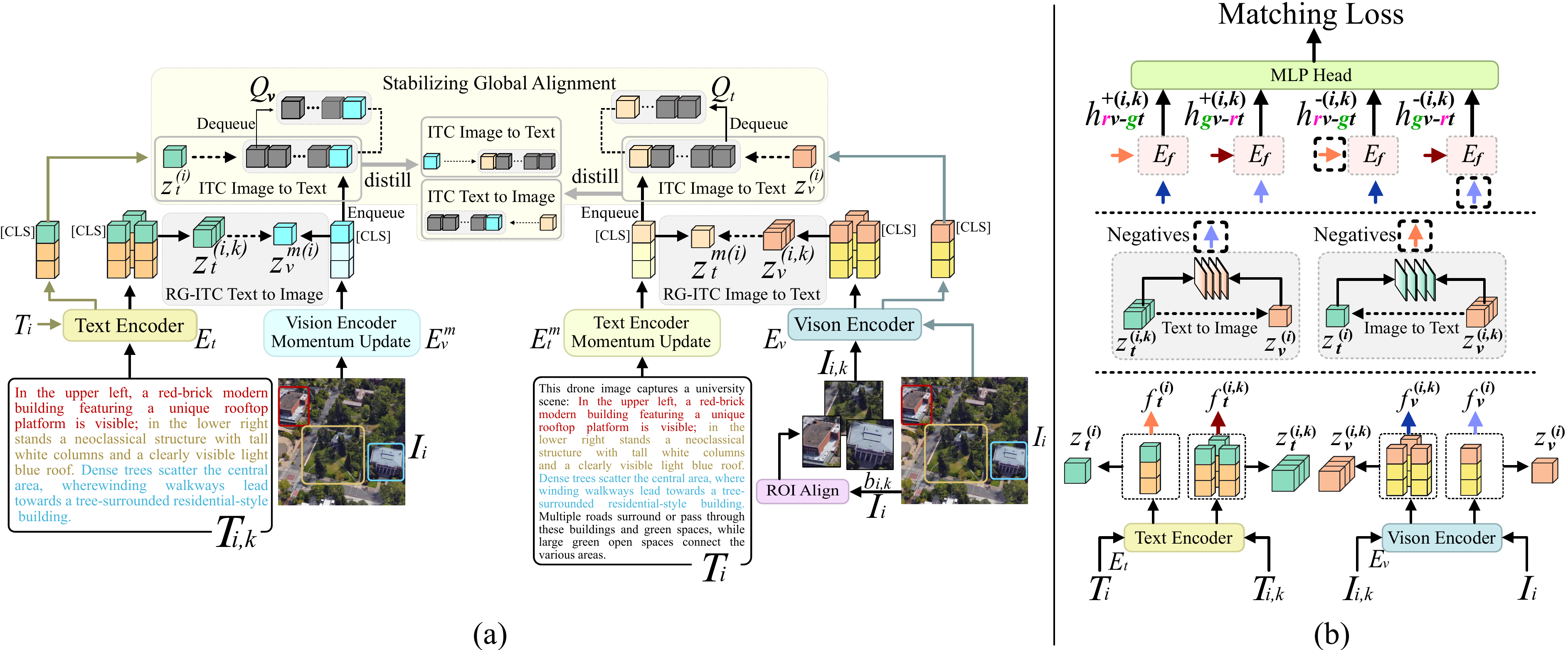}
    \vspace{-0.6cm}
    \caption{Team~[\textcolor{robo_blue}{rhao\_hur}]'s HCCM architecture overview. (a) Integration of global alignment ($\mathcal{L}_{ITC}$) with Region-Global Image-Text Contrastive Learning ($\mathcal{L}_{RG-ITC}$). (b) Region-Global Image-Text Matching Learning ($\mathcal{L}_{RG-ITM}$) with hard negative mining.}
  \label{fig:track4_rhao_hur_framework}
\end{figure*}

\noindent\textbf{\faLightbulbO~Key Innovations:}
\begin{itemize}
    \item \textit{Region-global image-text contrastive learning (RG-ITC):} Models semantic associations between local image regions and the global textual context (and vice versa), enabling the model to understand hierarchical relationships without relying on explicit entity segmentation or rigid part-whole constraints.
    
    \item \textit{Region-global image-text matching (RG-ITM):} Introduces a complementary consistency objective that evaluates whether local features align semantically with their corresponding global representations across modalities, reinforcing cross-modal coherence for compositional scene understanding.
    
    \item \textit{Momentum contrast and distillation (MCD):} Stabilizes training by maintaining momentum queues for negative sampling and generating soft target distributions that mitigate noise from ambiguous or incomplete textual descriptions, ensuring robust cross-modal alignment under real-world drone conditions.
\end{itemize}

\noindent\textbf{\faGear~Implementation Details:}\\
The framework builds upon the XVLM architecture, employing dual encoders for image and text modalities and a fusion encoder for fine-grained interactions. Region patches are extracted via ROI Align, and local-global pairs are constructed dynamically during training. The model is optimized using a multi-task loss combining ITC, ITM, RG-ITC, RG-ITM, and bounding-box regression objectives, weighted as $(w_{itc}, w_{itm}, w_{rg-itc}, w_{rg-itm}, w_{box}) = (0.25, 1.0, 0.25, 0.5, 0.1)$. Training is conducted on the GeoText-1652 dataset for six epochs with AdamW ($3\times10^{-5}$, weight decay 0.01) and a batch size of 24. Empirically, HCCM achieves state-of-the-art performance on GeoText-1652, reaching 28.8\% Recall@1 (image retrieval) and 14.7\% Recall@1 (text retrieval), outperforming prior methods such as HyCoCLIP and GeoText-1652 by a significant margin. On the ERA benchmark, the model attains 39.93\% mean Recall in zero-shot evaluation, surpassing all fine-tuned competitors, demonstrating superior compositional generalization. Visualization with GradCAM further confirms that HCCM grounds fine-grained entities (\eg, ``blue dome", ``solar panels") and relational structures (``buildings surrounded by roads") more accurately than global-only models. The results establish HCCM as a robust and interpretable solution for natural language-guided drone navigation, effectively bridging global alignment and local semantic reasoning.

\subsubsection{Team~ [\textcolor{robo_blue}{Xiaomi EV-AD VLA}]}
This team proposed a Caption-Guided Retrieval System (CGRS), which is a two-stage, coarse-to-fine retrieval framework that leverages VLMs to bridge the semantic gap between natural language descriptions and drone-view imagery. Their approach targets the long-standing issue in cross-modal drone navigation: while baseline models can achieve strong coarse alignment between images and text, they often lack fine-grained semantic precision, particularly in complex aerial scenes with repetitive or ambiguous structures. CGRS addresses this by introducing a caption-driven semantic refinement stage, transforming cross-modal retrieval into a more interpretable text-to-text similarity task.

\begin{figure*}[t]
    \centering
    \includegraphics[width=\linewidth]{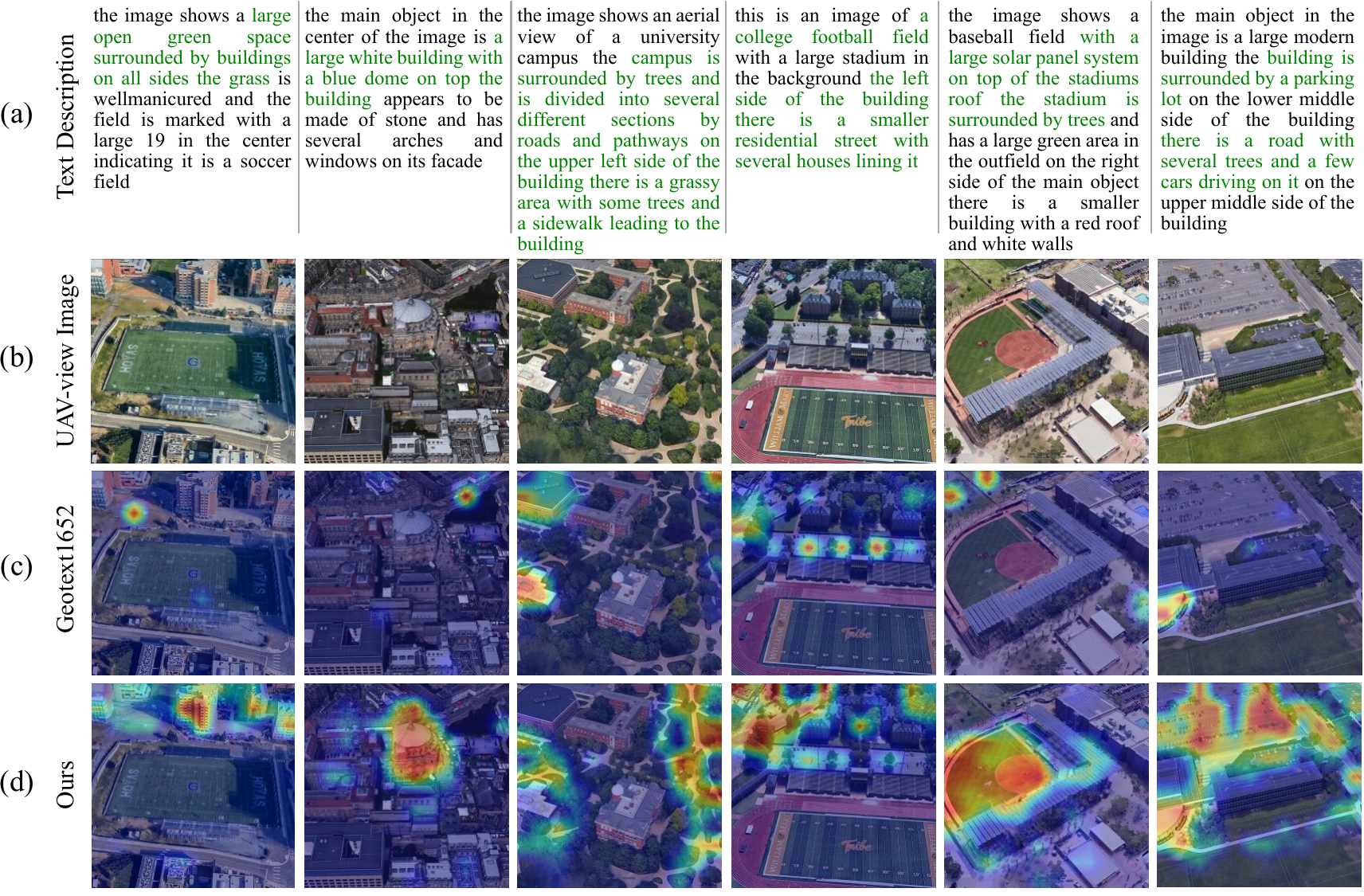}
    \vspace{-0.6cm}
    \caption{Team~[\textcolor{robo_blue}{rhao\_hur}]'s activation map visualization. (a) Text descriptions with key phrases highlighted. (b) UAV-view images. (c) Activation maps from GeoText-1652 baseline. (d) Activation maps from HCCM, showing improved grounding of fine-grained entities.}
    \label{fig:track4_rhao_hur_vis}
\end{figure*}

\noindent\textbf{\faLightbulbO~Key Innovations:}
\begin{itemize}
    \item \textit{Two-stage coarse-to-fine retrieval:} The system first performs a coarse retrieval using the GeoText-1652 baseline to select the top-20 candidate images for each query. These candidates are then refined through VLM-generated captions that encode spatial and semantic attributes, enabling fine-grained reranking.
    
    \item \textit{Caption-guided reranking via text-to-text similarity:} Instead of direct visual-text matching, CGRS generates detailed captions for each candidate using GPT-4o and compares them with the original query using a sentence encoder. This effectively transforms the retrieval problem into a semantic similarity comparison between natural language pairs, enhancing discriminative accuracy.
    
    \item \textit{Semantic fusion of coarse and fine retrieval signals:} The final ranking integrates visual and textual similarity scores through a weighted combination strategy, balancing geometric correspondence from the baseline with the semantic richness of caption-based reasoning.
\end{itemize}

\noindent\textbf{\faGear~Implementation Details:}\\
The framework builds upon the GeoText-1652~\cite{chu2024geotext-1652} baseline, employing a Swin Transformer for visual encoding and BERT for textual representation. The coarse model is trained for five epochs using AdamW ($3\times10^{-5}$) with spatial matching weight $\lambda=0.1$. For fine-grained refinement, captions are generated offline using GPT-4o with structured prompts emphasizing object layout, landmark relations, and spatial positions (\eg, ``center", ``top right"). Each caption averages 120–150 words, providing rich semantic cues. A BERT-based sentence encoder computes similarity between the query and generated captions, and a hybrid weight $\alpha=0.3$ fuses coarse and fine similarities. On the official RoboSense 2025 Track 4 benchmark, CGRS achieves R@1 = 31.33\%, R@5 = 49.09\%, and R@10 = 57.15\%, outperforming the baseline by +5–8\% across all metrics and securing second place among eight participating teams. Qualitative analysis shows that caption-guided reranking enhances semantic interpretability and effectively distinguishes visually similar yet semantically distinct aerial scenes. The results highlight the potential of combining retrieval-based pipelines with VLM-driven reasoning for robust and explainable cross-modal drone navigation.

\begin{figure*}[t]
    \centering
    \includegraphics[width=\textwidth]{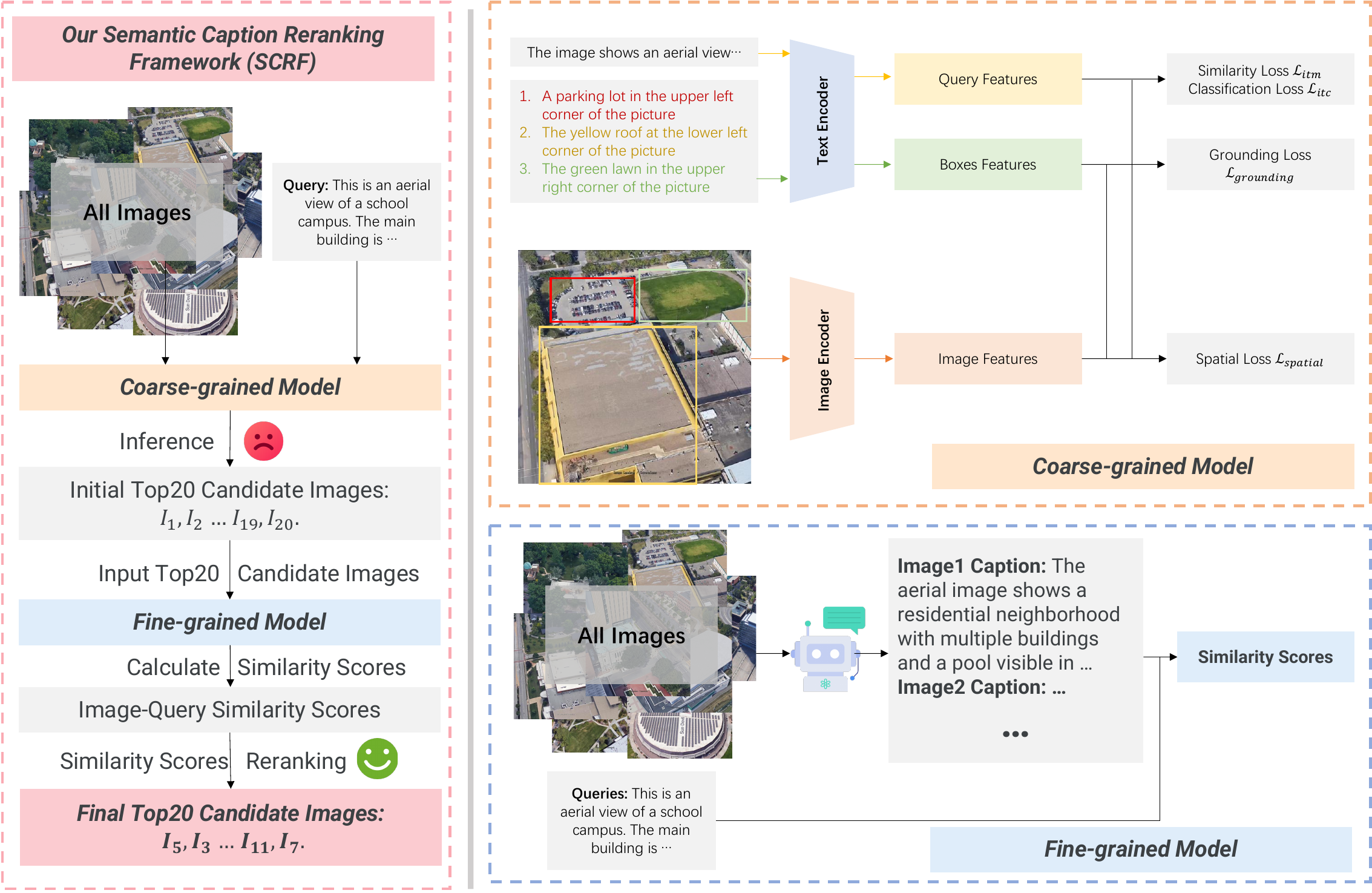}
    \vspace{-0.6cm}
    \caption{Team~[\textcolor{robo_blue}{Xiaomi EV-AD VLA}]'s Caption-Guided Retrieval System. The coarse-grained model retrieves top-20 candidates using the GeoText-1652 baseline, then the fine-grained model generates VLM captions and performs semantic reranking via text-to-text similarity.}
    \label{fig:track4_XiaomiEV_framework}
\end{figure*}
    
\begin{figure*}[t]
    \centering
    \includegraphics[width=\textwidth]{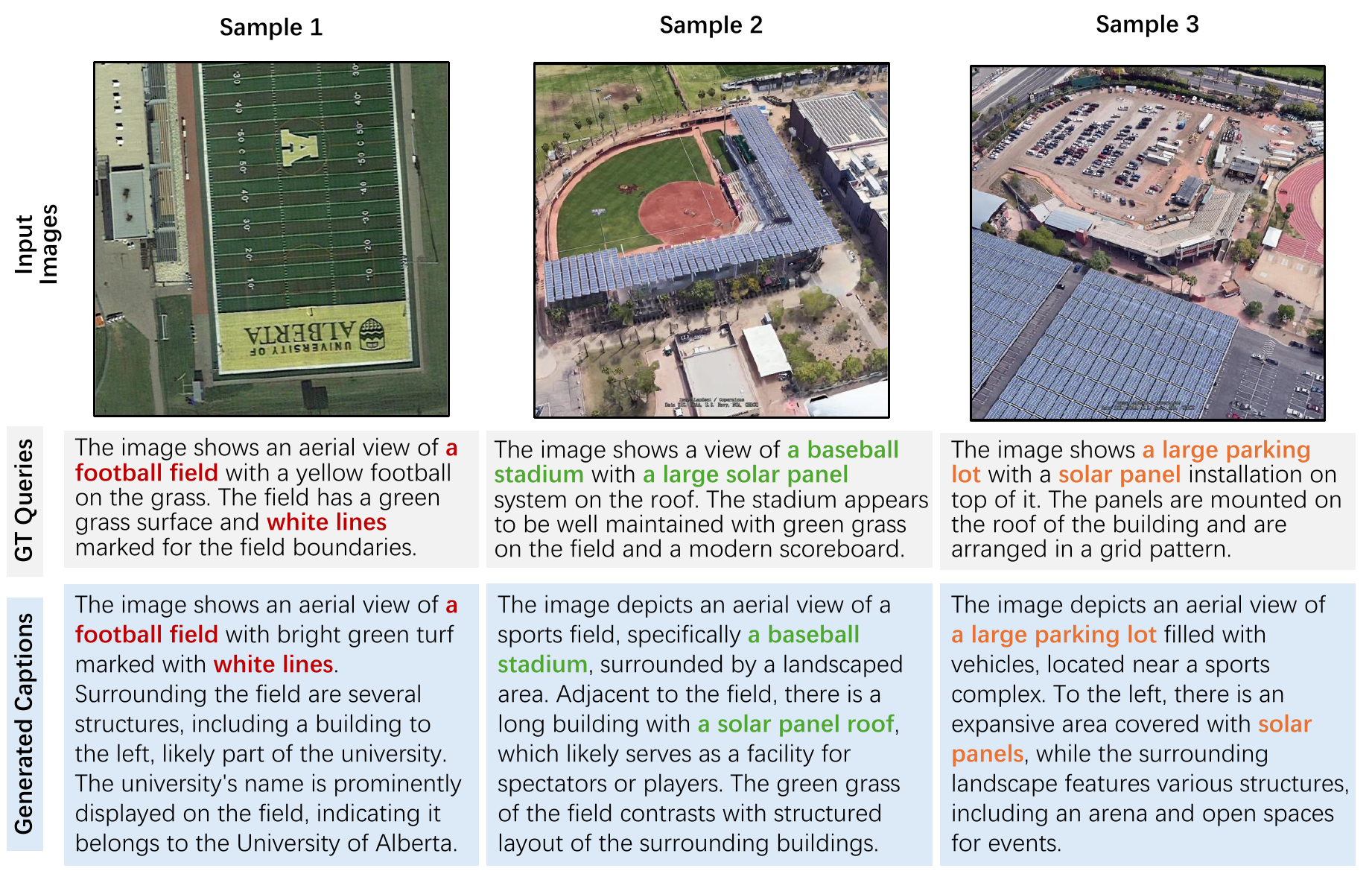}
    \vspace{-0.6cm}
    \caption{Team~[\textcolor{robo_blue}{Xiaomi EV-AD VLA}]'s qualitative retrieval results. Examples demonstrating caption-guided re-ranking's ability to distinguish semantically similar but contextually distinct aerial scenes.}
    \label{fig:track4_XiaomiEV_vis}
\end{figure*}

\subsubsection{Summary \& Discussion of Track 4}
This track investigates cross-modal drone navigation, where models must align natural language descriptions with aerial, satellite, and ground-view imagery. The task poses two central challenges: the severe domain gap between distinct viewpoints and modalities, and the need for fine-grained semantic understanding that connects global layout with localized visual cues. Track 4 evaluates retrieval systems that perform natural language–guided localization across platforms, emphasizing both precision and robustness under realistic visual corruptions and varying resolutions.

Across all submissions, two major methodological trends emerged. The first centers on \textit{hierarchical semantic alignment}, which integrates both global and regional cues to improve fine-grained cross-view matching. Advanced frameworks adopted multi-granularity contrastive or matching objectives to capture compositional relationships between local image regions and global textual semantics, thereby improving interpretability and resilience to spatial misalignment. The second trend is \textit{parameter-efficient and modular modeling}, where lightweight adaptation strategies, such as expert routing, domain-aligned preprocessing, and caption-guided refinement, achieve substantial performance gains without retraining large encoders. These approaches highlight that careful specialization and domain alignment can outperform brute-force scaling for cross-modal navigation.

Quantitatively, all top-performing systems achieved notable improvements over the GeoText-1652~\cite{chu2024geotext-1652} baseline, with relative gains of 5–15 \% in Recall@1 and consistent robustness under zero-shot evaluation. Methods that explicitly modeled local-global semantics and applied domain-aware data preprocessing delivered the strongest cross-view generalization, demonstrating that fine-grained language grounding is crucial for drone-level perception.  

Looking forward, future directions include expanding cross-modal grounding to 3D and temporal domains, enabling drones to interpret dynamic instructions and reason over sequential visual observations. Incorporating multi-sensor fusion (\eg, LiDAR–camera–language) and grounding under partial observability may further bridge the gap between visual retrieval and embodied drone navigation. Overall, Track 4 underscores that robust cross-modal alignment arises not only from high-capacity models, but from hierarchical reasoning and domain-aware design that allow drones to “understand” language in the wild.

\subsection{Track 5: Cross-Platform 3D Object Detection}
This section introduces the key innovations and implementation details of the three winning solutions in Track 5.
\subsubsection{Team~ [\textcolor{robo_blue}{Visionary}]}
This team introduced DataEngine, a unified pre-training and viewpoint normalization framework for cross-platform 3D object detection. The approach targets one of the fundamental challenges in LiDAR perception -- robustly transferring detection performance across heterogeneous robotic platforms such as vehicles, drones, and quadrupeds. Differences in LiDAR pose, scanning density, and motion patterns often cause geometric inconsistencies that severely limit cross-platform generalization. DataEngine addresses these issues by combining large-scale pre-training on diverse datasets, canonical viewpoint alignment, and test-time inference stabilization. Together, these components enable a single model to achieve consistent detection accuracy under widely varying viewpoints and sensing geometries. The proposed framework secured first place in both competition phases, achieving the strongest overall generalization among all submissions.

\noindent\textbf{\faLightbulbO~Key Innovations:}
\begin{itemize}
    \item \textit{Unified large-scale pre-training:} The model is pre-trained on a comprehensive corpus of six major LiDAR detection datasets (Waymo, nuScenes, ONCE, Pandaset, Lyft, and Argoverse2), collectively covering a wide range of sensor hardware and calibration settings. Unified class mapping ensures consistent label semantics, while range-adaptive masking prevents bias toward specific sensing configurations.
    
    \item \textit{Viewpoint normalization via ground-plane canonicalization:} A robust RANSAC-based module estimates the ground plane from each input scene and transforms all point clouds into a canonical coordinate frame aligned to the horizon. This normalization removes platform-specific biases, ensuring that identical objects are represented consistently regardless of sensor orientation or mounting height.
    
    \item \textit{Test-time augmentation and weighted box fusion:} During inference, multiple geometric augmentations -- rotations, flips, and scaling -- are applied, and predictions are consolidated using Weighted Box Fusion (WBF) to improve spatial precision and detection reliability under extreme viewpoint variations.
\end{itemize}

\noindent\textbf{\faGear~Implementation Details:}\\
The framework is implemented using the LION backbone, a window-based linear RNN detector built upon RetNet~\cite{sun2023retnet}. LION captures long-range spatial dependencies via efficient state-space modeling while maintaining scalability for large-scale pre-training. Training follows a two-stage process: six epochs of pre-training on the unified dataset collection, followed by two epochs of fine-tuning on the RoboSense Track 5 dataset using AdamW ($2\times10^{-4}$) and cosine learning rate scheduling. The system is trained on 8$\times$A100 GPUs and fine-tuned on 4$\times$RTX 3090s with mixed precision for efficiency.  Quantitatively, DataEngine achieves a \textbf{phase-1 mAP of 66.94} and \textbf{phase-2 mAP of 58.54}, outperforming all competitors by a clear margin. Ablation analysis shows additive gains from fine-tuning (+2.99) and viewpoint normalization (+5.09), confirming the synergy between canonicalization and large-scale pre-training. Qualitative evaluations further reveal that the model produces stable detections even under severe occlusion and elevation changes. Collectively, these results establish DataEngine as a scalable, platform-agnostic solution for LiDAR perception, illustrating the effectiveness of unified pre-training and geometric normalization in achieving cross-domain generalization for real-world robotics.

\begin{figure*}[t]
    \centering
    \includegraphics[width=\textwidth]{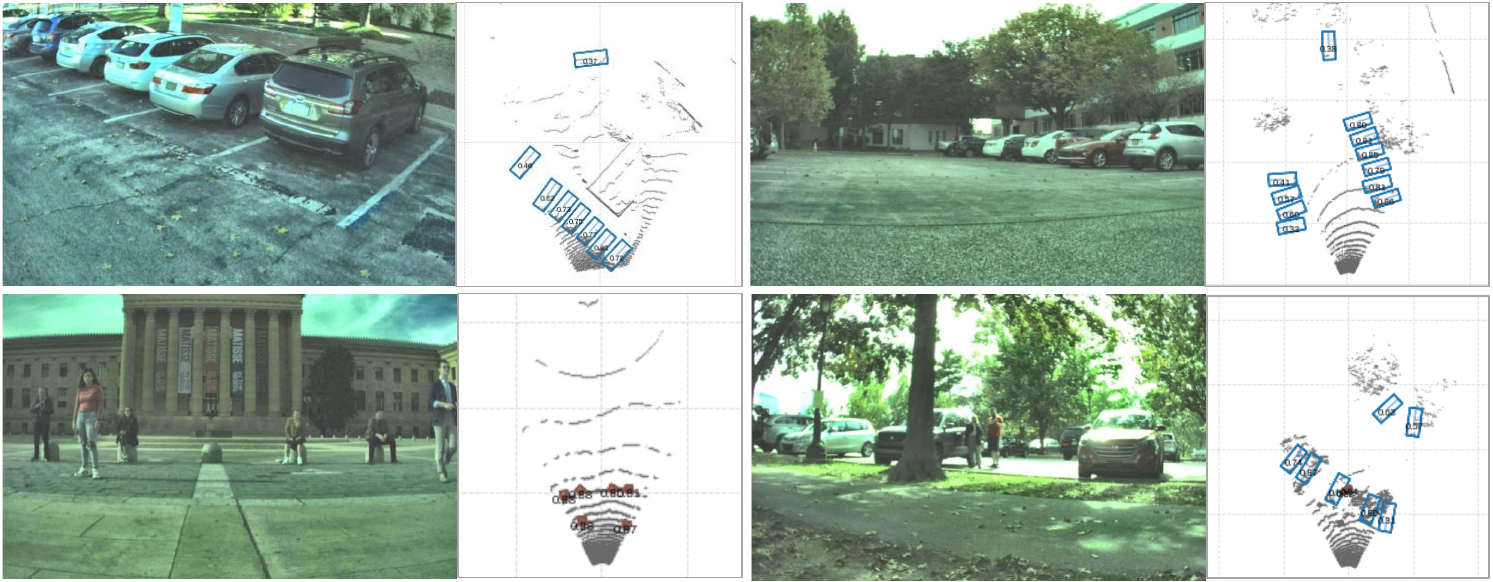}
    \vspace{-0.6cm}
    \caption{Team~[\textcolor{robo_blue}{Visionary}]'s qualitative detection results. Examples demonstrating DataEngine's robust performance across vehicle, drone, and quadruped platforms under varying viewpoints and occlusion conditions.}
    \label{fig:track5_Visionary_example}
\end{figure*}

\begin{figure*}[t]
    \centering
    \includegraphics[width=\textwidth]{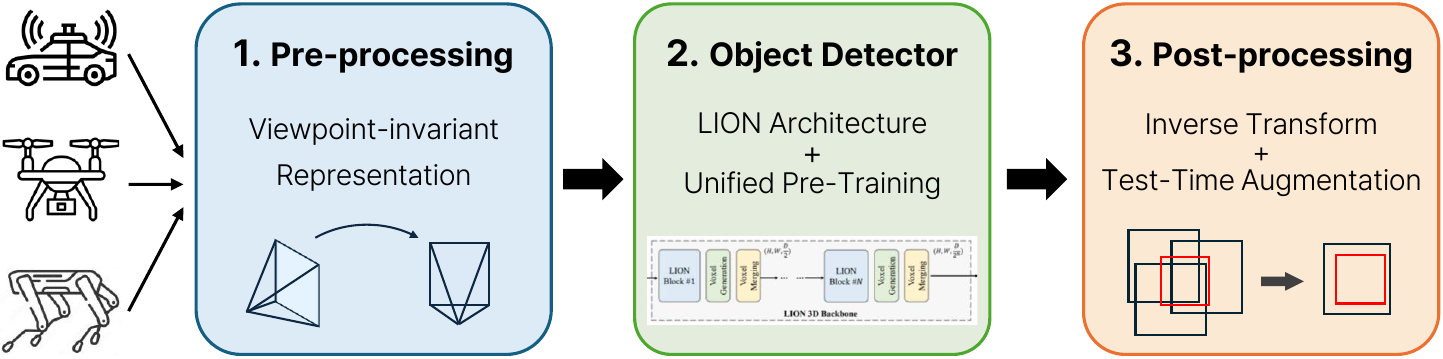}
    \vspace{-0.6cm}
    \caption{Team~[\textcolor{robo_blue}{Visionary}]'s three-stage DataEngine pipeline. Stage (1) Pre-processing: viewpoint normalization via ground-plane canonicalization. Stage (2) Detection: generalized LION detector processing. Stage (3) Post-processing: coordinate transformation and Test-Time Augmentation fusion.}
    \label{fig:track5_Visionary_method}
\end{figure*}

\subsubsection{Team~ [\textcolor{robo_blue}{Point Loom}]}
This team proposed a comprehensive framework for cross-platform 3D object detection that integrates \textit{physical-aware data augmentation} and \textit{class-specific model ensembling}. The approach addresses major challenges in adapting LiDAR detectors trained on vehicle-mounted sensors to unseen platforms such as drones and quadrupeds, where viewpoint, scanning height, and scene composition differ drastically. The framework explicitly models these geometric and semantic shifts through data-centric augmentations and category-tailored adaptation strategies. By combining scene-level resampling, viewpoint perturbation, and height-aware inference, the method enhances both geometric adaptability and semantic robustness across heterogeneous sensing domains.

\begin{figure*}
    \centering
    \includegraphics[width=0.85\linewidth]{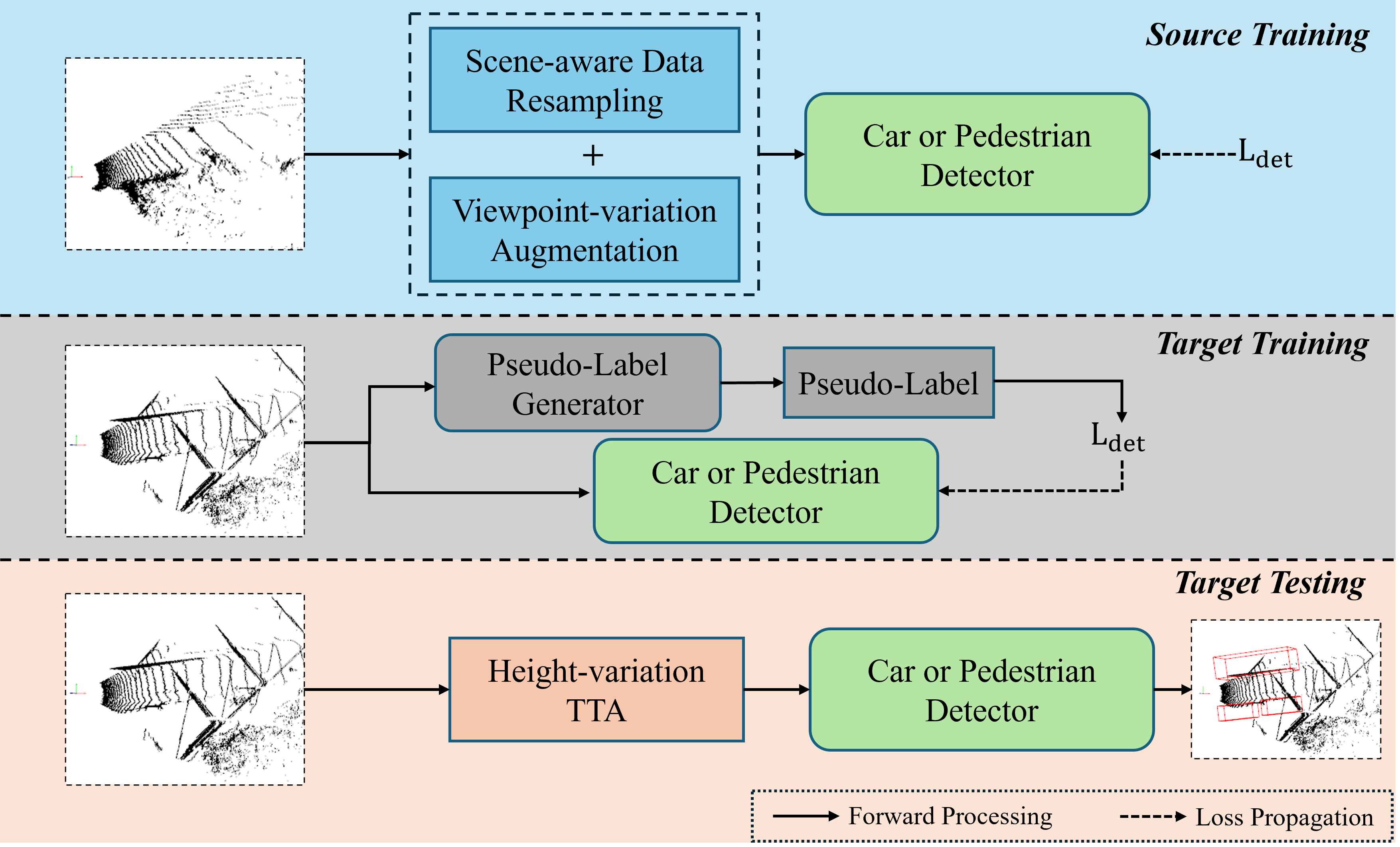}
    \vspace{-0.2cm}
    \caption{Team~[\textcolor{robo_blue}{Point Loom}]'s cross-platform detection framework. Source training employs scene resampling and viewpoint augmentation for category-specific detectors. Target training uses pseudo-label self-training. Target testing applies height-variation TTA.}

    \label{fig:track5_Point_Loom_framework}
\end{figure*}

\noindent\textbf{\faLightbulbO~Key Innovations:}
\begin{itemize}
    \item \textit{Physical-aware data augmentation:} Three augmentation modules -- scene-aware data resampling, viewpoint-variation augmentation, and height-variation test-time augmentation -- jointly simulate cross-platform disparities. Scene resampling enriches underrepresented environments (\eg, parking lots), while viewpoint perturbation randomly rotates training samples along the x/y axes to model cross-platform perspective shifts. Height-aware test-time augmentation further mitigates systematic biases induced by different LiDAR mounting heights.
    
    \item \textit{Class-specific model ensembling:} Instead of a single unified detector, separate models are trained for distinct object categories (\eg, car and pedestrian), each optimized with class-specific augmentations. This targeted specialization enhances generalization to unseen domains, with results later fused through non-maximum suppression to form the final detection output.
    
    \item \textit{Self-training–based adaptation:} During target-domain adaptation, the framework employs the ST3D~\cite{yang2021st3d} paradigm to iteratively refine pseudo labels on unlabeled data, achieving progressive alignment of source and target feature distributions without supervision.
\end{itemize}

\noindent\textbf{\faGear~Implementation Details:}\\
The method builds on the OpenPCDet~\cite{openpcdet2020} framework, adopting PV-RCNN~\cite{shishaoshuai2020pv} for Phase 1 (vehicle → drone) and PV-RCNN++~\cite{shi2023pv} for Phase 2 (vehicle → quadruped). The Adam one-cycle optimizer is used with learning rates of 0.01 (source) and 0.0005 (target). Data augmentations are applied per class: cars use both scene resampling and viewpoint variation, while pedestrians rely on viewpoint diversity alone. Height-aware test-time augmentation employs vertical offsets of $\{-0.4, 0.0, 0.4\}$ m for cars and $\{-0.6, 0.0, 0.6\}$ m for pedestrians. Quantitatively, the framework attains 62.25\% AP in Phase 1 (vehicle → drone) and an average of 55.66\% AP across classes in Phase 2 (vehicle → quadruped), exceeding the official baselines by over 19 points. Ablation studies confirm that the proposed augmentations and self-training contribute complementary improvements, while the class-specific ensemble provides an additional robustness boost under cross-platform conditions. These results demonstrate that modeling physical geometry and semantic diversity jointly offers a scalable path toward platform-agnostic LiDAR perception.

\subsubsection{Team~ [\textcolor{robo_blue}{DUTLu\_group}]}
This team proposed a domain-adaptive 3D detection framework that enhances cross-platform generalization through tailored data augmentation and self-training with pseudo-labels. Built upon PV-RCNN++~\cite{shi2023pv}, the approach mitigates geometric distribution shifts across vehicle, drone, and quadruped LiDAR platforms, where sensor viewpoint, vibration patterns, and spatial density differ significantly. The framework combines \textit{Cross-Platform Jitter Alignment (CJA)} to simulate platform-specific motion perturbations and \textit{ST3D-based self-training} to progressively adapt the detector to unlabeled target domains. By integrating these modules with a redesigned anchor-based proposal head, the system achieves strong cross-platform robustness without requiring additional annotations.

\begin{figure*}
    \centering
    \includegraphics[width=\linewidth]{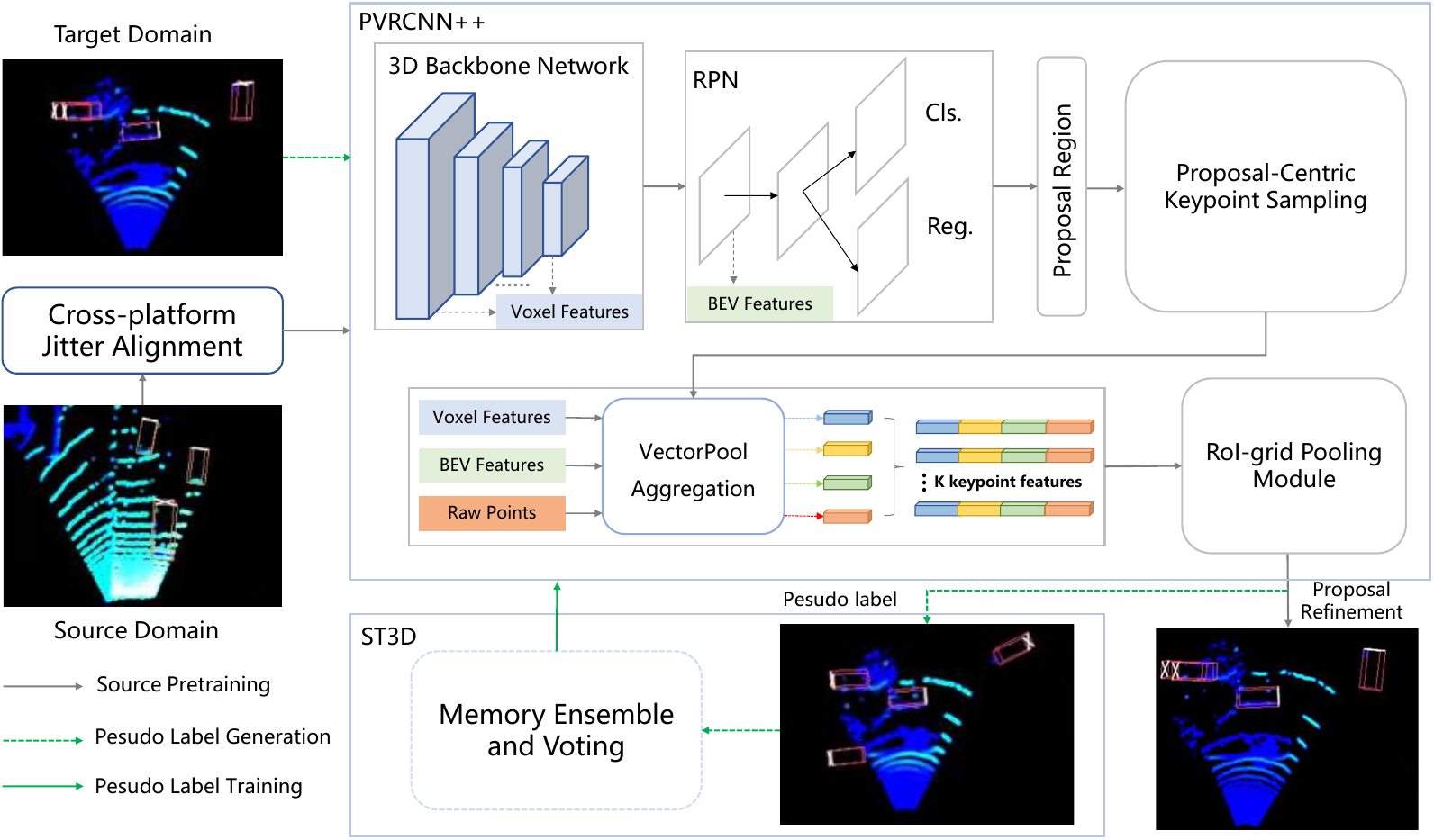}
    \vspace{-0.6cm}
    \caption{Team~[\textcolor{robo_blue}{DUTLu\_group}]'s domain-adaptive framework. Source domain processing with CJA augmentation and target domain pseudo-label refinement via ST3D module enable iterative cross-platform adaptation without additional annotations.}

    \label{fig:track5_DUTLu_group_framework}
\end{figure*}

\noindent\textbf{\faLightbulbO~Key Innovations:}
\begin{itemize}
    \item \textit{Cross-platform jitter alignment (CJA):} A data augmentation method that introduces controlled pitch–roll perturbations to vehicle-domain point clouds during pre-training, imitating the jitter patterns observed on drone and quadruped LiDAR setups. This alignment encourages the detector to learn viewpoint-invariant geometric features and improves generalization to unseen sensor poses.
    
    \item \textit{Iterative pseudo-label self-training:} Adopts the ST3D~\cite{yang2021st3d} paradigm to generate and refine pseudo-labels on unlabeled target-domain data. High-confidence detections from early iterations bootstrap later fine-tuning, progressively narrowing the inter-domain feature gap.
    
    \item \textit{AnchorHead-based proposal generation:} Replaces the original CenterHead in PV-RCNN++ with an AnchorHead containing multi-scale, multi-orientation priors. This modification alleviates center drift caused by sparse object observations at oblique viewing angles, improving proposal quality and recall in dynamic target domains.
\end{itemize}

\noindent\textbf{\faGear~Implementation Details:}\\
The framework is implemented with OpenPCDet~\cite{openpcdet2020} in PyTorch. Training proceeds in two stages: supervised source-domain pre-training followed by unsupervised domain adaptation via ST3D self-training. Both stages use the AdamW optimizer and OneCycle learning rate policy, with initial learning rates of 0.01 (source) and $1.5\times10^{-3}$ (target). For each training sample, pitch and roll jitter angles $(\Delta\theta, \Delta\phi)$ are uniformly sampled within predefined bounds and applied to both point clouds and annotations to preserve geometric consistency. During self-training, confidence thresholds are set to 0.7 (Phase 1) and class-specific 0.85/0.55 (Car/Pedestrian) in Phase 2.  

The final model achieves 62.67\% Car AP on the Phase 1 (vehicle → drone) target domain and 58.76\% Car AP / 49.81\% Pedestrian AP on Phase 2 (vehicle → quadruped), ranking third overall in Track 5. Ablation studies confirm that CJA alone improves Car AP@0.5 by +14.5 points, while combining CJA with ST3D yields an additional +11.0 gain, highlighting the complementary nature of geometric augmentation and iterative adaptation. The results demonstrate that explicitly modeling motion-induced perturbations and leveraging pseudo-label supervision form a practical, annotation-free pathway toward platform-agnostic 3D detection in real-world robotics.

\subsubsection{Team~ [\textcolor{robo_blue}{Hunter}]}
This team introduced SegSy3D (Segmentation-Guided Self-Training and Model Synergy), a cross-platform 3D object detection framework designed to mitigate large geometric and distributional gaps across vehicle, drone, and quadruped LiDAR platforms. Built upon Voxel R-CNN with Gaussian Blobs (GBlobs)~\cite{malic2025gblobs}, the framework unifies segmentation-guided pseudo-label refinement and model-synergy–based test-time adaptation to achieve robust unsupervised domain adaptation (UDA) without any target-domain annotations. By combining semantic cues from segmentation with temporal model ensembling, SegSy3D achieves strong cross-platform generalization for 3D object detection.

\begin{figure}
    \centering
    \includegraphics[width=\linewidth]{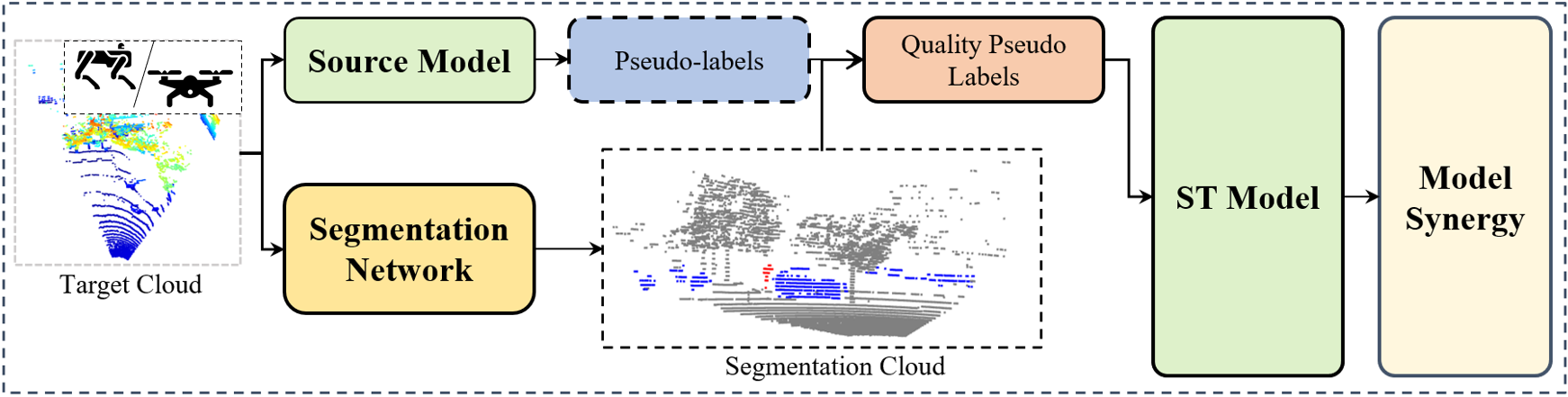} 
    \vspace{-0.6cm}
    \caption{Team~[\textcolor{robo_blue}{Hunter}]'s SegSy3D pipeline overview. Source detector generates pseudo-labels on target data, segmentation network refines them, refined labels drive self-training adaptation, and Model Synergy assembles historical checkpoints for robust inference.}

  \label{fig:track5_Hunter_network_sample}
\end{figure}

\begin{figure*}
    \centering
    \includegraphics[width=\linewidth]{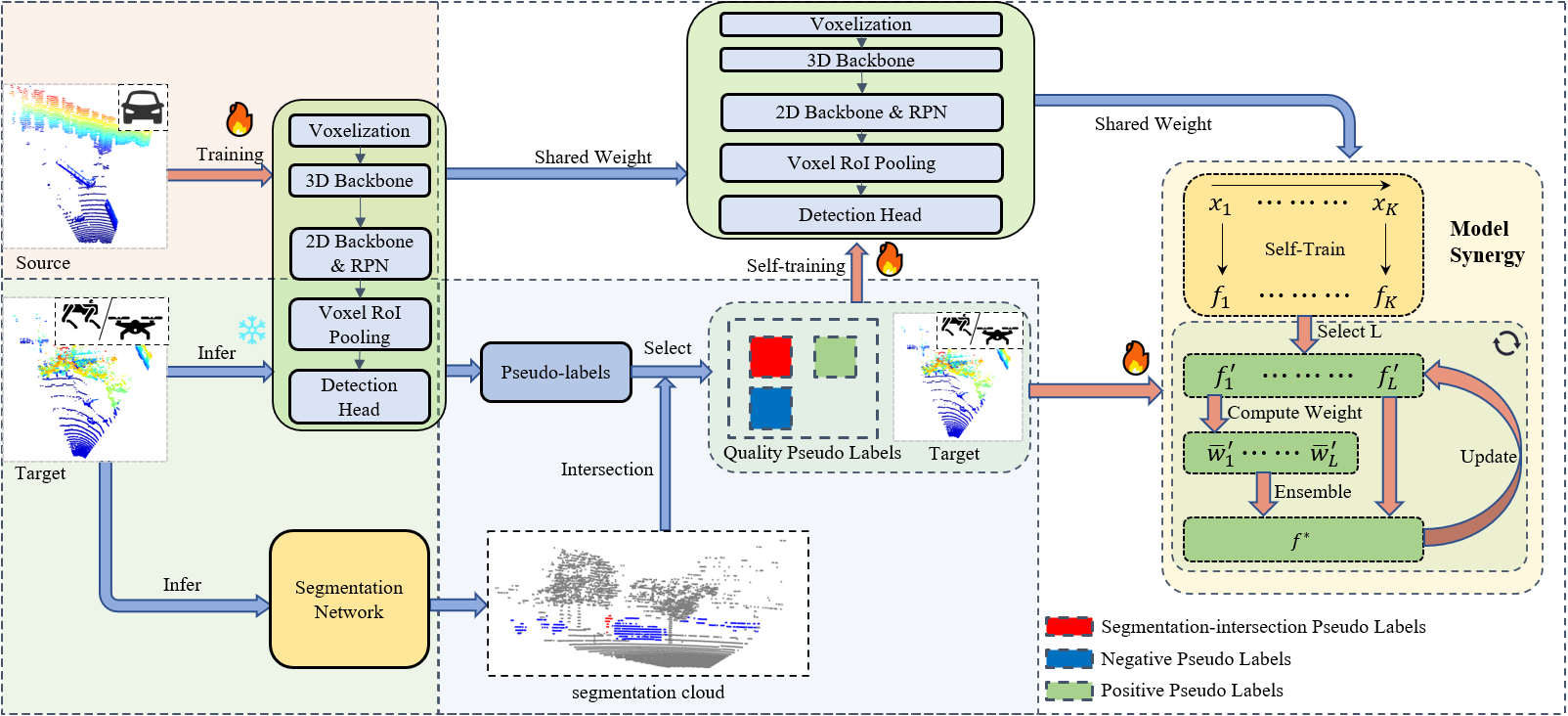} 
    \vspace{-0.6cm}
    \caption{Team~[\textcolor{robo_blue}{Hunter}]'s detailed SegSy3D architecture. The framework couples segmentation-guided pseudo-label refinement with Model Synergy (MOS) for multi-temporal adaptation and checkpoint fusion.}

  \label{fig:track5_Hunter_network}
\end{figure*}

\begin{figure*}
    \centering
    \includegraphics[width=\linewidth]{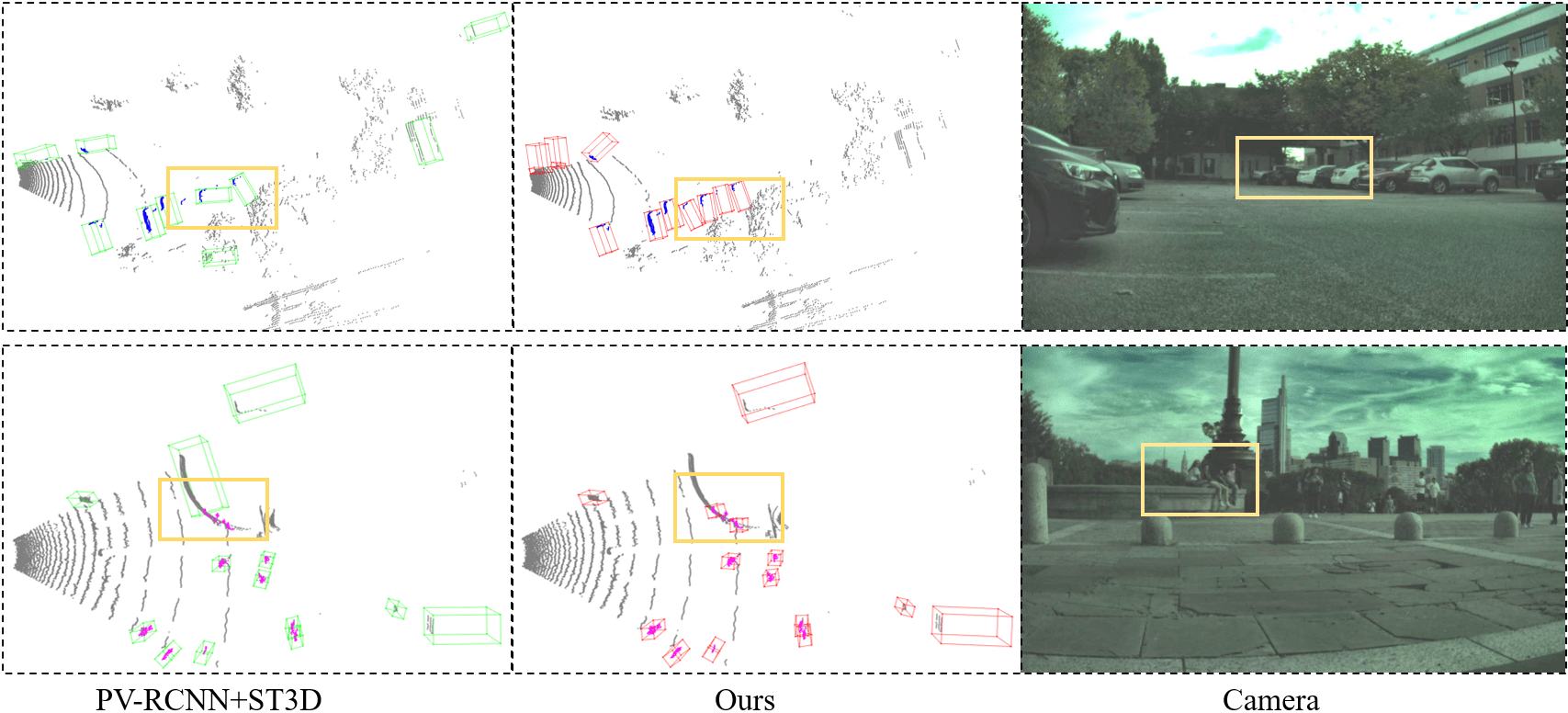} 
    \vspace{-0.6cm}
    \caption{Team~[\textcolor{robo_blue}{Hunter}]'s qualitative comparison on Phase~2 Quadruped domain. Left: baseline. Center: SegSy3D. Right: camera view. Yellow boxes highlight regions of interest; green/red boxes denote predictions; blue/magenta points show segmented cars/pedestrians.}

  \label{fig:track5_Hunter_label_show}
\end{figure*}

\noindent\textbf{\faLightbulbO~Key Innovations:}
\begin{itemize}
    \item \textit{Segmentation-guided pseudo-label refinement:} A source-trained segmentation model generates class-specific masks that intersect with detector predictions to refine pseudo-labels, recovering geometry-consistent low-confidence objects and filtering false positives. This dual-source label generation provides a more reliable supervision signal for self-training under domain shifts.
    
    \item \textit{Progressive self-training with semantic filtering:} Built on ST3D~\cite{yang2021st3d}, SegSy3D iteratively adapts the detector using refined pseudo-labels, enforcing geometric consistency and semantic balance across target domains. This progressive adaptation narrows the cross-platform feature gap and stabilizes learning.
    
    \item \textit{Model Synergy for test-time adaptation:} At inference, historical checkpoints with diverse representations are dynamically weighted and fused into a unified meta-model. This synergy captures complementary long-term knowledge, enabling multi-temporal adaptation and improved robustness to platform-specific variations.
\end{itemize}

\noindent\textbf{\faGear~Implementation Details:}\\
SegSy3D is implemented in PyTorch based on OpenPCDet~\cite{openpcdet2020}, with Voxel R-CNN~\cite{deng2021voxel} and GBlob-sVFE~\cite{malic2025gblobs} backbones. Source-domain training uses 30 epochs with Adam and a OneCycle scheduler (initial LR = 0.01), followed by 16 epochs of self-training on unlabeled target data at a reduced LR of 0.005. Input point clouds are voxelized at (0.1, 0.1, 0.15) m, and confidence thresholds of 0.7–0.85 control pseudo-label selection. Quantitatively, the framework achieves 56.14 AP $_{40}^{0.5}$ for Car, 55.07 AP $_{40}^{0.5}$ for Pedestrian, and an overall mAP of 55.61, surpassing all self-training baselines and reducing the cross-platform performance gap by over 25 points compared with direct source-only transfer. Ablation studies confirm complementary benefits: segmentation guidance improves pseudo-label precision and recall, while Model Synergy yields additional robustness under dynamic viewpoints. Collectively, SegSy3D demonstrates that integrating geometric semantics with temporal model fusion forms a scalable path toward annotation-free, platform-agnostic 3D perception.

\subsubsection{Team~ [\textcolor{robo_blue}{TeamArcN}]}
This team proposed UADA3D, an Unsupervised Adversarial Domain Adaptation framework for cross-platform 3D object detection. Their design targets the significant geometric and distributional gaps between vehicle-, drone-, and quadruped-mounted LiDAR systems, where sensor viewpoint, scan sparsity, and motion noise differ substantially. To overcome the limitations of pseudo-label–based approaches (\eg, ST3D++), which rely heavily on label quality and pre-trained teacher models, the proposed method instead aligns domain-invariant features through adversarial learning. By coupling the lightweight, point-based IA-SSD~\cite{zhang2022iassd} detector with the UADA3D adaptation module, this framework achieves efficient and stable domain transfer under large cross-platform variations.

\begin{figure*}[t]
    \centering
    \includegraphics[width=\linewidth]{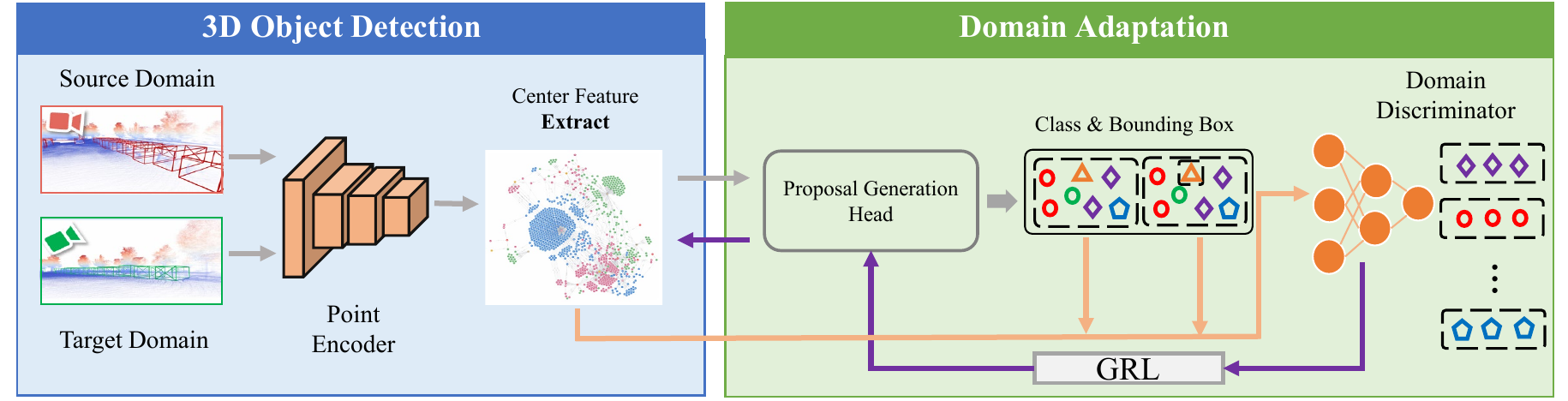}
    \vspace{-0.6cm}
    \caption{Team~[\textcolor{robo_blue}{TeamArcN}]'s UADA3D framework. The system combines 3D object detection with adversarial domain adaptation, where a discriminator (orange line) processes center features, bounding boxes, and classes, and GRL (purple line) enables domain-invariant feature learning.}

\label{fig:track5_TeamArcN_framework}  
\end{figure*}

\noindent\textbf{\faLightbulbO~Key Innovations:}
\begin{itemize}
    \item \textit{Adversarial domain alignment via Gradient Reversal Layer (GRL):} The framework introduces a domain discriminator that distinguishes between source and target features based on class-wise center embeddings, bounding box representations, and object categories. A GRL is used to backpropagate the discriminator loss, forcing the shared feature extractor to produce domain-invariant features and improving transferability across sensor viewpoints.
    
    \item \textit{Instance-aware feature masking:} IA-SSD’s point-based features are filtered through an instance-aware mask that preserves only foreground object features for domain discrimination. This minimizes background interference and enhances cross-domain feature alignment efficiency.
    
    \item \textit{Collaborative optimization of detection and adaptation:} Unlike decoupled two-stage UDA pipelines, the detection and adaptation losses are jointly optimized in a single training loop. The domain loss from UADA3D is backpropagated to the feature extractor and detection head of IA-SSD, forming a unified objective that balances detection accuracy with domain alignment.
\end{itemize}

\noindent\textbf{\faGear~Implementation Details:}\\
The model is built using IA-SSD~\cite{zhang2022iassd} as the backbone and OpenPCDet~\cite{openpcdet2020} as the training framework. Source-domain pre-training is performed on vehicle-platform data with Adam (LR = 0.01, batch size = 16, 60 epochs), followed by adversarial fine-tuning with UADA3D on unlabeled drone (Phase 1) and quadruped (Phase 2) data. The Gradient Reversal Layer (GRL) coefficient $\lambda$ is linearly annealed with $\lambda = \alpha(\frac{2}{1+e^{-\gamma p}} - 1)$, where $\alpha=0.1$ and $\gamma=10$. All training is conducted on a single NVIDIA RTX 4070 Super GPU under CUDA 11.8. Quantitatively, the model achieves 57.6\% Car AP@50 / 34.65\% Car AP@70 in Phase 1 (vehicle→drone) and 36.59\% Car AP@50 / 17.4\% Car AP@70 / 60.21\% Pedestrian AP@25 / 51.31\% Pedestrian AP@50 in Phase 2 (vehicle→quadruped), outperforming the baseline PV-RCNN + ST3D++ by 10–12 points on both classes. These results highlight the advantages of adversarial feature alignment over pseudo-labeling under large geometric shifts. The study demonstrates that coupling IA-SSD with UADA3D provides an efficient, scalable pathway for cross-platform 3D object detection, reducing reliance on annotation and improving adaptability to sparse or oblique LiDAR viewpoints.

\subsubsection{Summary \& Discussion of Track 5}
This track investigates the challenge of \textbf{cross-platform 3D object detection}, where LiDAR perception models must generalize across robotic platforms such as vehicles, drones, and quadrupeds. The task emphasizes robustness to drastic viewpoint, height, and motion differences that cause geometric distortions and distribution shifts. Participants were required to design frameworks capable of learning platform-invariant representations without access to target-domain labels, making this track a practical test of large-scale adaptability in 3D perception.

Two complementary research directions emerged.  
\textbf{First}, \textit{geometry-driven generalization} proved crucial for cross-platform robustness. High-performing methods adopted viewpoint-canonicalization or physical-aware augmentations to align scenes into consistent coordinate spaces, mitigating platform-specific biases. Ground-plane normalization, rotation perturbation, and jitter simulation were commonly employed to ensure viewpoint-invariant detection. These geometry-centric strategies directly address the spatial misalignment that arises when transferring models across vehicle, drone, and quadruped sensors.  

\textbf{Second}, \textit{domain-adaptive learning} played an equally vital role. Several approaches integrated unsupervised adaptation -- through self-training, pseudo-label refinement, or adversarial feature alignment -- to progressively bridge the distribution gap between source and target domains. Progressive self-labeling stabilized adaptation on unlabeled data, while adversarial objectives encouraged the detector to learn domain-invariant features. Other methods enhanced robustness via class-specific ensembling or segmentation-guided refinement, coupling semantic reasoning with geometric cues.

Overall, the leading frameworks consistently surpassed the PV-RCNN++ and BEVFusion-L baselines by large margins, achieving 10–20 point improvements in mAP across platforms. The results collectively reveal that cross-platform detection benefits most from the synergy of structural modeling and adaptive learning: canonicalizing geometry alone is insufficient without domain alignment, while self-training performs best when guided by geometric priors.  

Looking forward, future work may extend these ideas to unified multi-platform pre-training and multi-sensor fusion, where LiDAR, camera, and event data can be jointly aligned under shared spatial representations. The insights from Track 5 highlight that reliable 3D perception in heterogeneous robotic systems arises not from model scale alone, but from learning geometry and semantics that persist across platforms and motion domains.

\section{Discussions \& Future Directions}
\label{sec:discussions}

RoboSense 2025 highlights a central theme across all five tracks: modern perception systems can achieve strong accuracy in-distribution, yet their reliability still degrades noticeably under distribution shifts induced by sensor corruption, viewpoint changes, configuration mismatch, and platform diversity. While the top submissions substantially improved over the baselines, the winning solutions also reveal common design patterns and recurring bottlenecks. Below we summarize key observations and outline promising directions for future robustness-oriented robot sensing research.

\subsection{Cross-Track Observations}

\noindent\textbf{(1) Robustness often comes from training \emph{procedures} rather than architectures alone.}
Across the leaderboards, many leading approaches emphasize data-centric robustness strategies -- corruption-aware augmentation, stronger pretraining, improved sampling schedules, and better pseudo-label filtering -- which often yield larger gains than minor architectural changes. This suggests that robustness in embodied settings is frequently limited by how models are trained and evaluated, not only by model capacity.

\noindent\textbf{(2) Distribution shift is multi-factor and compound.}
RoboSense evaluates shifts that arise from different sources: visual corruption (Track~1), dynamic social interactions and partial observability (Track~2), sensor placement changes (Track~3), drastic cross-view semantics and scale mismatch (Track~4), and platform-induced geometry and motion differences (Track~5). In practice, these shifts are often compound. A major open challenge is to build systems that remain stable when multiple factors occur simultaneously, such as viewpoint changes under sensor degradation or cross-platform transfer under long-tail scenes.

\noindent\textbf{(3) Reliability requires \emph{calibration} and \emph{verification}, not only higher mean scores.}
Several top methods improve average metrics but can still exhibit brittle failure modes in rare cases: hallucinated answers under degraded inputs (Track~1), socially unacceptable shortcuts (Track~2), or unstable adaptation under noisy pseudo labels (Tracks~3/5). This motivates evaluation protocols that better capture tail risk, uncertainty calibration, and failure predictability, especially for safety-critical deployment.

\noindent\textbf{(4) Geometry remains a unifying bottleneck across modalities.}
Despite the diversity of tracks, a recurring challenge is robust geometric reasoning: localization under varying viewpoints and scales (Track~4), stable 3D detection under configuration/platform shifts (Tracks~3/5), and spatially grounded driving QA under corruptions (Track~1). Methods that incorporate geometry-aware representations, spatial priors, or structured alignment often demonstrate stronger transfer and stability.

\subsection{Future Directions}
\begin{itemize}
    \item \emph{Unified robustness protocols and compound shift benchmarks.}
    Future editions of RoboSense can further expand toward \emph{compound robustness} settings that jointly stress multiple shift factors. A unified protocol that standardizes corruption severity, cross-domain splits, and statistical significance testing would enable clearer attribution of improvements and stronger cross-track comparability.

    \item \emph{Reliable multimodal reasoning with uncertainty and abstention.}
    For language-grounded decision making and embodied interaction, robustness should include calibrated uncertainty, refusal/abstention mechanisms, and verifiable reasoning traces. Integrating uncertainty-aware decoding, self-consistency checks, and multimodal grounding constraints is a promising direction to reduce hallucination and improve trustworthiness.

    \item \emph{Sensor-agnostic representations and configuration-aware learning.}
    Tracks~3 and~5 suggest the importance of learning representations that are insensitive to sensor placement changes and platform-dependent geometry. Potential directions include explicit modeling of sensor extrinsics, canonicalization layers, simulation-based domain randomization, and self-supervised objectives that align observations across configurations without requiring labels.

    \item \emph{Stable adaptation under label scarcity.}
    Cross-domain adaptation remains limited by pseudo-label noise and optimization instability. Future work may benefit from more principled confidence modeling, robust self-training schedules, teacher-student frameworks with stronger regularization, and benchmarked protocols for preventing mode collapse during adaptation.

    \item \emph{Bridging perception, interaction, and downstream safety.}
    Finally, RoboSense motivates integrating robustness evaluation beyond isolated tasks, toward downstream behaviors and safety outcomes. For example, socially compliant navigation and driving QA can be coupled with safety-critical constraints and scenario-level evaluation, moving from static prediction metrics toward behaviorally meaningful robustness guarantees.
\end{itemize}

\section{Conclusion}
\label{sec:conclusion}

This report summarizes the RoboSense 2025 Challenge, a large-scale benchmark at IROS 2025 that advances robustness and adaptability in robot sensing across five complementary tracks: Driving with Language, Social Navigation, Cross-Sensor Placement 3D Object Detection, Cross-Modal Drone Navigation, and Cross-Platform 3D Object Detection. By providing standardized datasets, baseline systems, unified evaluation protocols, and verified leaderboards, RoboSense enables systematic study of real-world reliability under sensor corruption, viewpoint changes, configuration mismatch, and platform diversity.

Across 143 participating teams from 85 institutions in 16 countries, the top submissions consistently improved upon strong baselines, demonstrating meaningful progress in robust multimodal reasoning, socially compliant navigation, sensor-agnostic perception, cross-view alignment, and cross-platform adaptation. At the same time, the competition results highlight open challenges, particularly in compound distribution shifts, failure calibration, and stable learning under limited or noisy supervision. 

We hope RoboSense will serve as a sustained community effort that accelerates research toward perception systems that generalize reliably beyond training assumptions, ultimately supporting embodied agents that can sense robustly and adapt across platforms in real-world environments.

\section*{Acknowledgment}
This work is under the programme DesCartes and is supported by the National Research Foundation, Prime Minister’s Office, Singapore, under its Campus for Research Excellence and Technological Enterprise (CREATE) programme.

\section{Appendix}
\label{sec:appendix}

\subsection{Public Resources}
\label{sec:public_resources}
\begin{itemize}
    \item \textbf{Website:} \url{https://robosense2025.github.io}

    \item \textbf{Evaluation Server 1:} \url{https://www.codabench.org/competitions/9285}

    \item \textbf{Evaluation Server 2:} \url{https://eval.ai/web/challenges/challenge-page/2557}

    \item \textbf{Evaluation Server 3:} \url{https://www.codabench.org/competitions/9284}

    \item \textbf{Evaluation Server 4:} \url{https://www.codabench.org/competitions/10311}

    \item \textbf{Evaluation Server 5:} \url{https://www.codabench.org/competitions/9179}
\end{itemize}

\subsection{Organizers}
\label{sec:organizers}

\paragraph{Track Leaders}
\begin{itemize}
    \item 
    Lingdong Kong$^{1,2}$, 
    Shaoyuan Xie$^{3}$, 
    Zeying Gong$^{4}$, 
    Ye Li$^{5}$, 
    Meng Chu$^{6}$, and
    Ao Liang$^{1}$
    \item \noindent\textbf{Affiliations:}
    \\
    $^{1}$National University of Singapore
    \\
    $^{2}$CNRS@CREATE
    \\
    $^{3}$University of California, Irvine
    \\
    $^{4}$HKUST(GZ)
    \\
    $^{5}$University of Michigan, Ann Arbor
    \\
    $^{6}$HKUST
\end{itemize}

\paragraph{Track Organizers}
\begin{itemize}
    \item 
    Yuhao Dong$^{1}$, 
    Tianshuai Hu$^{2}$, 
    Ronghe Qiu$^{3}$, 
    Rong Li$^{3}$, 
    Hanjiang Hu$^{4}$,
    Dongyue Lu$^{5}$,
    Wei Yin$^{6}$, 
    Wenhao Ding$^{7}$,
    Linfeng Li$^{5}$, and
    Hang Song$^{8}$

    \item \noindent\textbf{Affiliations:}
    \\
    $^{1}$Nanyang Technological University, Singapore
    \\
    $^{2}$HKUST
    \\
    $^{3}$HKUST(GZ)
    \\
    $^{4}$Carnegie Mellon University
    \\
    $^{5}$National University of Singapore
    \\
    $^{6}$Horizon Robotics
    \\
    $^{7}$NVIDIA Research
    \\
    $^{8}$Xi'an Jiaotong University
\end{itemize}

\paragraph{Program Committee}
\begin{itemize}
    \item 
    Wenwei Zhang$^{1}$, 
    Yuexin Ma$^{2}$, 
    Junwei Liang$^{3}$, 
    Zhedong Zheng$^{4}$,
    Lai Xing Ng$^{5}$,
    Benoit R. Cottereau$^{6,7}$, 
    Wei Tsang Ooi$^{8}$, and 
    Ziwei Liu$^{9}$

    \item \noindent\textbf{Affiliations:}
    \\
    $^{1}$Shanghai AI Laboratory
    \\
    $^{2}$ShanghaiTech University
    \\
    $^{3}$HKUST(GZ)
    \\
    $^{4}$University of Macau
    \\
    $^{5}$Institute for Infocomm Research, A*STAR
    \\
    $^{6}$IPAL, CNRS IRL 2955, Singapore
    \\
    $^{7}$CerCo, CNRS UMR 5549, Université Toulouse III
    \\
    $^{8}$National University of Singapore
    \\
    $^{9}$Nanyang Technological University, Singapore
\end{itemize}

\subsection{Technical Committee}
\label{sec:technical_committee}
\begin{itemize}
    \item 
    Zhanpeng Zhang$^1$, Weichao Qiu$^1$, and Wei Zhang$^1$

    \item \noindent\textbf{Affiliations:}\\
    $^1$HUAWEI Noah's Ark Lab
\end{itemize}

\subsection{Teams \& Affiliations}
\label{sec:teams_affiliations}

\subsubsection{Track 1: Driving with Language}

\paragraph{Team~ [\textcolor{robo_blue}{TQL}]}
\begin{itemize}
    \item \textbf{Team Members:}
    \\
    Jiangpeng Zheng$^1$, Ji Ao$^2$, Siyu Wang$^2$, and Guang Yang$^1$
    \item \textbf{E-mail:}
    \\ 
    \url{aoji.vip@gmail.com}
    \item \textbf{Affiliations:}
    \\
    $^1$Tianjin University of Technology
    \\
    $^2$Independent Researcher
\end{itemize}

\paragraph{Team~ [\textcolor{robo_blue}{UCAS-CSU}]}
\begin{itemize}
    \item \textbf{Team Members:}
    \\
    Aodi Wu$^1$ and Xubo Luo$^1$
    \item \textbf{E-mail:}
    \\ 
    \url{wuaodi20@mails.ucas.ac.cn}
    \item \textbf{Affiliation:}
    \\
    $^1$University of Chinese Academy of Sciences
\end{itemize}

\paragraph{Team~ [\textcolor{robo_blue}{AutoRobots}]}
\begin{itemize}
    \item \textbf{Team Members:}
    \\
    Hanshi Wang$^1$, Xijie Gong$^1$, Yixiang Yang$^1$, Qianli Ma$^1$, and Zhipeng Zhang$^1$
    \item \textbf{E-mail:}
    \\ 
    \url{hanshi.wang.cv@outlook.com}
    \item \textbf{Affiliation:}
    \\
    $^1$AutoLab, School of Artificial Intelligence, Shanghai Jiao Tong University
\end{itemize}

\paragraph{Team~ [\textcolor{robo_blue}{CVML}]}
\begin{itemize}
    \item \textbf{Team Members:}
    \\
    Seungjun Yu$^1$, Junsung Park$^1$, Youngsun Lim$^1$, and Hyunjung Shim$^1$
    \item \textbf{E-mail:}
    \\ 
    \url{seungjunyu@kaist.ac.kr}
    \item \textbf{Affiliation:}
    \\
    $^1$KAIST
\end{itemize}

\paragraph{Team~ [\textcolor{robo_blue}{UQMM}]}
\begin{itemize}
    \item \textbf{Team Members:}
    \\
    Yuxia Fu$^1$, Djamahl Etchegaray$^1$, and Yadan Luo$^1$
    \item \textbf{E-mail:}
    \\ 
    \url{yuxia.fu@uq.edu.au}
    \item \textbf{Affiliation:}
    \\
    $^1$The University of Queensland
\end{itemize}

\subsubsection{Track 2: Social Navigation}

\paragraph{Team~ [\textcolor{robo_blue}{Are Ivan}]}
\begin{itemize}
    \item \textbf{Team Members:}
    \\
    Zihao Zhang$^1$, Yu Zhong$^{1,2}$, Enzhu Gao$^3$, Xinhan Zheng$^4$, Xueting Wang$^4$, Shouming Li$^5$, Yunkai Gao$^3$, Siming Lan$^3$, Mingfei Han$^6$, and Xing Hu$^1$
    \item \textbf{E-mail:}
    \\
    \url{zhangzihao@ict.ac.cn}
    \item \textbf{Affiliations:}
    \\
    $^1$Institute of Computing Technology, Chinese Academy of Sciences
    \\
    $^2$University of Chinese Academy of Sciences
    \\
    $^3$Institute of AI for Industries
    \\
    $^4$University of Science and Technology of China
    \\
    $^5$Beijing University of Technology
    \\
    $^6$Mohamed Bin Zayed University of Artificial Intelligence
\end{itemize}

\paragraph{Team~ [\textcolor{robo_blue}{Xiaomi EV-AD VLA}]}
\begin{itemize}
    \item \textbf{Team Members:}
    \\
    Erjia Xiao$^1$, Lingfeng Zhang$^2$, Yingbo Tang$^3$, Hao Cheng$^1$, Renjing Xu$^1$, Wenbo Ding$^2$, Lei Zhou$^4$, Long Chen$^4$, Hangjun Ye$^4$, and Xiaoshuai Hao$^4$
    \item \textbf{E-mail:}
    \\
    \url{haoxiaoshuai714@163.com}
    \item \textbf{Affiliations:}
    \\
    $^1$The Hong Kong University of Science and Technology (Guangzhou)
    \\
    $^2$Tsinghua Shenzhen International Graduate School, Tsinghua University
    \\
    $^3$Institute of Automation, Chinese Academy of Sciences
    \\
    $^4$Xiaomi EV
\end{itemize}

\paragraph{Team~ [\textcolor{robo_blue}{AutoRobot}]}
\begin{itemize}
    \item \textbf{Team Members:}
    \\
    Wenxiang Shi$^1$, Jingmeng Zhou$^1$, Weijun Zeng$^1$, and Zhipeng Zhang$^1$
    \item \textbf{E-mail:}
    \\
    \url{swx-luna@sjtu.edu.cn}
    \item \textbf{Affiliations:}
    \\
    $^1$AutoLab, School of Artificial Intelligence, Shanghai Jiao Tong University
\end{itemize}

\paragraph{Team~ [\textcolor{robo_blue}{DUO}]}
\begin{itemize}
    \item \textbf{Team Member:}
    \\
    Faduo Liang$^1$
    \item \textbf{E-mail:}
    \\ 
    \url{737175218lfd@gmail.com}
    \item \textbf{Affiliation:}
    \\
    $^1$South China University of Technology
\end{itemize}

\paragraph{Team~ [\textcolor{robo_blue}{CityU-ASL}]}
\begin{itemize}
    \item \textbf{Team Members:}
    \\
    Yang Li$^1$, Congfei Li$^1$, and Yuxiang Sun$^1$
    \item \textbf{E-mail:}
    \\ 
    \url{yang.li.mne@my.cityu.edu.hk}
    \item \textbf{Affiliation:}
    \\
    $^1$City University of Hong Kong
\end{itemize}

\subsubsection{Track 3: Sensor Placement}

\paragraph{Team~ [\textcolor{robo_blue}{LRP}]}
\begin{itemize}
    \item \textbf{Team Members:}
    \\
    Dusan Malic$^{1,2}$, Christian Fruhwirth-Reisinger$^{1,2}$, Alexander Prutsch$^1$, Wei Lin$^3$, Samuel Schulter$^4$, and Horst Possegger$^1$
    \item \textbf{E-mail:}
    \\
    \url{dusan.malic@tugraz.at}
    \item \textbf{Affiliations:}
    \\
    $^1$Institute of Visual Computing, Graz University of Technology
    \\
    $^2$Christian Doppler Laboratory for Embedded Machine Learning
    \\
    $^3$Institute for Machine Learning, Johannes Kepler University Linz
    \\
    $^4$Amazon
\end{itemize}

\paragraph{Team~ [\textcolor{robo_blue}{Point Loom}]}
\begin{itemize}
    \item \textbf{Team Members:}
    \\
    Shuangzhi Li$^1$, Junlong Shen$^1$, and Xingyu Li$^1$
    \item \textbf{E-mail:}
    \\
    \url{shuangzh@ualberta.ca}
    \item \textbf{Affiliation:}
    \\
    $^1$University of Alberta
\end{itemize}

\paragraph{Team~ [\textcolor{robo_blue}{Smartqiu}]}
\begin{itemize}
    \item \textbf{Team Member:}
    \\
    Kexin Xu$^1$
    \item \textbf{E-mail:}
    \\
    \url{kxu10@ualberta.ca}
    \item \textbf{Affiliation:}
    \\
    $^1$University of Alberta
\end{itemize}

\paragraph{Team~ [\textcolor{robo_blue}{DZT328}]}
\begin{itemize}
    \item \textbf{Team Members:}
    \\
    Zihang Wang$^{1,2}$, Yiming Peng$^1$, Guanyu Zong$^1$, Xu Li$^1$, Binghao Wang$^3$, Hao Wei$^4$, Yongxin Ma$^5$, Yunke Shi$^{1,2}$, Shuaipeng Liu$^{1,2}$, and Dong Kong$^6$
    \item \textbf{E-mail:}
    \\
    \url{wangzihanggg@hotmail.com}
    \item \textbf{Affiliations:}
    \\
    $^1$Southeast University
    \\
    $^2$Ruimove.ai
    \\
    $^3$Jiangsu University of Science and Technology
    \\
    $^4$Zhejiang University
    \\
    $^5$Shandong University
    \\
    $^6$Shandong University of Science and Technology
\end{itemize}

\paragraph{Team~ [\textcolor{robo_blue}{seu\_zwk}]}
\begin{itemize}
    \item \textbf{Team Members:}
    \\
    Wenkai Zhu$^1$, Wang Xu$^2$, Linru Li$^3$, Longjie Liao$^1$, Jun Yan$^4$, Benwu Wang$^1$, Xueliang Ren$^1$, Xiaoyu Yue$^{1,5}$, Jixian Zheng$^1$, and Jinfeng Wu$^1$
    \item \textbf{E-mail:}
    \\ 
    \url{zwk.seu@gmail.com}
    \item \textbf{Affiliations:}
    \\
    $^1$Southeast University
    \\
    $^2$Shanghai Jiao Tong University
    \\
    $^3$Nanjing Normal University
    \\
    $^4$Chongqing University
    \\
    $^5$Momenta.ai
\end{itemize}

\subsubsection{Track 4: Cross-Modal Drone Navigation}

\paragraph{Team~ [\textcolor{robo_blue}{TeleAI}]}
\begin{itemize}
    \item \textbf{Team Members:}
    \\
    Linfeng Li$^{1,2}$, Jian Zhao$^1$, Zepeng Yang$^1$, Yuhang Song$^3$, Bojun Lin$^3$, Tianle Zhang$^1$, Yuchen Yuan$^1$, Chi Zhang$^1$, and Xuelong Li$^1$
    \item \textbf{E-mail:}
    \\
    \url{cslineng@gmail.com}
    \item \textbf{Affiliations:}
    \\
    $^1$TeleAI, China Telecom
    \\
    $^2$East China Normal University
    \\
    $^3$National Tsing Hua University
\end{itemize}

\paragraph{Team~ [\textcolor{robo_blue}{rhao\_hur}]}
\begin{itemize}
    \item \textbf{Team Members:}
    \\
    Hao Ruan$^1$, Jinliang Lin$^1$, Zhiming Luo$^1$, Yu Zang$^1$, and Cheng Wang$^1$
    \item \textbf{E-mail:}
    \\
    \url{ruanhao@stu.xmu.edu.cn}
    \item \textbf{Affiliation:}
    \\
    $^1$Xiamen University
\end{itemize}

\paragraph{Team~ [\textcolor{robo_blue}{Xiaomi EV-AD VLA}]}
\begin{itemize}
    \item \textbf{Team Members:}
    \\
    Lingfeng Zhang$^1$, Erjia Xiao$^2$, Yuchen Zhang$^3$, Haoxiang Fu$^4$, Ruibin Hu$^5$, Yanbiao Ma$^6$, Wenbo Ding$^1$, Long Chen$^7$, Hangjun Ye$^7$, and Xiaoshuai Hao$^7$
    \item \textbf{E-mail:}
    \\
    \url{haoxiaoshuai714@163.com}
    \item \textbf{Affiliations:}
    \\
    $^1$Tsinghua Shenzhen International Graduate School, Tsinghua University
    \\
    $^2$The Hong Kong University of Science and Technology (Guangzhou)
    \\
    $^3$Georgia Institute of Technology
    \\
    $^4$National University of Singapore
    \\
    $^5$The Chinese University of Hong Kong
    \\
    $^6$Gaoling School of Artificial Intelligence, Renmin University of China
    \\
    $^7$Xiaomi EV
\end{itemize}

\subsubsection{Track 5: Cross-Platform 3D Object Detection}

\paragraph{Team~ [\textcolor{robo_blue}{Visionary}]}
\begin{itemize}
    \item \textbf{Team Members:}
    \\
    Youngseok Kim$^1$, Sihwan Hwang$^{1,2}$, and Hyeonjun Jeong$^{1,2}$
    \item \textbf{E-mail:}
    \\ \url{youngseok.kim@visionary.run}
    \item \textbf{Affiliations:}
    \\
    $^1$Visionary Inc.
    \\
    $^2$KAIST
\end{itemize}

\paragraph{Team~ [\textcolor{robo_blue}{Point Loom}]}
\begin{itemize}
    \item \textbf{Team Members:}
    \\
    Shuangzhi Li$^1$, Junlong Shen$^1$, and Xingyu Li$^1$
    \item \textbf{E-mail:}
    \\
    \url{shuangzh@ualberta.ca}
    \item \textbf{Affiliation:}
    \\
    $^1$University of Alberta
\end{itemize}

\paragraph{Team~ [\textcolor{robo_blue}{DUTLu\_group}]}
\begin{itemize}
    \item \textbf{Team Members:}
    \\
    Xiyan Feng$^1$, Wenbo Zhang$^1$, Lu Zhang$^1$, Yunzhi Zhuge$^1$, Huchuan Lu$^1$, and You He$^2$
    \item \textbf{E-mail:}
    \\
    \url{fxy@mail.dlut.edu.cn}
    \item \textbf{Affiliations:}
    \\
    $^1$Dalian University of Technology
    \\
    $^2$Shenzhen International Graduate School, Tsinghua University
\end{itemize}

\paragraph{Team~ [\textcolor{robo_blue}{Hunter}]}
\begin{itemize}
    \item \textbf{Team Members:}
    \\
    Yongchun Lin$^1$, Huitong Yang$^2$, Liang Lei$^1$, Haoang Li$^3$, Xinliang Zhang$^4$, Zhiyong Wang$^4$, and Xiaofeng Wang$^4$
    \item \textbf{E-mail:}
    \\
    \url{linyongchun@mails.gdut.edu.cn}
    \item \textbf{Affiliations:}
    \\
    $^1$Guangdong University of Technology
    \\
    $^2$The University of Queensland
    \\
    $^3$The Hong Kong University of Science and Technology (Guangzhou)
    \\
    $^4$SenseTime
\end{itemize}

\paragraph{Team~ [\textcolor{robo_blue}{TeamArcN}]}
\begin{itemize}
    \item \textbf{Team Members:}
    \\
    Shurui Qin$^1$, Wei Cong$^1$, and Yao He$^1$
    \item \textbf{E-mail:}
    \\
    \url{glfgpbt@163.com}
    \item \textbf{Affiliation:}
    \\
    $^1$South China University of Technology
\end{itemize}

\subsection{Track Reports}
\label{sec:track_reports}

\subsubsection{Track 1: Driving with Language}
\begin{itemize}
    \item Jiangpeng Zheng, Ji Ao, Guang Yang, and Siyu Wang: \textbf{Enhancing Multi-View Driving VLMs via Pseudo-Label Pretraining and Long-Tail Balancing} [\href{https://github.com/robosense2025/toolkit/blob/main/technical_reports/RoboSense2025_Track1_TQL.pdf}{Download}]

    \item Aodi Wu and Xubo Luo: \textbf{Enhancing VLMs for Autonomous Driving through Task-Specific Prompting and Spatial Reasoning} [\href{https://github.com/robosense2025/toolkit/blob/main/technical_reports/RoboSense2025_Track1_UCAS_CSU.pdf}{Download}]

    \item Hanshi Wang, Xijie Gong, Yixiang Yang, Qianli Ma, and Zhipeng Zhang: \textbf{Task Aware Prompt Routing and CoT Augmented Fine Tuning for Driving VQA} [\href{https://github.com/robosense2025/toolkit/blob/main/technical_reports/RoboSense2025_Track1_AutoRobots.pdf}{Download}]

    \item Seungjun Yu, Junsung Park, Youngsun Lim, and Hyunjung Shim: \textbf{Towards Robust Autonomous Driving Question-Answering through Metadata-Grounded Context and Task-Specific Prompts} [\href{https://github.com/robosense2025/toolkit/blob/main/technical_reports/RoboSense2025_Track1_AutoRobots.pdf}{Download}]

    \item Yuxia Fu, Djamahl Etchegaray, and Yadan Luo: \textbf{Driving Robustly through Corruptions: Multi-Source LoRA Fine-Tuning of Driving VLMs for Multi-View Reasoning} [\href{https://github.com/robosense2025/toolkit/blob/main/technical_reports/RoboSense2025_Track1_UQMM.pdf}{Download}]
\end{itemize}

\subsubsection{Track 2: Social Navigation}
\begin{itemize}
    \item Zihao Zhang, Yu Zhong, Enzhu Gao, Xinhan Zheng, Xueting Wang, Shouming Li, Yunkai Gao, Siming Lan, Mingfei Han, and Xing Hu: \textbf{PER-Falcon: Positive-Episode Replay for Future-Aware Social Navigation} [\href{https://github.com/robosense2025/toolkit/blob/main/technical_reports/RoboSense2025_Track2_AreIvan.pdf}{Download}]

    \item Erjia Xiao, Lingfeng Zhang, Yingbo Tang, Hao Cheng, Renjing Xu, Wenbo Ding, Lei Zhou, Long Chen, Hangjun Ye, and Xiaoshuai Hao: \textbf{Learning to Navigate Socially Through Proactive Risk Perception} [\href{https://github.com/robosense2025/toolkit/blob/main/technical_reports/RoboSense2025_Track2_XiaomiEV_AD_VLA.pdf}{Download}]

    \item Wenxiang Shi, Jingmeng Zhou, Weijun Zeng, and Zhipeng Zhang: \textbf{From Imitation to Interaction: A Two-Stage Training Paradigm for Social Navigation} [\href{https://github.com/robosense2025/toolkit/blob/main/technical_reports/RoboSense2025_Track2_AutoRobot.pdf}{Download}]

    \item Faduo Liang: \textbf{Enhancing Robust Social Navigation with Cutout: Handling Human Occlusion in Dynamic Environments} [\href{https://github.com/robosense2025/toolkit/blob/main/technical_reports/RoboSense2025_Track2_DUO.pdf}{Download}]

    \item Yang Li, Congfei Li, and Yuxiang Sun: \textbf{Towards Socially Compliant Navigation: Hybrid Parameter Optimization for Falcon in Dynamic Environments} [\href{https://github.com/robosense2025/toolkit/blob/main/technical_reports/RoboSense2025_Track2_CityU_ASL.pdf}{Download}]
\end{itemize}

\subsubsection{Track 3: Sensor Placement}
\begin{itemize}
    \item Dusan Malic, Christian Fruhwirth-Reisinger, Alexander Prutsch, Wei Lin, Samuel Schulter, and Horst Possegger: \textbf{GBlobs: Local LiDAR Geometry for Improved Sensor Placement Generalization} [\href{https://github.com/robosense2025/toolkit/blob/main/technical_reports/RoboSense2025_Track3_LRP.pdf}{Download}]

    \item Junlong Shen, Shuangzhi Li, and Xingyu Li: \textbf{Robust 3D Object Detection under Sensor Placement Variability} [\href{https://github.com/robosense2025/toolkit/blob/main/technical_reports/RoboSense2025_Track3_PointLoom.pdf}{Download}]

    \item Kexin Xu: \textbf{Towards Generalizable 3D Object Detection Across Sensor Placements} [\href{https://github.com/robosense2025/toolkit/blob/main/technical_reports/RoboSense2025_Track3_Smartqiu.pdf}{Download}]

    \item Zihang Wang, Yiming Peng, Guanyu Zong, Xu Li, Binghao Wang, Hao Wei, Yongxin Ma, Yunke Shi, Shuaipeng Liu, and Dong Kong: \textbf{PlaceRecover: A Transformer-based Point Cloud Recovery Network with Implicit Neural Representations for Robust LiDAR Placement Adaptation} [\href{https://github.com/robosense2025/toolkit/blob/main/technical_reports/RoboSense2025_Track3_DZT328.pdf}{Download}]

    \item Wenkai Zhu, Xu Li, Wang Xu, Linru Li, Longjie Liao, Benwu Wang, Xueliang Ren, Xiaoyu Yue, JiXian Zheng, and Jinfeng Wu: \textbf{Layout-Robust LiDAR 3D Object Detection via Multi-Representation Fusion} [\href{https://github.com/robosense2025/toolkit/blob/main/technical_reports/RoboSense2025_Track3_seu_zwk.pdf}{Download}]
\end{itemize}

\subsubsection{Track 4: Cross-Modal Drone Navigation}
\begin{itemize}
    \item Linfeng Li, Jian Zhao, Zepeng Yang, Yuhang Song, Bojun Lin, Tianle Zhang, Yuchen Yuan, Chi Zhang, and Xuelong Li: \textbf{A Parameter-Efficient Mixture-of-Experts Framework for Cross-Modal Geo-Localization} [\href{https://github.com/robosense2025/toolkit/blob/main/technical_reports/RoboSense2025_Track4_TeleAI.pdf}{Download}]

    \item Hao Ruan, Jinliang Lin, Zhiming Luo, Yu Zang, and Cheng Wang: \textbf{HCCM: Hierarchical Cross-Granularity Contrastive and Matching Learning for Cross-Modal Drone Navigation} [\href{https://github.com/robosense2025/toolkit/blob/main/technical_reports/RoboSense2025_Track4_rhao_hur.pdf}{Download}]

    \item Lingfeng Zhang, Erjia Xiao, Yuchen Zhang, Haoxiang Fu, Ruibin Hu, Yanbiao Ma, Wenbo Ding, Long Chen, Hangjun Ye, and Xiaoshuai Hao: \textbf{Caption-Guided Retrieval System for Cross-Modal Drone Navigation} [\href{https://github.com/robosense2025/toolkit/blob/main/technical_reports/RoboSense2025_Track4_XiaomiEV_AD_VLA.pdf}{Download}]
\end{itemize}

\subsubsection{Track 5: Cross-Platform 3D Object Detection}
\begin{itemize}
    \item Youngseok Kim, Sihwan Hwang, and Hyeonjun Jeong: \textbf{DataEngine: Unified Pre-Training and Viewpoint Normalization for Cross-Platform 3D Object Detection} [\href{https://github.com/robosense2025/toolkit/blob/main/technical_reports/RoboSense2025_Track5_Visionary.pdf}{Download}]

    \item Shuangzhi Li, Junlong Shen, and Xingyu Li: \textbf{Robust 3D Object Detection via Physical-Aware Augmentation and Class-Specific Model Ensembling} [\href{https://github.com/robosense2025/toolkit/blob/main/technical_reports/RoboSense2025_Track5_PointLoom.pdf}{Download}]

    \item Xiyan Feng, Wenbo Zhang, Lu Zhang, Yunzhi Zhuge, Huchuan Lu, and You He: \textbf{Towards Cross-Platform Generalization: Domain Adaptive 3D Detection with Augmentation and Pseudo-Labeling} [\href{https://github.com/robosense2025/toolkit/blob/main/technical_reports/RoboSense2025_Track5_DUTLu_group.pdf}{Download}]

    \item Yongchun Lin, Huitong Yang, Liang Lei, Haoang Li, Xinliang Zhang, Zhiyong Wang, and Xiaofeng Wang: \textbf{SegSy3D: Segmentation-Guided Self-Training and Model Synergy for Cross-Platform 3D Detection} [\href{https://github.com/robosense2025/toolkit/blob/main/technical_reports/RoboSense2025_Track5_Hunter.pdf}{Download}]

    \item Shurui Qin, Gan Sun, Yao He, and Wei Cong: \textbf{Unsupervised Domain Adaptation for 3D Object Detection via Adversarial Learning} [\href{https://github.com/robosense2025/toolkit/blob/main/technical_reports/RoboSense2025_Track5_TeamArcN.pdf}{Download}]
\end{itemize}

{\small
\bibliographystyle{ieeenat_fullname}
\bibliography{main}
}

\end{document}